\RequirePackage{fix-cm}
\documentclass[twocolumn]{svjour3}          
\smartqed  
\usepackage{natbib}
\usepackage{graphicx}
\usepackage{epstopdf}
\usepackage{amsmath}
\usepackage{amssymb}
\usepackage{fixltx2e}
\usepackage[hidelinks]{hyperref}
\usepackage{nicefrac}
\usepackage{multirow}
\usepackage{amsmath,amssymb}
\usepackage{algorithmic}
\usepackage{subfigure}
\usepackage{color}
\usepackage{array}
\usepackage{tabu}
\usepackage{adjustbox}
\usepackage{color}

\begin{document}

\title{Hierarchical Cellular Automata for Visual Saliency
}


\author{Yao~Qin*     \and
        Mengyang~Feng* \and
        Huchuan Lu    \and
        Garrison~W.~Cottrell
}


\institute{Yao Qin \at
              University of California, San Diego\\
              \email{yaq007@eng.ucsd.edu}           
           \and
           Mengyang Feng \at
              Dalian University of Technology\\
              \email{mengyangfeng@gmail.com}
           \and
           Huchuan Lu \at
              Dalian University of Technology\\
              \email{lhchuan@dlut.edu.cn}
           \and
           Garrsion W. Cottrell \at
              University of California, San Diego\\
              \email{gary@eng.ucsd.edu}
          \and
          * Equal Contribution
}
\date{Received: date / Accepted: date}

\maketitle

\begin{abstract}
Saliency detection, finding the most important parts of an image, has become increasingly popular in computer vision. In this paper, we introduce Hierarchical Cellular Automata (HCA) -- a temporally evolving model to intelligently detect salient objects. HCA consists of two main components: Single-layer Cellular Automata (SCA) and Cuboid Cellular Automata (CCA). As an unsupervised propagation mechanism, Single-layer Cellular Automata can exploit the intrinsic relevance of similar regions through interactions with neighbors. Low-level image features as well as high-level semantic information extracted from deep neural networks are incorporated into the SCA to measure the correlation between different image patches. With these hierarchical deep features, an impact factor matrix and a coherence matrix are constructed to balance the influences on each cell's next state. The saliency values of all cells are iteratively updated according to a well-defined update rule. Furthermore, we propose CCA to integrate multiple saliency maps generated by SCA at different scales in a Bayesian framework. Therefore, single-layer propagation and multi-layer integration are jointly modeled in our unified HCA. Surprisingly, we find that the SCA can improve all existing methods that we applied it to, resulting in a similar precision level regardless of the original results. The CCA can act as an efficient pixel-wise aggregation algorithm that can integrate state-of-the-art methods, resulting in even better results. Extensive experiments on four challenging datasets demonstrate that the proposed algorithm outperforms state-of-the-art conventional methods and is competitive with deep learning based approaches.
\end{abstract}

\section{Introduction}
\label{introduction}
Humans excel in identifying visually significant regions in a scene corresponding to salient objects. Given an image, people can quickly tell what attracts them most. In the field of computer vision, however, performing the same task is very challenging, despite dramatic progress in recent years. To mimic the human attention system, many researchers focus on developing computational models that locate regions of interest in the image. Since accurate saliency maps can assign relative importance to the visual contents in an image, saliency detection can be used as a pre-processing procedure to narrow the scope of visual processing and reduce the cost of computing resources. As a result, saliency detection has raised a great amount of attention~\citep{achanta2009frequency,goferman2010context} and has been incorporated into various computer vision tasks, such as visual tracking~\citep{mahadevan2009saliency}, object retargeting~\citep{ding2011importance,sun2011scale} and image categorization~\citep{siagian2007rapid,kanan2010robust}.
\begin{figure}
  \centering
  \includegraphics[width=8.35cm]{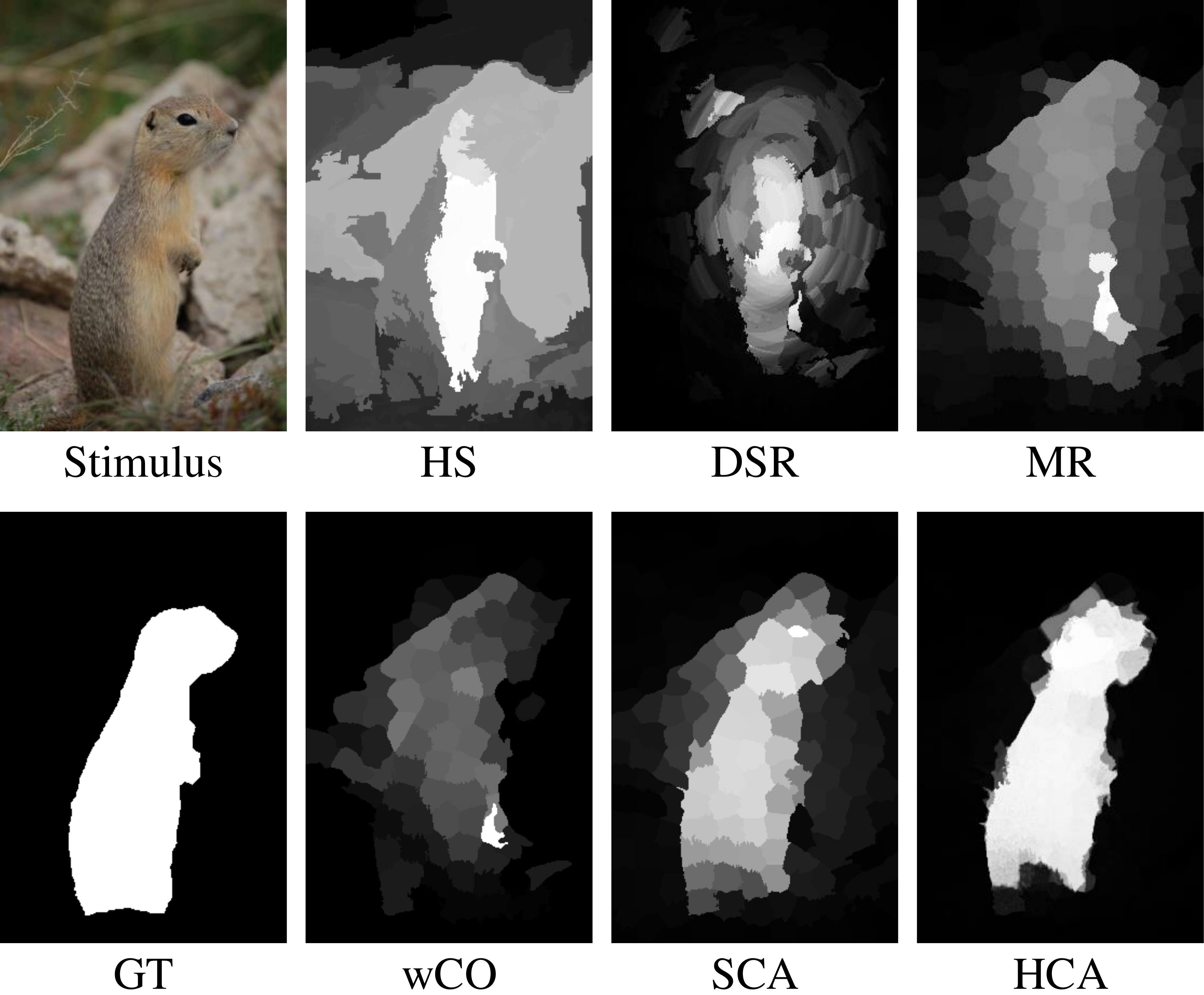}\\
  \caption{An example illustrates that conventional saliency detection methods based on handcrafted low-level features fail in complex circumstances. From top left to bottom right: stimulus, HS~\citep{yan2013hierarchical}, DSR~\citep{li2013saliency}, MR~\citep{yang2013saliency}, ground truth, wCO~\citep{zhu2014saliency}, and our method SCA and HCA.}\label{contrastfail}
  \vspace{-5mm}
\end{figure}
\begin{figure*}
  \centering
  \includegraphics[width=17.46cm]{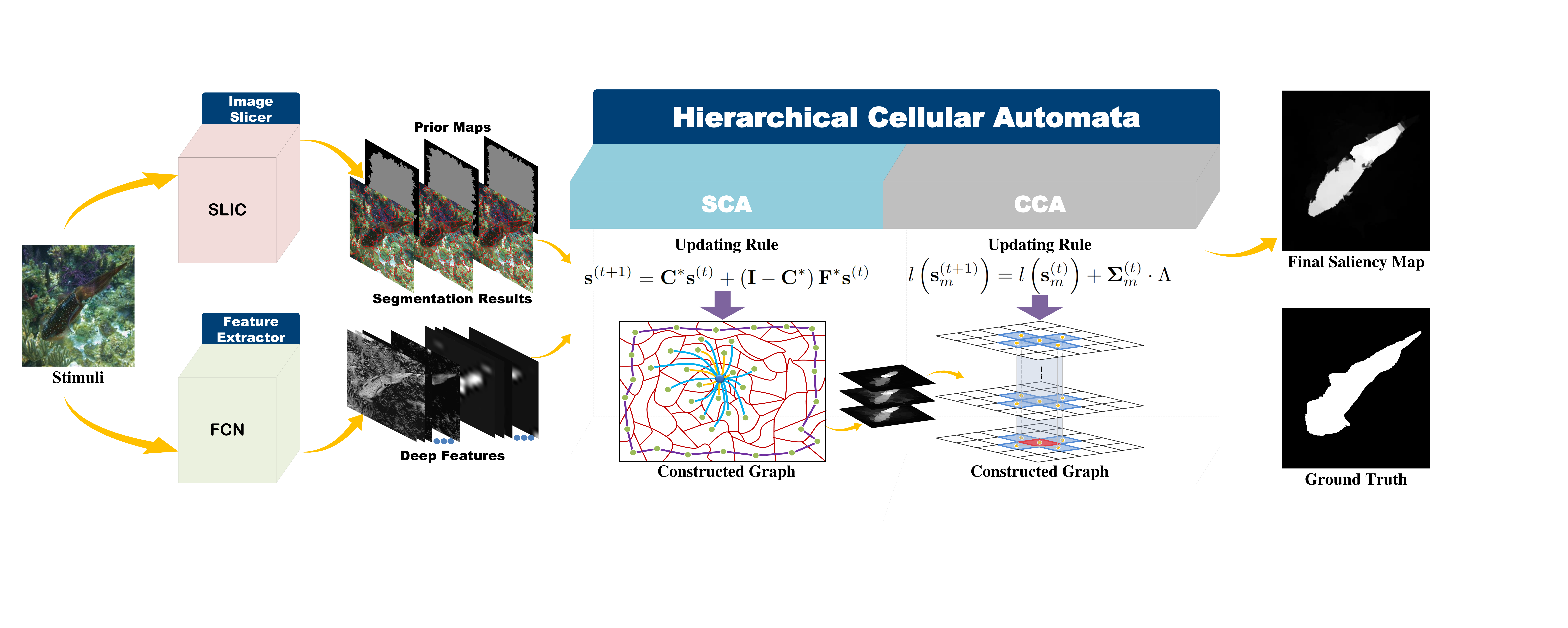}\\
  \caption{The pipeline of our proposed Hierarchical Cellular Automata. First, the stimulus is segmented into multi-scale superpixels, and superpixels on the image boundary are selected as seeds for the propagation of the background (Section.~\ref{initialmaps}). Then FCN-32s~\citep{long2015fully} is used as a feature extractor to obtain deep features (Section.~\ref{deepfeature}). The generated prior maps and deep features are both fed into the Single-Layer Cellular Automata (Section.~\ref{SCA}) to create multi-scale saliency maps. Finally, we integrate these saliency maps via the Cuboid Cellular Automata (Section.~\ref{CCA}) to obtain our ultimate result.}\label{pipeline}
\end{figure*}
Results in perceptual research show that contrast is one of the decisive factors in the human visual attention system~\citep{itti2001computational,reinagel1999natural}, suggesting that salient objects are most likely in the region of the image that significantly differs from its surroundings. Many conventional saliency detection methods focus on exploiting local and global contrast based on various handcrafted image features, e.g., color features~\citep{liu2011learning,cheng2015global}, focusness~\citep{jiang2013salientufo}, textual distinctiveness~\citep{scharfenberger2013statistical}, and structure descriptors~\citep{shi2013pisa}. Although these methods perform well on simple benchmarks, they may fail in some complex situations where the handcrafted low-level features do not help salient objects stand out from the background. For example, in Figure.~\ref{contrastfail}, the prairie dog is surrounded by low-contrast rocks and bushes. It is challenging to detect the prairie dog as a salient object with only low-level saliency cues. However, humans can easily recognize the prairie dog based on its category as it is semantically salient in high-level cognition and understanding.

In addition to the limitation of low-level features, the large variations in object scales also restrict the accuracy of saliency detection. An appropriate scale is of great importance in extracting the salient object from the background. One of the most popular ways to detect salient objects of different sizes is to construct multi-scale saliency maps and then aggregate them with pre-defined functions, such as averaging or a weighted summation. In most existing methods~\citep{Wang_2016_CVPR, li2015visual,Li_2014_CVPR,Zhou_2014_CVPR,borji2015salient}, however, these constructed saliency maps are usually integrated in a simple and heuristic way, which may directly limit the precision of saliency aggregation.

To address these two obvious problems, we propose a novel method named Hierarchical Cellular Automata (HCA) to extract the salient objects from the background efficiently. A Hierarchical Cellular Automata consists of two main components: Single-layer Cellular Automata (SCA) and Cuboid Cellular Automata (CCA). First, to improve the features, we use fully convolutional networks~\citep{long2015fully} to extract deep features due to their successful application to semantic segmentation. It has been demonstrated that \emph{deep features} are highly versatile and have stronger representational power than traditional handcrafted features~\citep{krizhevsky2012imagenet,farabet2013learning,girshick2014rich}. Low-level image features and high-level saliency cues extracted from deep neural networks are used by an SCA to measure the similarity of neighbors. With these hierarchical deep features, the SCA iteratively updates the saliency map through interactions with similar neighbors. Then the salient object will naturally emerge from the background with high consistency among similar image patches. Secondly,
to detect multi-scale salient objects, we apply the SCA at different scales and integrate them with the CCA based on Bayesian inference. Through interactions with neighbors in a cuboid zone, the integrated saliency map can highlight the foreground and suppress the background. An overview of our proposed HCA is shown in Figure.~\ref{pipeline}.

Furthermore, the Hierarchical Cellular Automata is capable of optimizing other saliency detection methods. If a saliency map generated by one of the existing methods is used as the prior map and fed into HCA, it can be improved to the state-of-the-art precision level. Meanwhile, if multiple saliency maps generated by different existing methods are used as initial inputs, HCA can naturally fuse these saliency maps and achieve a result that outperforms each method.

In summary, the main contributions of our work include:\\
(1) We propose a novel Hierarchical Cellular Automata to adaptively detect salient objects of different scales based on hierarchical deep features. The model effectively improves all of the methods we have applied it to to state-of-the-art precision levels and is relatively insensitive to the original maps.\\
(2) Single-layer Cellular Automata serve as a propagation mechanism that exploits the intrinsic relevance of similar regions via interactions with neighbors. \\
(3) Cuboid Cellular Automata can integrate multiple saliency maps into a more favorable result under the Bayesian framework.

\section{Related Work}\label{related work}
\vspace{-2mm}
\subsection{Salient Object Detection}
\vspace{-3mm}
Methods of saliency detection can be divided into two categories: top-down (task-driven) methods and bottom-up (data-driven) methods. Approaches like~\citep{alexe2010object,marchesotti2009framework,ng2002spectral,yang2012top} are typical top-down visual attention methods that require supervised learning with manually labeled ground truth. To better distinguish salient objects from the background, high-level category-specific information and supervised methods are incorporated to improve the accuracy of saliency maps. In contrast, bottom-up methods usually concentrate on low-level cues such as color, intensity, texture and orientation to construct saliency maps~\citep{hou2007saliency,jiang2011automatic,klein2011center,sun2012saliency,tong2015salient,yan2013hierarchical}. Some global bottom-up approaches tend to build saliency maps by calculating the holistic statistics on uniqueness of each element over the whole image~\citep{cheng2015global,perazzi2012saliency,bruce2005saliency}.

As saliency is defined as a particular part of an image that visually stands out compared to their neighboring regions or the rest of image, one of the most used principles, \emph{contrast prior}, measures the saliency of a region according to the color contrast or geodesic distance against its surroundings~\citep{cheng2013efficient,cheng2015global,jiang2011automatic,jiang2013submodular,klein2011center,perazzi2012saliency,wang2011automatic}. Recently, the \emph{boundary prior} has been introduced in several methods based on the assumption that regions along the image boundaries are more likely to be the background~\citep{jiang2013salient,li2013saliency,wei2012geodesic,yang2013saliency,borji2015salient,shen2012unified}, although this takes advantage of photographer's bias and is less likely to be true for active robots. Considering the connectivity of regions in the background, \citet{wei2012geodesic} define the saliency value for each region as the shortest-path distance towards the boundary. \citet{yang2013saliency} use manifold ranking to infer the saliency score of image regions according to their relevance to boundary superpixels. Furthermore, in~\citep{jiang2013saliency}, the contrast against the image border is used as a new regional feature vector to characterize the background.

However, one of the fundamental problems with all these conventional saliency detection methods is that the features used are not representative enough to capture the contrast between foreground and background, and this limits the precision of saliency detection. For one thing, low-level features cannot help salient objects stand out from a low-contrast background with similar visual appearance. Also, the extracted global features are weak in capturing semantic information and have much poorer generalization compared to the deep features used in this paper.

\vspace{-5mm}
\subsection{Deep Neural Networks}
\vspace{-3mm}
Deep convolutional neural networks have recently achie-
ved a great success in various computer vision tasks, including image classification~\citep{krizhevsky2012imagenet,szegedy2015going}, object detection~\citep{girshick2014rich,hariharan2014simultaneous,szegedy2013deep} and semantic segmentation~\citep{long2015fully,pinheiro2014recurrent}. With the rapid development of deep neural networks, researchers have begun to construct effective neural networks for saliency detection~\citep{zhao2015saliency, li2015visual, zou2015harf, wang2015deep, Li2016DeepSaliency, ssd2016eccv}. In~\citep{zhao2015saliency}, Zhao \emph{et al.} propose a unified multi-context deep neural network taking both global and local context into consideration. Li \emph{et al.}~\citep{li2015visual} and Zou \emph{et al.}~\citep{zou2015harf} explore high-quality visual features extracted from DNNs to improve the accuracy of saliency detection. DeepSaliency in~\citep{Li2016DeepSaliency} is a multi-task deep neural network using a collaborative feature learning scheme between two correlated tasks, saliency detection and semantic segmentation, to learn better feature representation. One leading factor for the  success of deep neural networks is the powerful expressibility and strong capacity of deep architectures that facilitate learning high-level features with semantic information~\citep{hariharan2015hypercolumns,ma2015hierarchical}.

In~\citep{donahue2014decaf}, Donahue~\emph{et al.} point out that features extracted from the activation of a deep convolutional network can be repurposed to other generic tasks. Inspired by this idea, we use the hierarchical deep features extracted from fully convolutional networks~\citep{long2015fully} to represent smaller image regions. The extracted deep features incorporate low-level features as well as high-level semantic information of the image and can be fed into our Hierarchical Cellular Automata to measure the similarity of different image patches.
\vspace{-6mm}
\subsection{Cellular Automata}
\vspace{-3mm}
Cellular Automata are a model of computation first proposed by \citet{von1951general}. They can be described as a temporally evolving system with simple construction but complex self-organizing behavior. A Cellular Automaton consists of a lattice of cells with discrete states, which evolve in discrete time steps according to specific rules. Each cell's next state is determined by its current state as well as its nearest neighbors' states. Cellular Automata have been applied to simulate the evolution of many complicated dynamical systems~\citep{batty2007cities,chopard2005cellular,cowburn2000room,de2003stochastic,martins2008continuous,pan2016opinion}.
Considering that salient objects are spatially coherent, we introduce Cellular Automata into this field and propose Single-layer Cellular Automata as an unsupervised propagation mechanism to exploit the intrinsic relationships of neighboring elements of the saliency map and eliminate gaps between similar regions.

In addition, we propose a method to combine multiple saliency maps generated by different algorithms, or combine saliency maps at different scales through what we call Cuboid Cellular Automata (CCA). In CCA, states of the automaton are determined by a cuboid neighborhood corresponding to automata at the same location as well as their adjacent neighbors in different saliency maps. An illustration of the idea is in Figure~\ref{graph}(b). In this setting, the saliency maps are iteratively updated  through interactions among neighbors in the cuboid zone. The state updates are determined through Bayesian evidence combination rules. Variants of this type of approach have been used before~\citep{rahtu2010segmenting,xie2011visual,xie2013bayesian, li2013saliency}. \citet{xie2013bayesian} use the low-level visual cues derived from a convex hull to compute the observation likelihood. \citet{li2013saliency} construct saliency maps through dense and sparse reconstruction and propose a Bayesian algorithm to combine them. Using Bayesian updates to combine saliency maps puts the algorithm for Cuboid Cellular Automata on a firm theoretical foundation.

\vspace{-6mm}
\section{Proposed algorithm}\label{proalg}
\vspace{-3mm}
In this paper, we propose an unsupervised Hierarchical Cellular Automata (HCA) for saliency detection, composed of two sub-units, a Single-layer Cellular Automata (SCA), and a Cuboid Cellular Automata (CCA), as described below. First, we construct prior maps of different scales with superpixels on the image boundary chosen as the background seeds. Then, hierarchical deep features are extracted from fully convolutional networks~\citep{long2015fully} to measure the similarity of different superpixels. Next, we use SCA to iteratively update the prior maps at different scales based on the hierarchical deep features. Finally, a CCA is used to integrate the multi-scale saliency maps using Bayesian evidence combination. Figure.~\ref{pipeline} shows an overview of our proposed method.
\vspace{-5mm}
\subsection{Background Priors}\label{initialmaps}
\vspace{-2mm}

Recently, there have been various mathematical models proposed to generate a coarse saliency map to help locate potential salient objects in an image~\citep{tong2015bootstrap,zhu2014saliency,Gong_2015_CVPR}. Even though prior maps are effective in improving detection precision, they still have several drawbacks. For example, a poor prior map may greatly limit the accuracy of the final saliency map if it incorrectly estimates the location of the objects or classifies the foreground as the background. Also, the computational time to construct a prior map can be excessive. Therefore, in this paper, we build a quite simple and time-efficient prior map that only provides the propagation seeds for HCA, which is quite insensitive to the prior map and is able to refine this coarse prior map into an improved saliency map.

First, we use the efficient Simple Linear Iterative Clustering (SLIC) algorithm~\citep{achanta2010slic} to segment the image into smaller superpixels in order to capture the essential structural information of the image. Let $s_i\in$~$\mathbb{R}$ denote the saliency value of the superpixel $i$ in the image. Based on the assumption that superpixels on the image boundary tend to have a higher probability of being the background, we assign a close-to-zero saliency value to the boundary superpixels. For others, we assign a uniform value as their initial saliency values,
\begin{equation}\label{impactfactor}
  {s_{i}} = \left\{ {\begin{array}{*{20}{c}}
   {0.001} & {i \in {\text{boundary}}} \\
   0.5 & i\not\in{\text{boundary}}. \\
\end{array}} \right.
\end{equation}
Considering the great variation in the scales of salient objects, we segment the image into superpixels at $M$ different scales, which are displayed in Figure.~\ref{pipeline} (\textbf{Prior Maps}).
\vspace{-4mm}
\subsection{Deep Features from FCN}\label{deepfeature}
\vspace{-2mm}
 As is well-known, the features in the last layers of CNNs encode semantic abstractions of objects, and are robust to appearance variations, while the early layers contain low-level image features, such as color, edge, and texture. Although high-level features can effectively discriminate the objects from various backgrounds, they cannot precisely capture the fine-grained low-level information due to their low spatial resolution. Therefore, a combination of these deep features is preferred compared to any individual feature map.

In this paper, we use the feature maps extracted from the fully-convolutional network (FCN-32s~\citep{long2015fully}) to encode object appearance. The input image to FCN-32s is resized to $500 \times 500$, and a 100-pixel padding is added to the four boundaries. Due to subsampling and pooling operations in the CNN, the outputs of each convolutional layer in the FCN framework are not at the same resolution. Since we only care about the features corresponding to the original image, we need to 1) crop the feature maps to get rid of the padding; 2) resize each feature map to the input image size via the nearest neighbor interpolation. Then each feature map can be aggregated using a simple linear combination as:
 \vspace{-2mm}
\begin{equation}\label{hh}
  \vspace{-2mm}
g(\textbf{r}_i,\textbf{r}_j)=\sum_{l=1}^L {\rho_l} \cdot \| {df^{l}_i-df^{l}_j}\|_2,
\end{equation}
where $df_i^l$ denotes the deep features of superpixel $i$ on the $l$-th layer and $\rho_l$ is a weighting of the importance of the $l$-th feature map, which we set by cross-validation. The weights are constrained to sum to 1:  $\sum^L_{l=1}\rho_l = 1$. Each superpixel is represented by the mean of the deep features of all contained pixels. The computed $g(\textbf{r}_i,\textbf{r}_j)$ is used to measure the similarity between superpixels.
\vspace{-4mm}
\subsection{Hierarchical Cellular Automata}
\vspace{-2mm}
Hierarchical Cellular Automata (HCA) is a unified framework composed of single-layer propagation (Single-layer Cellular Automata) and multi-layer aggregation (Cuboid Cellular Automata). It can generate saliency maps at different scales and integrate them to get a fine-grained saliency map. We will discuss SCA and CCA respectively in Sections~\ref{SCA} and ~\ref{CCA}.
\vspace{-3mm}
\subsubsection{Single-layer Cellular Automata}\label{SCA}
\vspace{-2mm}
In Single-layer Cellular Automata (SCA), each cell denotes a superpixel generated by the SLIC algorithm. SLIC takes the number of desired superpixels as a parameter, so by using different numbers of superpixels with SCA, we can obtain maps at different scales. In this section, we assume one scale, denoted $m$. We denote the number of superpixels in scale $m$ by $n_m$, but we omit the subscript $m$ in most notation in this section for clarity, \textit{e.g.}, $\textbf{F}$ for $\textbf{F}_m$, $\textbf{C}$ for $\textbf{C}_m$ and $\textbf{s}$ for $\textbf{s}_m$.

\begin{figure}
  \centering
  \includegraphics[width=8.4cm]{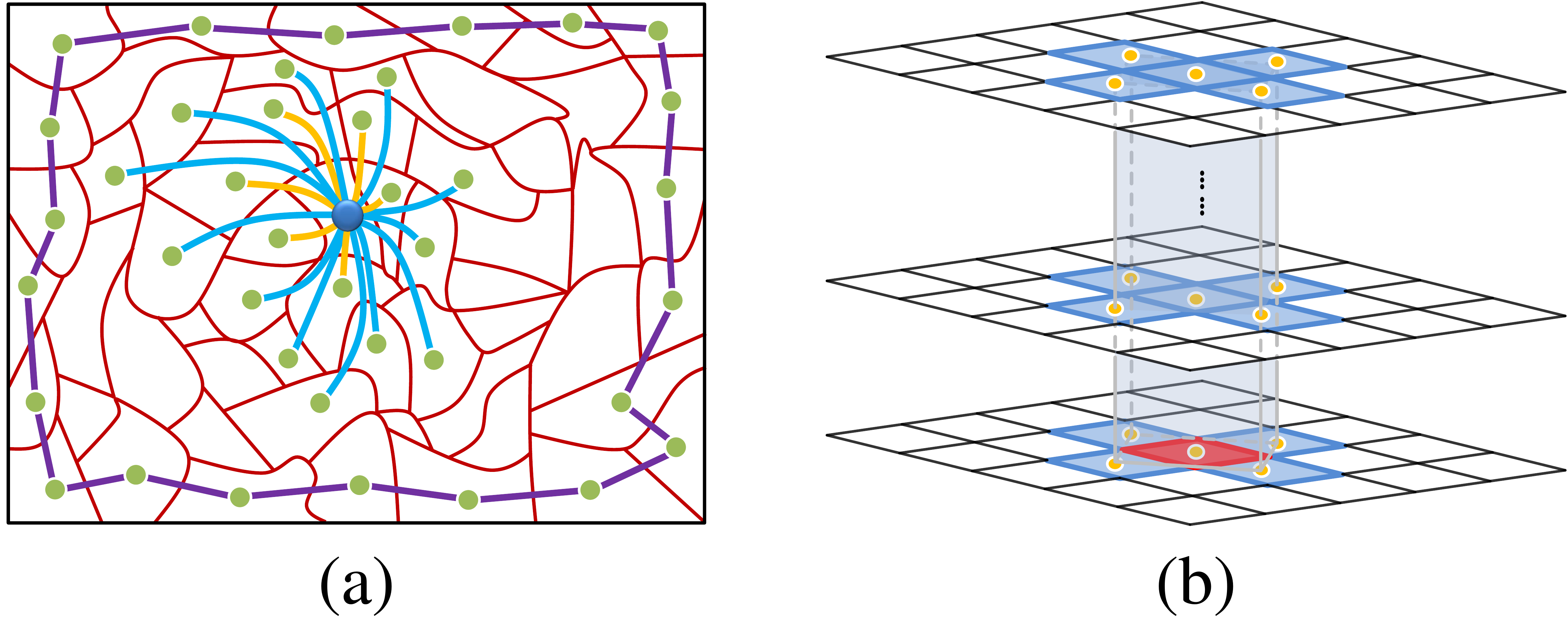}\\
    \vspace{-3mm}
    \caption{The constructed graph models used in our algorithm. (a) is used in SCA, the orange lines and the blue lines represent the connections between the blue center cell and its 2-layer neighbors. The purple lines indicate that superpixels on the image boundaries are all connected to each other; (b) is used in CCA, a cell (e.g.~the red pixel in the bottom layer) is connected to the pixels with the same coordinates in other layers as well as their four adjacent neighbors (e.g.~cells in blue color). All these pixels construct a cuboid interaction zone.}\label{graph}
  \vspace{-4mm}
\end{figure}

We make three major modifications to the previous cellular automata models~\citep{smith1972real,von1951general} for saliency detection.
First, the states of cells in most existing Cellular Automata models are discrete~\citep{von1966theory,wolfram1983statistical}. However, in this paper, we use the saliency value of each superpixel as its state, which is continuous between 0 and 1.
Second, we give a broader definition of the neighborhood that is similar to the concept of $z$-layer neighborhood (here $z=2$) in graph theory. The $z$-layer neighborhood of a cell includes adjacent  cells as well as those sharing common boundaries with its adjacent cells. Also, we assume that superpixels on the image boundaries are all connected to each other because all of them serve as background seeds. The connections between the neighbors are clearly illustrated in Figure.~\ref{graph} (a). Finally, instead of uniform influence of the neighbors , the influence is based on the similarity between the neighbor to the cell in feature space, as explained next.

\textbf{Impact Factor Matrix:}\label{IFM}
Intuitively, neighbors with more similar features have a greater influence on the cell's next state. The similarity of any pair of superpixels is measured by a pre-defined distance in feature space. For the $m$-th saliency map, which has $n_m$ superpixels in total, we construct an impact factor matrix $\textbf{F}\in \mathbb{R}^{n_m \times n_m}$. Each element  $f_{ij}$ in $\textbf{F}$ defines the impact factor of superpixel $i$ to $j$ as:
  \vspace{-1mm}
  \begin{equation}\label{impactfactor}
  {f_{ij}} = \left\{ {\begin{array}{*{20}{c}}
   {\exp (\frac{{ - g(\textbf{r}_i,\textbf{r}_j)}}{{{\sigma_f ^2}}})} & {j \in \text{NB}(i)}  \\
   0 & j = i~\text{or}~\text{otherwise},  \\
\end{array}} \right.
\end{equation}
where $g(\textbf{r}_i,\textbf{r}_j)$ is a function that computes the distance between the superpixel $i$ and $j$ in feature space with $\textbf{r}_i$ as the feature descriptor of superpixel $i$. In this paper, we use the weighted distance of hierarchical deep features computed by Eqn.~(\ref{hh}) to measure the similarity between neighbors. $\sigma_f$ is a parameter that controls the strength of similarity and $\text{NB}(i)$ is the set of the neighbors of the cell $i$. In order to normalize the impact factors, a degree matrix $\textbf{D}=diag\{d_{1},d_{2},\cdots,d_{n_m}\}$ is constructed, where $d_{i}=\sum_{i}f_{ij}$. Finally, a row-normalized impact factor matrix can be calculated as $\textbf{F}^{\ast} = \textbf{D}^{-1} \cdot \textbf{F}$.

\textbf{Coherence Matrix: }\label{CM}
Given that each cell's next state is determined by its current state as well as its neighbors, we need to balance the importance of these two factors. On the one hand, if a superpixel is quite different from all its neighbors in the feature space, its next state will be primarily based on itself. On the other hand, if a cell is similar to its neighbors, it should be assimilated by the local environment. To this end, we build a coherence matrix $\textbf{C} = diag\{c_{1},c_{2},\cdots,c_{n_m}\}$ to promote the evolution among all cells. Each cell's coherence towards its current state is initially computed as: ${c_i} = \frac{1}{{\max ({f_{ij}})}}$, so it is inversely proportional to its maximum similarity to its neighbors. We normalize this to be in a range $c_{i} \in [~b~, a+b~]$, where $a$ and $b$ are parameters, via:
\vspace{-3mm}
\begin{equation}\label{c*}
c_i^ *  = a \cdot \frac{{{c_i} - \min \left( {{c_j}} \right)}}{{\max \left( {{c_j}} \right) - \min \left( {{c_j}} \right)}} + b,
\end{equation}
where the min and max are computed over $j = 1, 2,$ ..., $n_m$. Based on preliminary experiments, we set the constants $a$ and $b$ in Eq.~(\ref{c*}) to 0.6 and 0.2. If $a$ is fixed to 0.6, our results are insensitive to the value of $b$ in the interval $[~0.1~,0.3~]$. The final, normalized coherence matrix is then:  $\textbf{C}^{\ast} = diag \{c_{1}^{\ast}, c_{2}^{\ast}, \cdots, c_{n_m}^{\ast}\}$.

\textbf{Synchronous Update Rule: }\label{updating}
In Cellular Automata, all cells will simultaneously update their states according to the update rule, which is a key point in Cellular Automata, as it controls whether the ultimate evolving state is chaotic or stable~\citep{wolfram1983statistical}. Here, we define the synchronous update rule based on the impact factor matrix $\textbf{F}^ * \in \mathbb{R}^{n_m\times n_m}$ and coherence matrix $\textbf{C}^ * \in \mathbb{R}^{n_m\times n_m}$:
\vspace{-1mm}
\begin{equation}\label{rulefunc}
{\textbf{s}^{(t + 1)}} = {\textbf{C}^ * } {\textbf{s}^{(t)}} + \left( {\textbf{I} - {\textbf{C}^ * }} \right) {\textbf{F}^ * } {\textbf{s}^{(t)}},
\vspace{-1mm}
\end{equation}
where $\textbf{I}$ is the identity matrix of dimension ${n_m\times n_m}$ and $\textbf{s}^{(t)}\in \mathbb{R}^{n_m}$ denotes the saliency map at time $t$. When $t = 0$, $\textbf{s}^{(0)}$ is the prior map generated by the method introduced in Section.~\ref{initialmaps}. After $T_S$ time steps (a time step is defined as one update of all cells), the saliency map can be represented as $\textbf{s}^{(T_S)}$. It should be noted that the update rule is invariant over time; only the cells' states $\textbf{s}^{(t)}$ change over iterations.
\begin{figure}
  \includegraphics[width=8.4cm]{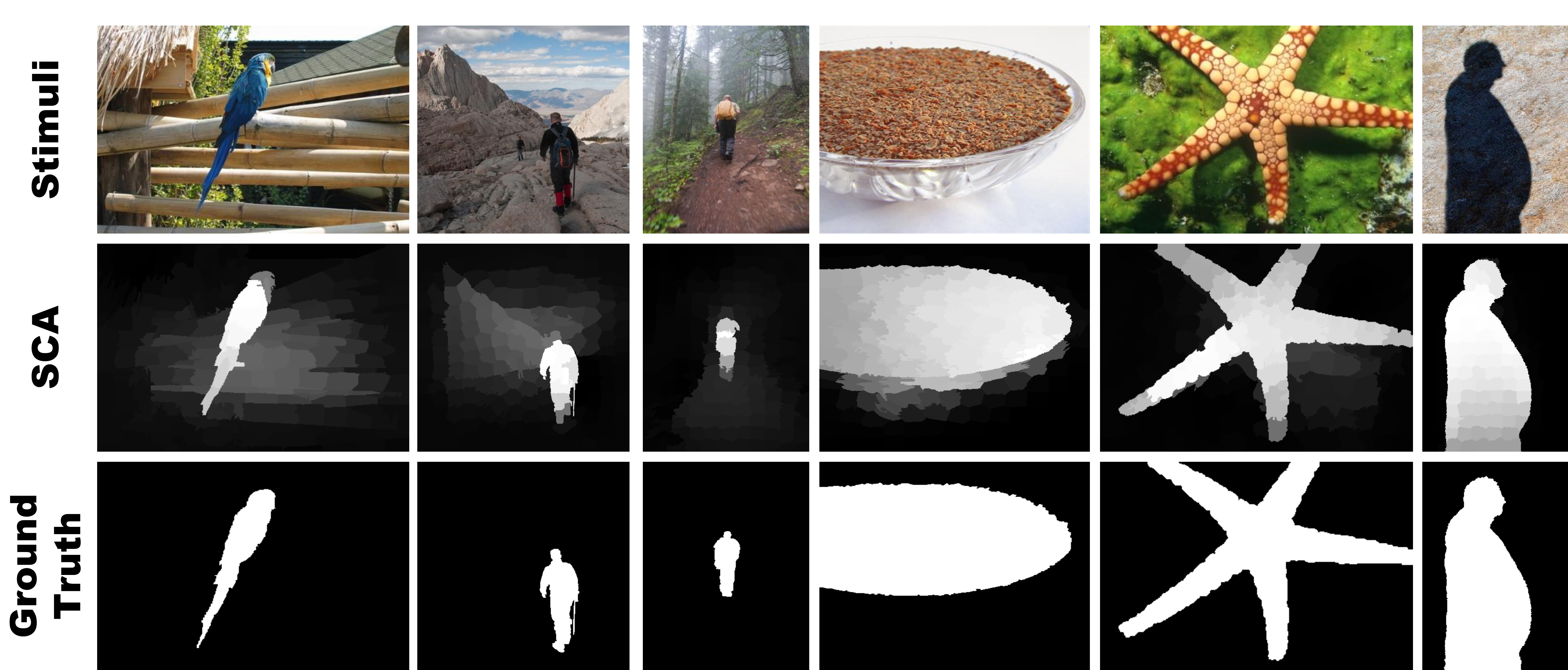}\\
    \vspace{-3mm}
    \caption{Saliency maps generated by SCA ($n_m =200$). The first three columns show that salient objects can be precisely detected when the saliency appears in the center of the image. The last three columns indicate that SCA can still have good performance even when salient objects touch the image boundary.}\label{sm-show1}
\vspace{-4mm}
\end{figure}

Our synchronous update rule is based on the generalized intrinsic characteristics of most images.
First, superpixels belonging to the foreground usually share similar feature representations. By exploiting the correlation between neighbors, the SCA can enhance saliency consistency among similar regions and develop a steady local environment.
Second, it can be observed that there is a high contrast between the object and its surrounding background in feature space. Therefore, a clear boundary will naturally emerge between the object and the background, as the cell's saliency value is greatly influenced by its similar neighbors. With boundary-based prior maps, salient objects can be naturally highlighted after the evolution of the system due to the connectivity and compactness of the object, as exemplified in Figure.~\ref{sm-show1}. Specifically, even though part of the salient object is incorrectly selected as the background seed, the SCA can adaptively increase their saliency values under the influence of the local environment. The last three columns in Figure.~\ref{sm-show1} show that when the object touches the image boundary, the results achieved by the SCA are still satisfying.

\vspace{-3mm}
\subsubsection{Cuboid Cellular Automata}\label{CCA}
\vspace{-2mm}
 To better capture the salient objects of different scales, we propose a novel method named Cuboid Cellular Automata (CCA) to incorporate $M$ different saliency maps generated by SCA under $M$ scales, each of which serves as a layer of the Cuboid Cellular Automata. In CCA, each cell corresponds to a pixel, and the saliency values of all pixels constitute the set of cells'
states. The number of all pixels in an image is denoted as \emph{H}. Unlike the definition of a neighborhood in Section.~\ref{SCA} and Multi-layer Cellular Automata in~\citep{Qin_2015_CVPR}, here pixels with the same coordinates in different saliency maps as well as their 4-connected pixels are all regarded as neighbors. That is, for any cell in a saliency map, it should have $5M-1$ neighbors, constructing a cuboid interaction zone. The hierarchical graph is presented in Figure.~\ref{graph} (b) to illustrate the connections between neighbors.

The saliency value of pixel $i$ in the $m$-th saliency map at time $t$ is its probability of being the foreground $F$, represented as $s_{m,i}^{(t)} = P(i \in_m^{(t)} F) $, while $1 - s_{m,i}^{(t)}$ is its probability of being the background $B$, denoted as $1 - s_{m,i}^{(t)} = P(i \in_m^{(t)} B)$. We binarize each map with an adaptive threshold using  Otsu's method~\citep{otsu1975threshold}, which is computed from the initial saliency map and does not change over time. The threshold of the $m$-th saliency map is denoted by $\gamma_{m}$. If pixel $i$ in the $m$-th binary map is classified as foreground at time $t$ ($s_{m,i}^{(t)} \geq \gamma_m)$, then it will be denoted as $\eta_{m,i}^{(t)}=+1$. Correspondingly, $\eta_{m,i}^{(t)}=-1$ means that pixel $i$ is binarized as background ($s_{m,i}^{(t)} < \gamma_m$).

If pixel $i$ belongs to the foreground, the probability that one of its neighboring pixels $j$ in the $m$-th binary map is classified as foreground at time $t$ is denoted as $P(~\eta_{m,j}^{(t)} = +1|i \in^{(t)}_m F~)$. In the same way, the probability $P(~\eta_{m,j}^{(t)} = -1|i \in_m^{(t)} B~)$ represents that the pixel $j$ is binarized as $B$ conditioned on that pixel $i$ belongs to the background at time $t$. We make the assumption that $P(~\eta_{m,j}^{(t)} = +1|i \in^{(t)}_m F~)$ is the same for all the pixels in any saliency map and it does not change over time. Additionally, it is reasonable to assume that $P(~\eta_{m,j}^{(t)} = +1|i \in^{(t)}_m F~) = P(~\eta_{m,j}^{(t)} = -1|i \in_m^{(t)} B~)$. Therefore, we use a constant $\lambda$ to denote these two probablities:
\begin{equation}
P(~\eta_{m,j}^{(t)} = +1|i \in^{(t)}_m F~) = P(~\eta_{m,j}^{(t)} = -1|i \in_m^{(t)} B~) = \lambda.
\end{equation}
Then the posterior probability $P(i \in_m^{(t)} F|\eta _{m,j}^{(t)} = +1)$ can be calculated as follows:
 \vspace{-1mm}
\begin{equation}\label{eqnp}
\begin{split}
&~P\left( {i \in^{(t)}_m F\left| {{\eta _{m,j}^{(t)}} =  + 1} \right.} \right)\\
 \propto &~P\left( {i \in^{(t)}_m F} \right)P\left( {{\eta _{m,j}^{(t)}} =  + 1\left| {i \in^{(t)}_m F} \right.} \right) \\
 =&~ {s_{m,i}^{(t)}} \cdot \lambda
\end{split}
\end{equation}

In order to get rid of the normalizing constant in Eqn.~(\ref{eqnp}), we define the prior ratio $\Omega(i \in^{(t)}_m F)$ as:
\vspace{-2mm}
\begin{equation}\label{eqnq}
\vspace{-2mm}
\Omega \left( {i \in^{(t)}_m F} \right) = \frac{{P\left( {i \in^{(t)}_m F} \right)}}{{P\left( {i \in^{(t)}_m B} \right)}} = \frac{{{s_{m,i}^{(t)}}}}{{1 - {s_{m,i}^{(t)}}}}.
\end{equation}
Combining Eqn.~(\ref{eqnp}) and Eqn.~(\ref{eqnq}), the posterior ratio $\Omega(i \in^{(t)}_m F~|~\eta_{m,j}^{(t)}=+1)$ turns into:
\begin{equation}\label{eqn10}
\vspace{-4mm}
\begin{split}
\Omega \left( {i \in^{(t)}_m F\left| {{\eta _{m,j}^{(t)}} =  + 1} \right.} \right) &= \frac{{P\left( {i \in^{(t)}_m F\left| {{\eta _{m,j}^{(t)}} =  + 1} \right.} \right)}}{{P\left( {i \in^{(t)}_m B\left| {{\eta _{m,j}^{(t)}} =  + 1} \right.} \right)}} \\
&= \frac{{{s_{m,i}^{(t)}}}}{{1 - {s_{m,i}^{(t)}}}} \cdot \frac{\lambda }{{1 - \lambda }}.
\end{split}
\end{equation}

As the posterior probability $P(i \in^{(t)}_m F|\eta _{m,j}^{(t)} = +1)$  represents the probability of pixel $i$ of being the foreground $F$ conditioned on that its neighboring pixel $j$ in the $m$-th saliency map is binarized as foreground at time t, $P(i \in^{(t)}_m F|\eta _{m,j}^{(t)} = +1)$ can also be used to represent the probability of pixel $i$ of being the foreground $F$ at time $t+1$. Then,
\begin{equation}\label{eqns}
s_{m,i}^{(t+1)} = P(i \in^{(t)}_m F|\eta _{m,j}^{(t)} = +1).
\end{equation}
According to Eqn.~(\ref{eqn10}) and Eqn.~(\ref{eqns}), we can get:
\begin{equation}\label{eqn11}
\vspace{-2mm}
\begin{split}
\frac{{s_{m,i}^{(t + 1)}}}{{1 - s_{m,i}^{(t + 1)}}} &=  \frac{P(i \in^{(t)}_m F|\eta _{m,j}^{(t)} = +1)}{1-P(i \in^{(t)}_m F|\eta _{m,j}^{(t)} = +1)}\\
&= \frac{P(i \in^{(t)}_m F|\eta _{m,j}^{(t)} = +1)}{P(i \in^{(t)}_m B|\eta _{m,j}^{(t)} = +1)}\\
& = \frac{{{s_{m,i}^{(t)}}}}{{1 - {s_{m,i}^{(t)}}}} \cdot \frac{\lambda }{{1 - \lambda }}.
\end{split}
\end{equation}
It is much easier to deal with the logarithm of this quantity  because the changes in logodds will be additive. So Eqn.~(\ref{eqn11}) turns into:
\begin{equation}\label{eqn12}
l\left( {s_{m,i}^{(t + 1)}} \right) = l\left( {s_{m,i}^{(t )}} \right) +\mathrm{\Lambda},
\end{equation}
where $l\left( {s_{m,i}^{(t + 1)}} \right)= \ln(\frac{{s_{m,i}^{(t + 1)}}}{{1 - s_{m,i}^{(t + 1)}}})$  and $\mathrm{\Lambda}=\ln(\frac{\lambda}{1-\lambda})$ is a constant. The intuitive explanation for Eqn.~(\ref{eqn12}) is that: if a pixel observes that one of its neighbors is binarized as foreground, it ought to increase its saliency value; otherwise, it should decrease its saliency value. Therefore, Eqn.~(\ref{eqn12}) requires $\mathrm{\Lambda} > 0$. In this paper, we empirically set $\mathrm{\Lambda} = 0.05$.

As each pixel has $5M-1$ neighbors in total, the pixel will decide its action (increase or decrease it saliency value) based on all its neighbors' current states. Assuming the contribution of each neighbor is conditionally independent, we derive the synchronous update rule from Eqn.~(\ref{eqn12}) as:
\vspace{-1mm}
\begin{equation}\label{mcaupdating_px}
l\left( {\textbf{s}_m^{(t + 1)}} \right) = l\left( {\textbf{s}_m^{(t)}} \right) +\boldsymbol{\mathrm{\Sigma}}_m^{(t)} \cdot \mathrm{\Lambda},
\end{equation}
where $\textbf{s}^{(t)}_m \in \mathbb{R}^{H}$ is the $m$-th saliency map at time $t$ and $H$ is the number of pixels in the image. $\boldsymbol{\mathrm{\Sigma}}_m^{(t)} \in \mathbb{R}^{H}$ can be computed by:
\vspace{-3mm}
\begin{equation}\label{mcaupdating}
\boldsymbol{\mathrm{\Sigma}}_m^{(t)}=\sum\limits_{j=1}^{5}\sum\limits_{k=1}^{M} \delta(k=m,j>1) \cdot \text{sign} \left( \textbf{s}_{j,k}^{(t)} -\gamma _k \cdot \textbf{1} \right),
\end{equation}
where $M$ is the number of different saliency maps, $\textbf{s}_{j,k}^{(t)} \in \mathbb{R}^{H}$ is a vector containing the saliency values of the j-th neighbor for all the pixels in the $m$-th saliency map at time $t$ and $\textbf{1} = [1,~1, \cdots, ~1]^\top \in \mathbb{R}^H$. We use $\delta(k=m,j>1)$ to represent the occasion that the cell only has 4 neighbors instead of 5 in the $m$-th saliency map when it is in the $m$-th saliency map. After $T_C$ iterations, the final integrated saliency map $\textbf{s}^{(T_C)}$ is calculated by
\vspace{-4mm}
\begin{equation}
{\textbf{s}^{{(T_C)}}} = \frac{1}{M}\sum\limits_{m = 1}^M {\textbf{s}_m^{{(T_C)}}}.
\end{equation}
\vspace{-4mm}
\begin{figure}
\centering
\includegraphics[width=8.4cm]{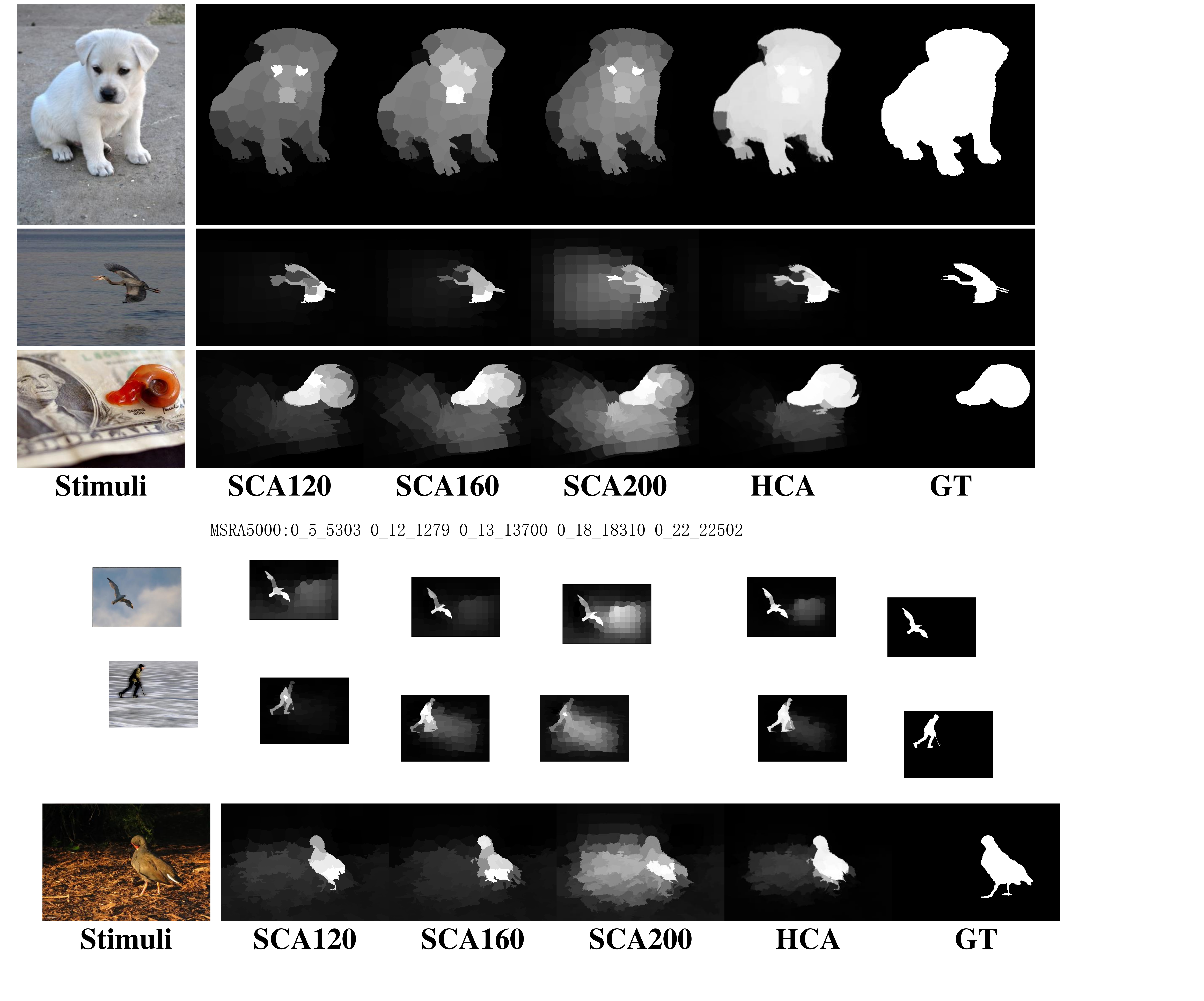}\\
\vspace{1mm}
\caption{Visual comparison of saliency maps generated by SCA at different scales ($n_1$ = 120, $n_2 = $ 160 and $n_3$ = 200) and HCA.}\label{hcas}
\end{figure}

In this paper, we use CCA to integrate saliency maps generated by SCA at $M=3$ scales. This combination is denoted as HCA, and the visual saliency maps generated by HCA can be seen in Figure.~\ref{hcas}. Here we use the notation SCA$n$ to denote SCA applied with $n$ superpixels. We can see that the detected objects in the integrated saliency maps are uniformly highlighted and much closer to the ground truth.

\vspace{-4mm}
\subsection{Consistent Optimization}\label{consist}
\vspace{-2mm}
\subsubsection{Single-layer Propagation}\label{sop}
\begin{figure}
  \center
  \includegraphics[width=8.4cm]{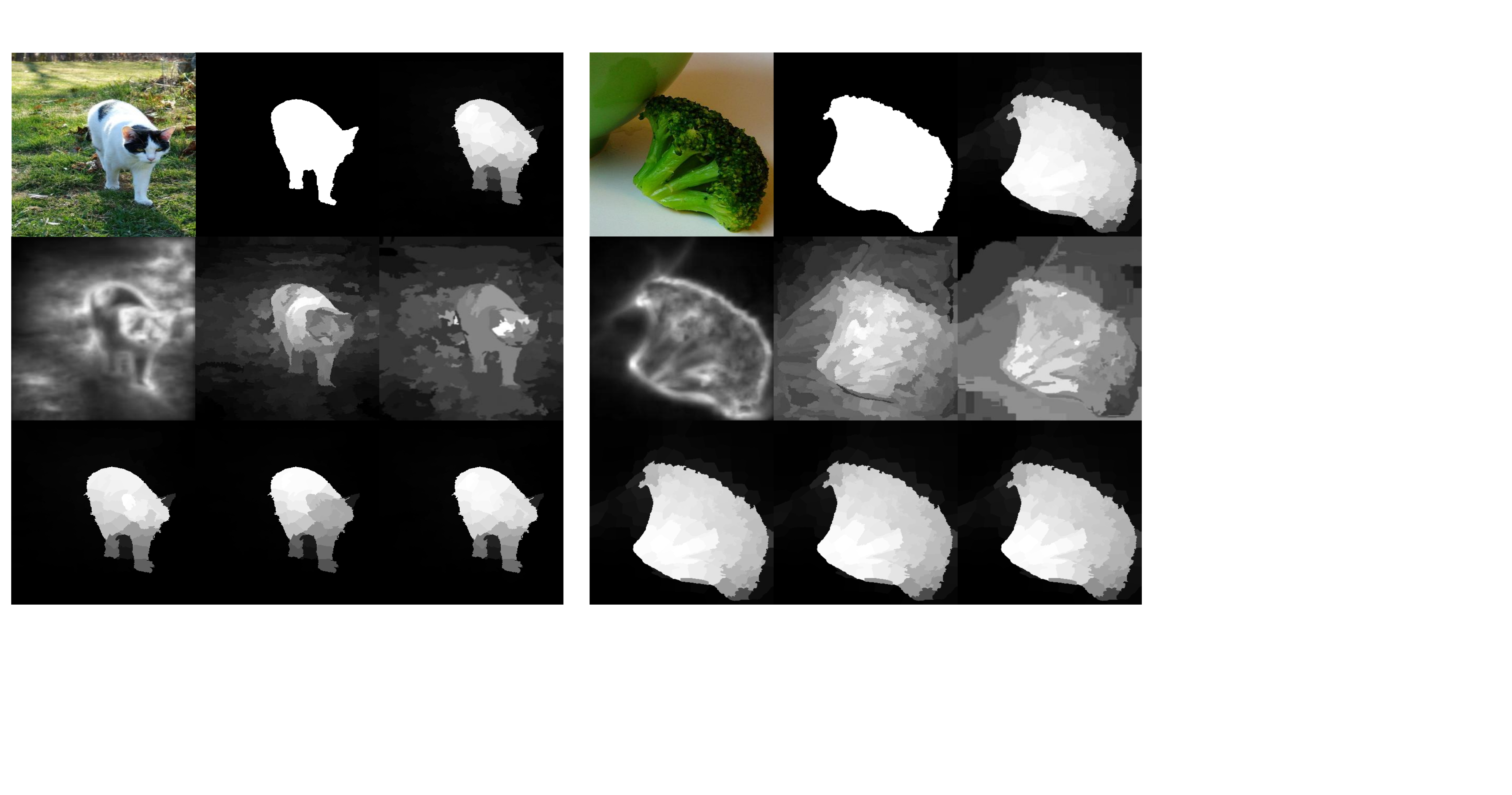}\\
  \caption{Comparison of saliency maps generated by different methods and their optimized results via Single-layer Cellular Automata. The first row is respectively input images, ground truth and saliency maps generated by our proposed SCA with 200 superpixels. The second row displays original saliency maps generated by three traditional methods (from left to right: CAS~\citep{goferman2010context}, LR~
  \citep{shen2012unified}, RC~\citep{cheng2015global}). The third row is their corresponding optimized results by SCA with 200 superpixels.}\label{sca-effect-visual}
  \vspace{-3mm}
\end{figure}

\vspace{-2mm}
Due to the connectivity and compactness of the object,
the salient part of an image will naturally emerge with the Single-layer Cellular Automaton, which serves as a propagation mechanism. Therefore, we use the saliency maps generated by several well-known methods as the prior maps and refresh them
according to the synchronous update rule. The saliency
maps achieved by CAS~\citep{goferman2010context}, LR~\citep{shen2012unified} and RC~\citep{cheng2015global} are taken as $\textbf{s}^{(0)}$ in Eqn.~(\ref{rulefunc}). The optimized results via SCA are shown in Figure.~\ref{sca-effect-visual}. We can see that the foreground is uniformly highlighted and a clear object contour naturally emerges with the automatic single-layer propagation mechanism.
Even though the original saliency maps are not particularly good, all of them are
significantly improved to a similar accuracy level after evolution. That means our method is independent of prior maps and
can make a consistent and efficient optimization towards state-of-the-art
methods.
\vspace{-6mm}
 \subsubsection{Pixel-wise Integration}\label{pwa}
\begin{figure}
  \center
  \subfigure[Saliency aggregation of three conventional methods]{\includegraphics[width=8.4cm]{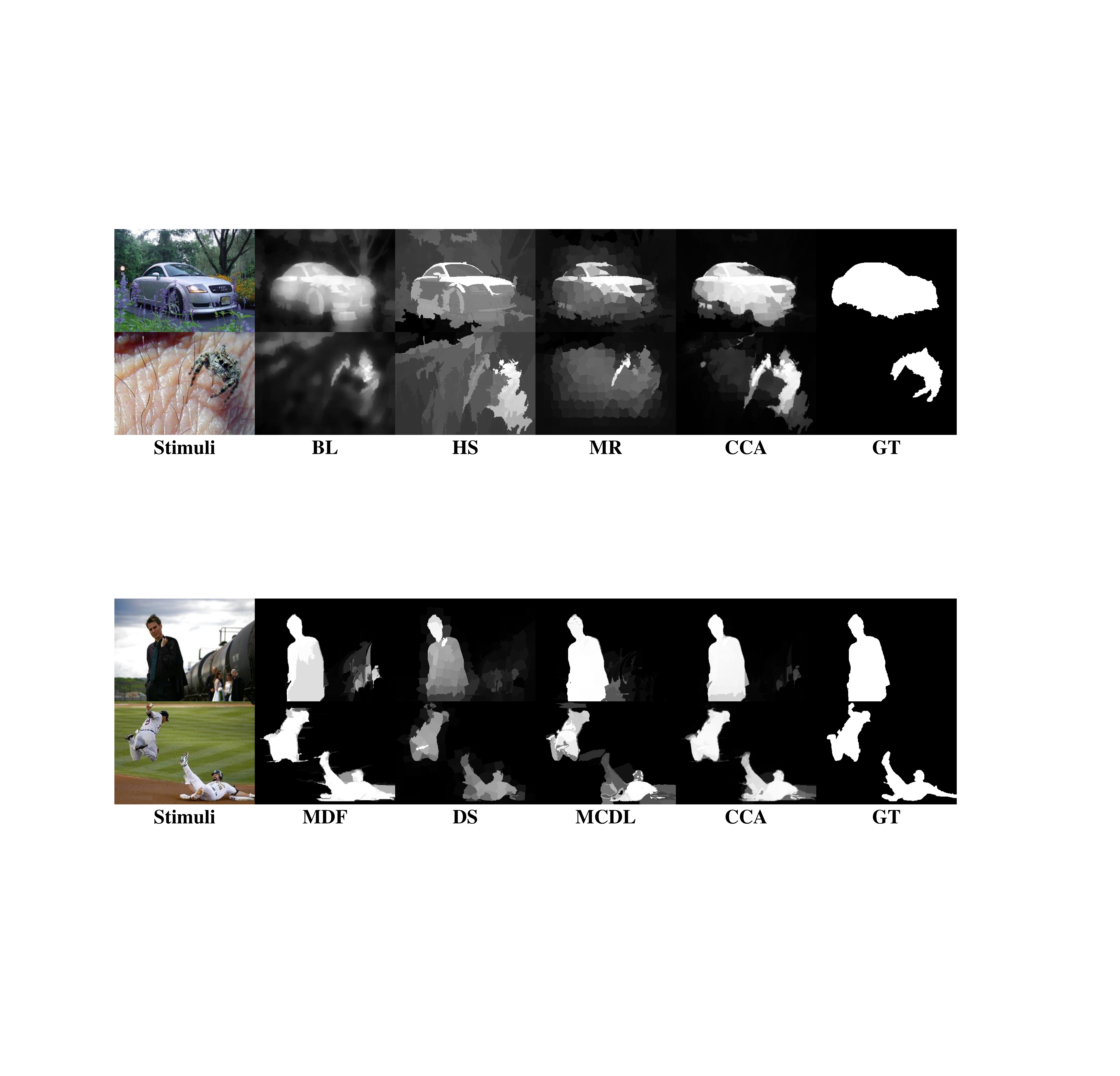}}\\
  \vspace{-3mm}
  \subfigure[Saliency aggregation of three deep learning methods]{\includegraphics[width=8.4cm]{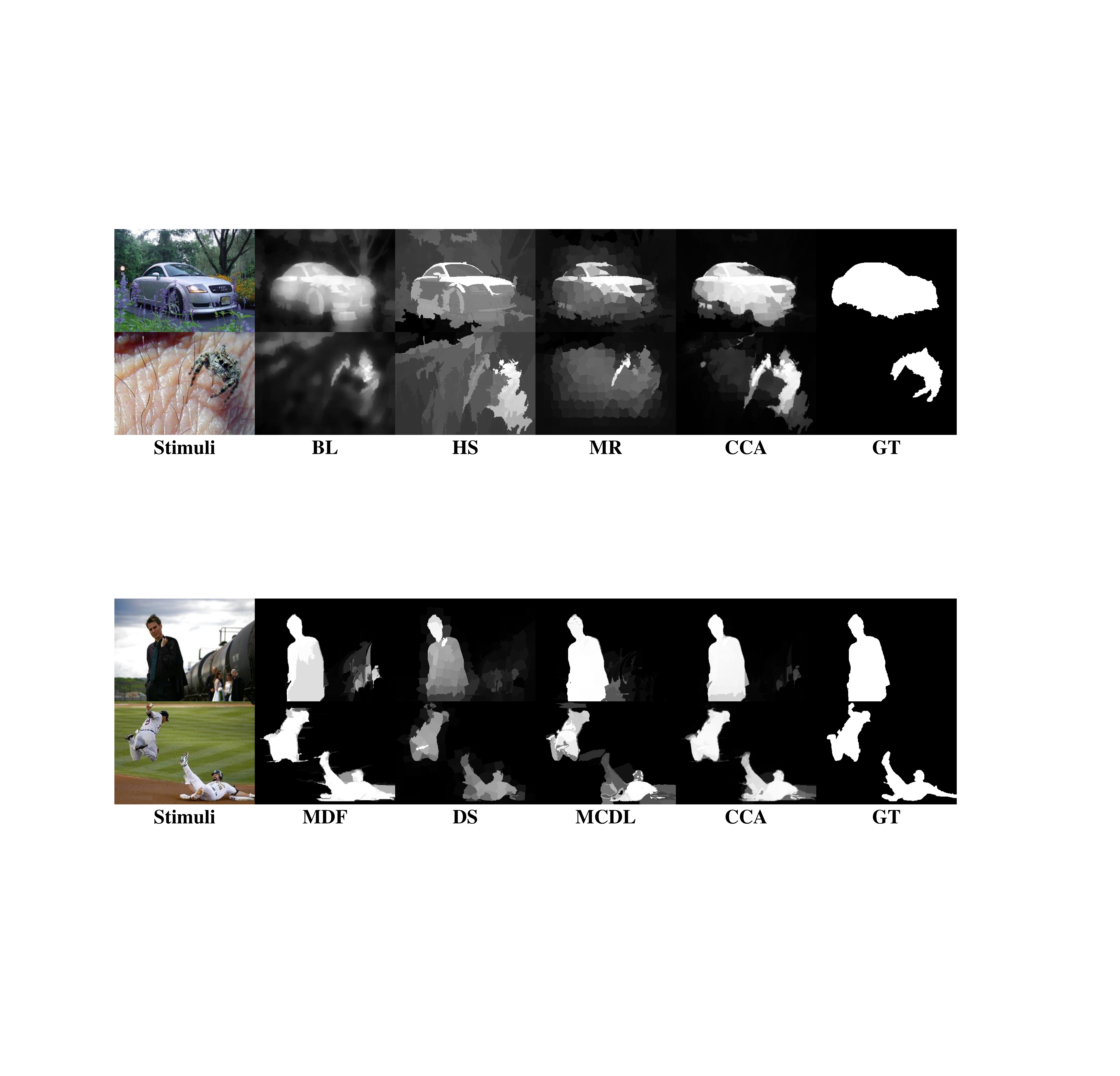}}
  \caption{Effects of pixel-wise saliency aggregation with Cuboid Cellular Automata. We integrate saliency maps generated by three conventional algorithms: BL~\citep{tong2015bootstrap}, HS~\citep{yan2013hierarchical} and MR~\citep{yang2013saliency} in (a) and incorporate saliency maps generated by three deep learning methods: MDF~\citep{li2015visual}, DS~\citep{Li2016DeepSaliency}, and MCDL~\citep{zhao2015saliency} in (b). The integrated result is denoted as CCA.}\label{mca-effect-visual}
\end{figure}

 \vspace{-2mm}
A variety of methods have been developed for visual saliency detection, and each of them has its advantages and limitations. As shown in Figure.~\ref{mca-effect-visual}, the performance of a saliency detection method varies with individual images. Each method can work well for some images or some parts of the images but none of them can perfectly handle all the images. Furthermore, different methods may complement each other. To take advantage of the superiority of each saliency map, we use Cuboid Cellular Automata to aggregate two groups of saliency maps, which are generated by three conventional algorithms: BL~\citep{tong2015bootstrap}, HS~\citep{yan2013hierarchical} and MR~\citep{yang2013saliency} and three deep learning methods: MDF~\citep{li2015visual} and DS~\citep{Li2016DeepSaliency}  and MCDL~\citep{zhao2015saliency}. Each of them serves as a layer of Cellular Automata $\textbf{s}_m^{(0)}$ in Eqn.~(\ref{mcaupdating_px}). Figure.~\ref{mca-effect-visual} shows that our proposed pixel-wise aggregation method, Cuboid Cellular Automata, can appropriately integrate multiple saliency maps and outperforms each one. The saliency objects on the aggregated saliency map are consistently highlighted and much closer to the ground truth.
\vspace{-3mm}
\subsubsection{SCA + CCA = HCA}
\begin{figure}
\center
  \includegraphics[width=8.4cm]{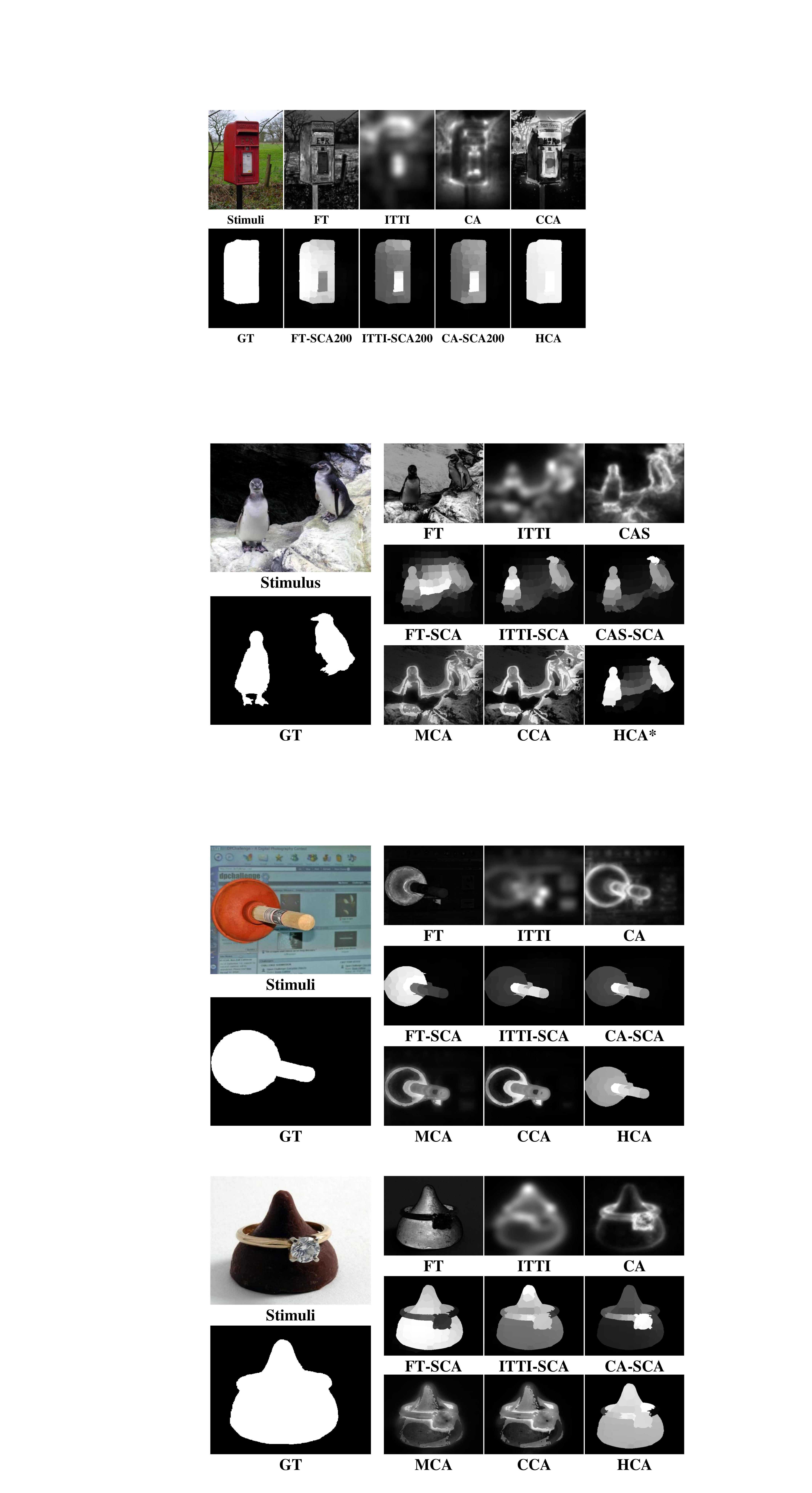}\\
  \caption{Effects of holistic optimization by Hierarchical Cellular Automata. We use MCA~\citep{Qin_2015_CVPR}, CCA and HCA to integrate saliency maps generated by three classic methods: FT~\citep{achanta2009frequency}, ITTI~\citep{itti1998model} and CAS~\citep{goferman2010context}. Their respective saliency maps optimized by SCA with 200 superpixels are shown in the second row. Note that HCA* uses as input the saliency maps processed by SCA (the second row) and applies CCA to them, while the MCA and CCA models are applied directly to the first row.}\label{cca-hca-visual-cmp}
  \vspace{-3mm}
\end{figure}

\vspace{-3mm}
Here we show that when CCA is applied to some (poor) prior maps, it does not perform as well as when the prior map is post-processed by SCA. This motivates their combination into HCA. As is shown in Figure.~\ref{cca-hca-visual-cmp}, when the candidate saliency maps are not well constructed, both CCA and MCA~\citep{Qin_2015_CVPR} fail to detect the salient object. Unlike CCA and MCA, HCA overcomes this limitation through incorporating single-layer propagation (SCA) together with pixel-wise integration (CCA) into a unified framework. The salient objects can be intelligently detected by HCA regardless of the original performance of the candidate methods. When we use HCA to integrate existing methods, the optimized results will be denoted as HCA*.

\vspace{-7mm}
\section{Experiments}
\vspace{-3mm}
In order to demonstrate the effectiveness of our proposed algorithms, we compare the results on four challenging datasets: ECSSD~\citep{yan2013hierarchical}, MSRA5000 \citep{liu2011learning}, PASCAL-S~\citep{li2014secrets} and HKU-IS~\citep{li2015visual}. The Extended Complex Scene Saliency Dataset (ECSSD) contains 1000 images with multiple objects of different sizes. Some of the images come from the challenging Berkeley-300 dataset. MSRA-
5000 contains more comprehensive images with complex background. The PASCAL-S dataset derives from the validation set of PASCAL VOC2010~\citep{everingham2010pascal} segmentation challenge and contains 850 natural images. The last dataset, HKU-IS, contains 4447 challenging images and their pixel-wise saliency annotation. In this paper, we use ECSSD as the validation dataset to help choose the feature maps in FCN~\citep{long2015fully}.

We compare our algorithm with 20 classic or
state-of-the-art methods including ITTI~\citep{itti1998model}, FT~\citep{achanta2009frequency}, CAS~\citep{goferman2010context}, LR~\citep{shen2012unified},
 XL13~\citep{xie2013bayesian},
DSR\\\citep{li2013saliency}, HS~\citep{yan2013hierarchical}, UFO~\citep{jiang2013salientufo}, MR~\citep{yang2013saliency},
DRFI~\citep{jiang2013salient}, wCO~\citep{zhu2014saliency}, RC~\citep{cheng2015global}, HDCT~\citep{kim2014salient}, BL~\citep{tong2015bootstrap}, BSCA~\citep{Qin_2015_CVPR}, LEGS~\citep{wang2015deep}, MCDL~\citep{zhao2015saliency}, MDF~\citep{li2015visual}, DS~\citep{Li2016DeepSaliency}, and SSD-HS~\citep{ssd2016eccv}, where the last 5 methods are deep learning-based methods. The results of different methods are either provided by authors or achieved by running available code or binaries. The code and results of HCA will be publicly available at our project site \footnote{ \url{https://github.com/ArcherFMY/HCA_saliency_codes}}.
\vspace{-6mm}
\subsection{Parameter Setup}
\vspace{-2mm}
For the Single-layer Cellular Automaton, we set the number of iterations $T_S = 20$. $\sigma_f^{2}$ in Eq.~(\ref{impactfactor}) is set to 0.1 as in~\citep{yang2013saliency}. For the Cuboid Cellular Automata, we set the number of iterations $T_C = 3$. We determined empirically that SCA and CCA converge by 20 and 3 iterations, respectively. We choose $M=3$ and run SCA with $n_1 = 120, n_2 = 160, n_3 = 200$ superpixels to generate multi-scale saliency maps for CCA.

\begin{figure*}
  \includegraphics[width=17.4cm]{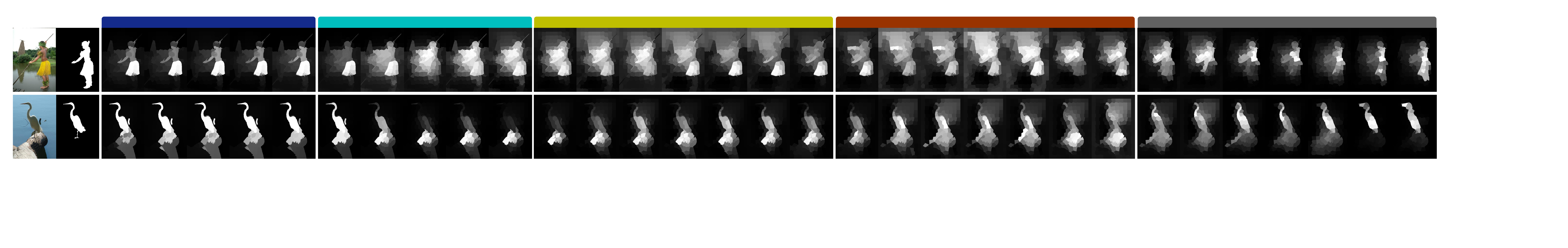}\\
  \vspace{-3mm}
  \caption{Visual comparison of saliency maps with different layers of deep features. The left two columns are the input images and their ground truth. Other columns present the saliency maps with different layers of deep features. The color bars on the top stand for different convolutional layers (see Figure.~\ref{dnn-layers}(a) and (b)).}\label{visual-dnn-layers}
\vspace{-5mm}
\end{figure*}

\vspace{-5mm}
\begin{figure}
  \subfigure[F-measure bars]{\includegraphics[width=3.84cm]{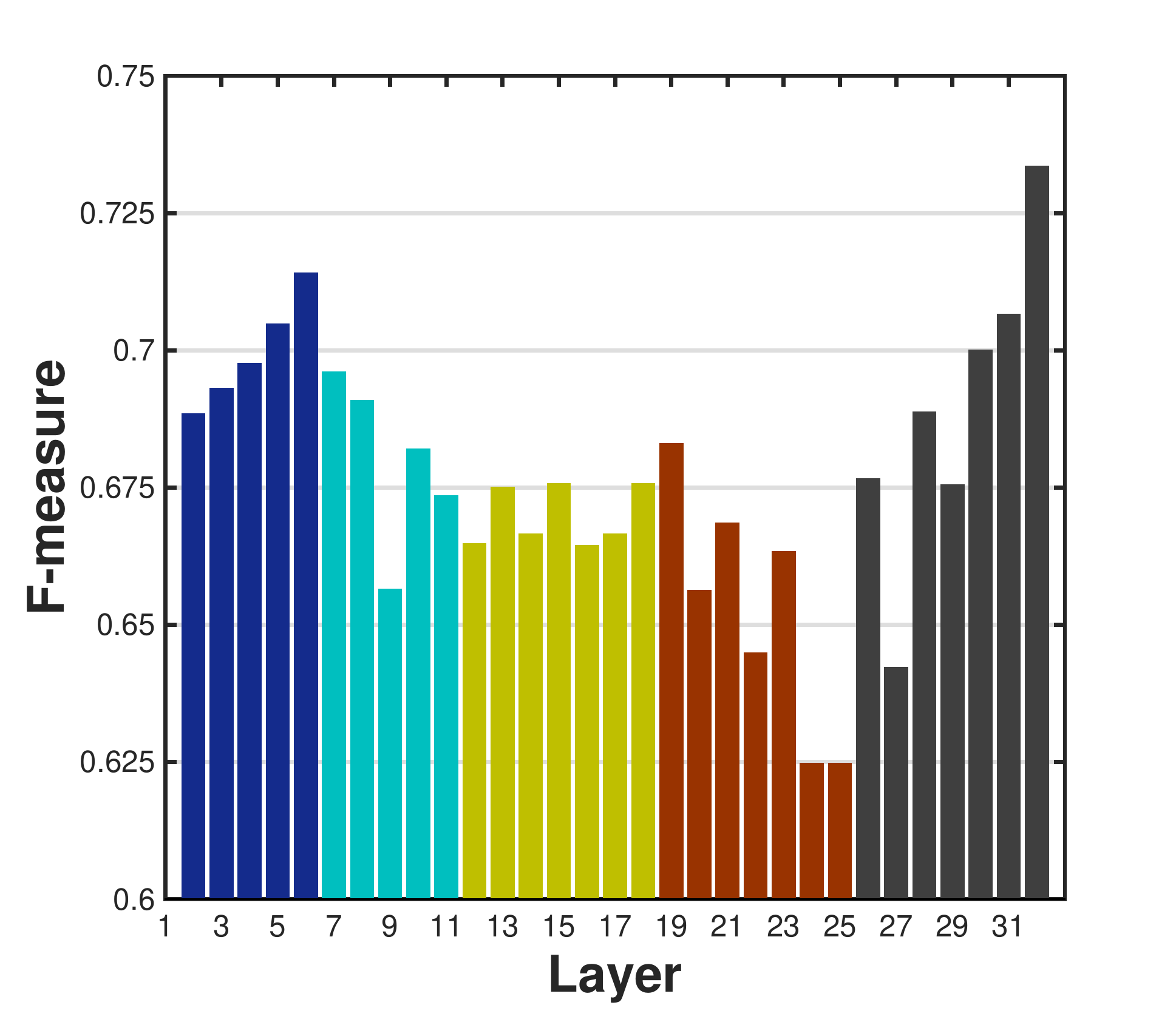}}
  \includegraphics[width=.57cm]{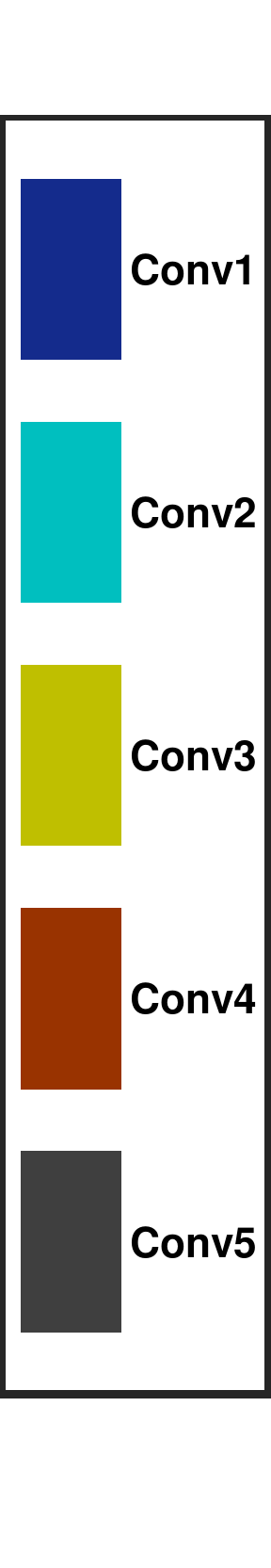}
  \subfigure[MAE bars]{\includegraphics[width=3.84cm]{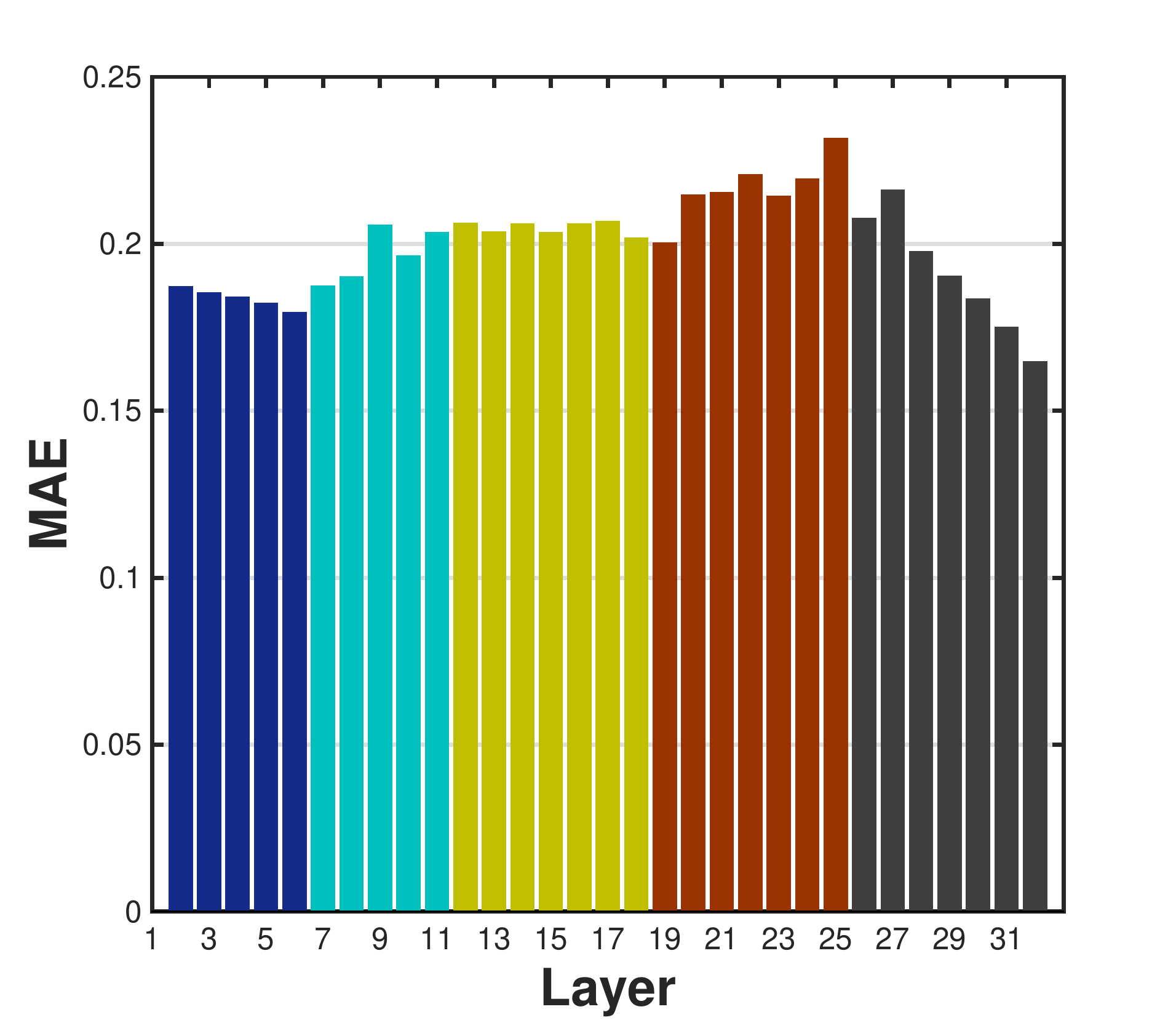}}\\
  \subfigure[Precision-Recall curves comparison]{\includegraphics[width=3.8cm]{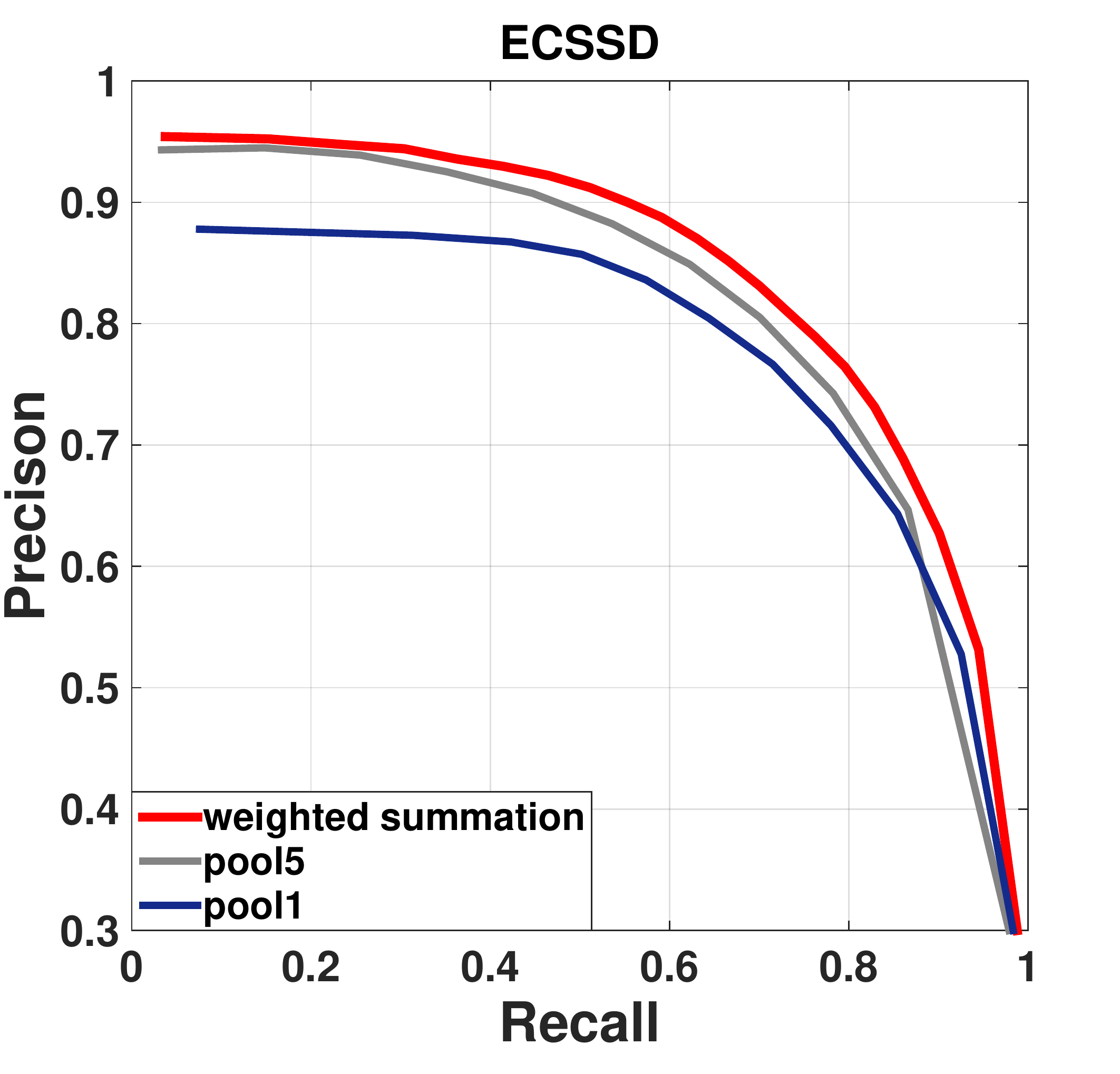}
  \includegraphics[width=.57cm]{./legend1.pdf}
  \includegraphics[width=3.8cm]{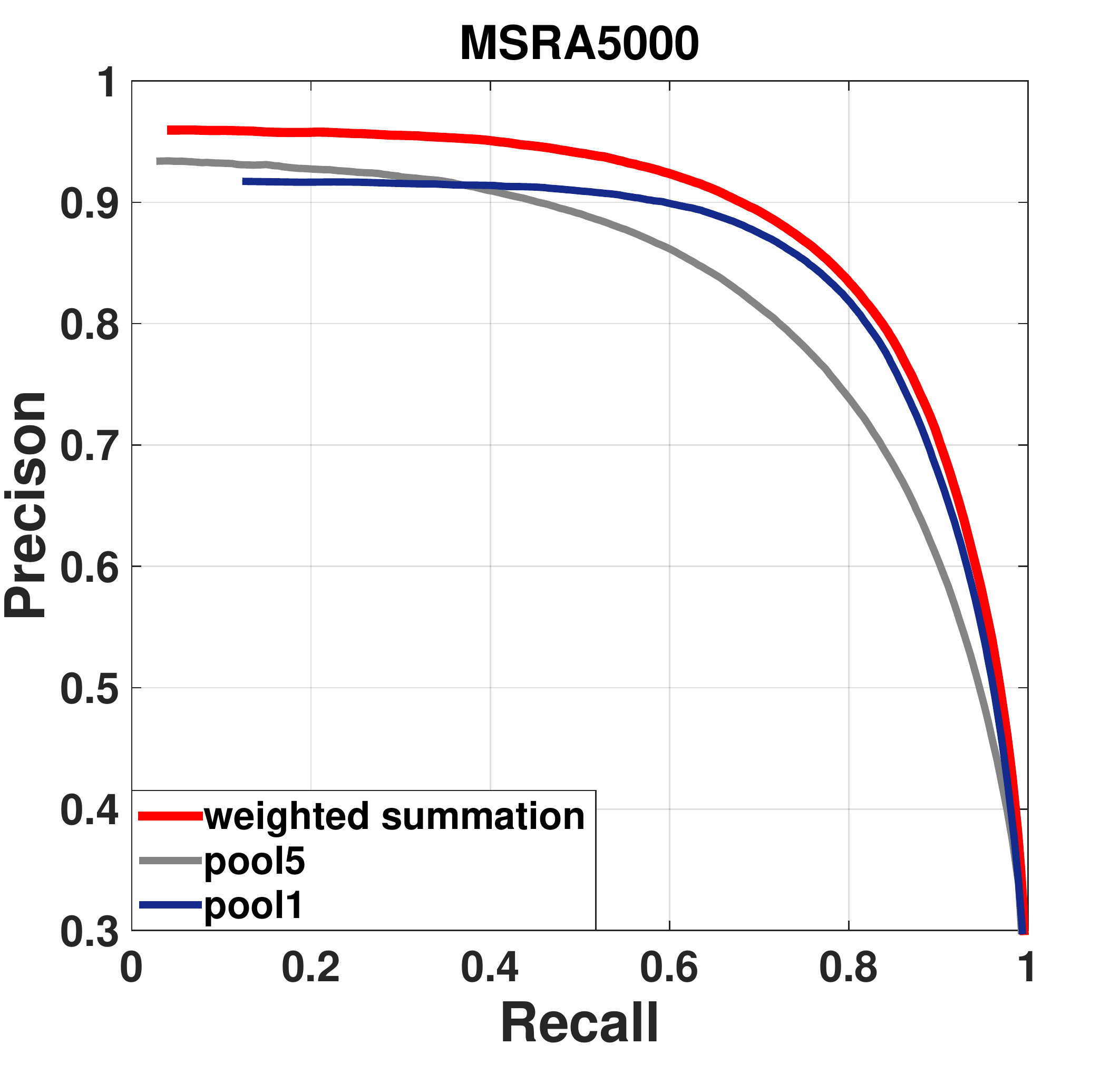}}
  \caption{(a) The F-measure score for each layer in FCN-32s on ECSSD; (b) The MAE score for each layer in FCN-32s on ECSSD; (c) Precision-Recall curves of SCA using deep features extracted from \texttt{pool1} and \texttt{pool5} as well as a weighted summation of these two layers of deep features.}\label{dnn-layers}
    \vspace{-4mm}
\end{figure}
\vspace{-3mm}
\subsection{Evaluation Metrics}\label{PEM}
\vspace{-2mm}
We evaluate all methods by standard Precision-Recall (PR) curves via binarizing the saliency map with a threshold sliding from 0 to 255 and then comparing the binary maps with the ground truth. Specifically,
\begin{equation}
\text{precision} = \frac{|SF \cap GF|}{|SF|}, \text{recall} = \frac{|SF \cap GF|}{|GF|},
\end{equation}
where $SF$ is the set of the pixels segmented as the foreground, $GF$ denotes the set of the pixels belonging to the foreground in the ground truth, and $|\cdot|$ refers to the number of elements in a set.
In many cases, high precision and recall are both required. These are combined in the F-measure to obtain a single figure of merit, parameterized by $\beta$:
\vspace{-2mm}
\begin{equation}
{F_\beta } = \frac{{\left( {1 + {\beta ^2}} \right) \cdot \text{precision} \cdot \text{recall}}}{{{\beta ^2} \cdot \text{precision} + \text{recall}}}
\vspace{-2mm}
\end{equation}
where $\beta^2$ is set to 0.3 as suggested in~\citep{achanta2009frequency} to emphasize the precision. To complement these two measures, we also use mean absolute error (MAE) to quantitatively measure the average difference between the saliency map $\mathbf{s} \in \mathbb{R}^H$ and the ground truth $\textbf{g} \in \mathbb{R}^H$ in pixel level:
\vspace{-2mm}
\begin{equation}
\text{MAE} = \frac{1}{H}\sum\limits_{i = 1}^H {\left| {s_i - g_i} \right|}.
\vspace{-2mm}
\end{equation}
MAE indicates how similar a saliency map is compared to the ground truth, and is of great importance for different applications, such as image segmentation and cropping~\citep{perazzi2012saliency}.
\vspace{-5mm}
\subsection{Validation of the Proposed Algorithm}
\vspace{-2mm}
\subsubsection{Feature Analysis}
\vspace{-2mm}
In order to construct the Impact Factor matrix, we need to choose the features that will enter into Eqn.(~\ref{hh}). Here we analyze the efficacy of the features in different layers of a deep network in order to choose these feature layers. In deep neural networks, earlier convolutional layers capture fine-grained low-level information, e.g.,~colors, edges and texture, while later layers capture high-level semantic features. In order to select the best feature layers in the FCN~\citep{long2015fully}, we use ECSSD as a validation dataset to measure the performance of deep features extracted from different layers. The function $g(\textbf{r}_i, \textbf{r}_j)$ in Eqn.~(\ref{impactfactor}) can be computed as
\vspace{-2mm}
\begin{equation}
g(\textbf{r}_i, \textbf{r}_j) = \left\| {df_i^l - df_j^l} \right\|_2,
\vspace{-2mm}
\end{equation}
where $df_i^l$ denotes the deep features of superpixel $i$ on the $l$-th layer. The outputs of convolutional layers, relu layers and pooling layers are all regarded as a feature map. Therefore, we consider in total 31 layers of fully convolutional networks. We do not take the last two convolutional layers into consideration as their spatial resolutions are too low.

We use the F-measure (the higher, the better) and mean absolute error (MAE) (the lower, the better) to evaluate the performance of different layers on the ECSSD dataset.  The results are shown in Figure.~\ref{dnn-layers}~(a) and (b). The F-measure score is obtained by thresholding the saliency maps at twice the mean saliency value.  We use this convention for all of the subsequent F-measure results. The x-index in Figure.~\ref{dnn-layers} (a) and (b) refers to convolutional, ReLu, and pooling layers as implemented in the FCN. We can see that deep features extracted from the pooling layer in \texttt{Conv1} and \texttt{Conv5} can achieve the best two F-measure scores, and also perform well on MAE. The saliency maps in Figure.~\ref{visual-dnn-layers} correspond to the bars in Figure~\ref{dnn-layers}. Here it is visually apparent that salient objects are better detected with the final pooling layers of \texttt{Conv1} and \texttt{Conv5} . Therefore, in this paper, we combine the feature maps from \texttt{pool1} and \texttt{pool5} with a simple linear combination. Eqn. (\ref{hh}) then turns into:
 \vspace{-1mm}
\begin{equation}\label{2layers}
g(\textbf{r}_i,\textbf{r}_j)={{\rho_1} \cdot \left\| {df^{5}_i-df^{5}_j} \right\|_2 + {\rho _2} \cdot \left\| {df^{31}_i-df^{31}_j} \right\|_2},
\end{equation}
where $\rho_1$ and $\rho_2$ balance the weight of \texttt{pool1} and \texttt{pool5}. In this paper, we empirically set $\rho_1=0.375$ and $\rho_2=0.625$ and apply them to all other datasets.

To test the effectiveness of the integrated deep features, we show the Precision-Recall curves of Single-layer Cellular Automata with each layer of deep features as well as the integrated deep features on two datasets. The Precision-Recall curves in Figure.~\ref{dnn-layers}~(c) demonstrate that hierarchical deep features outperform single-layer features, as they contain both category-level semantics and fine-grained details.
  \vspace{-4mm}
\subsubsection{Component Effectiveness}
\begin{figure}
\centering
\includegraphics[width=4.15cm]{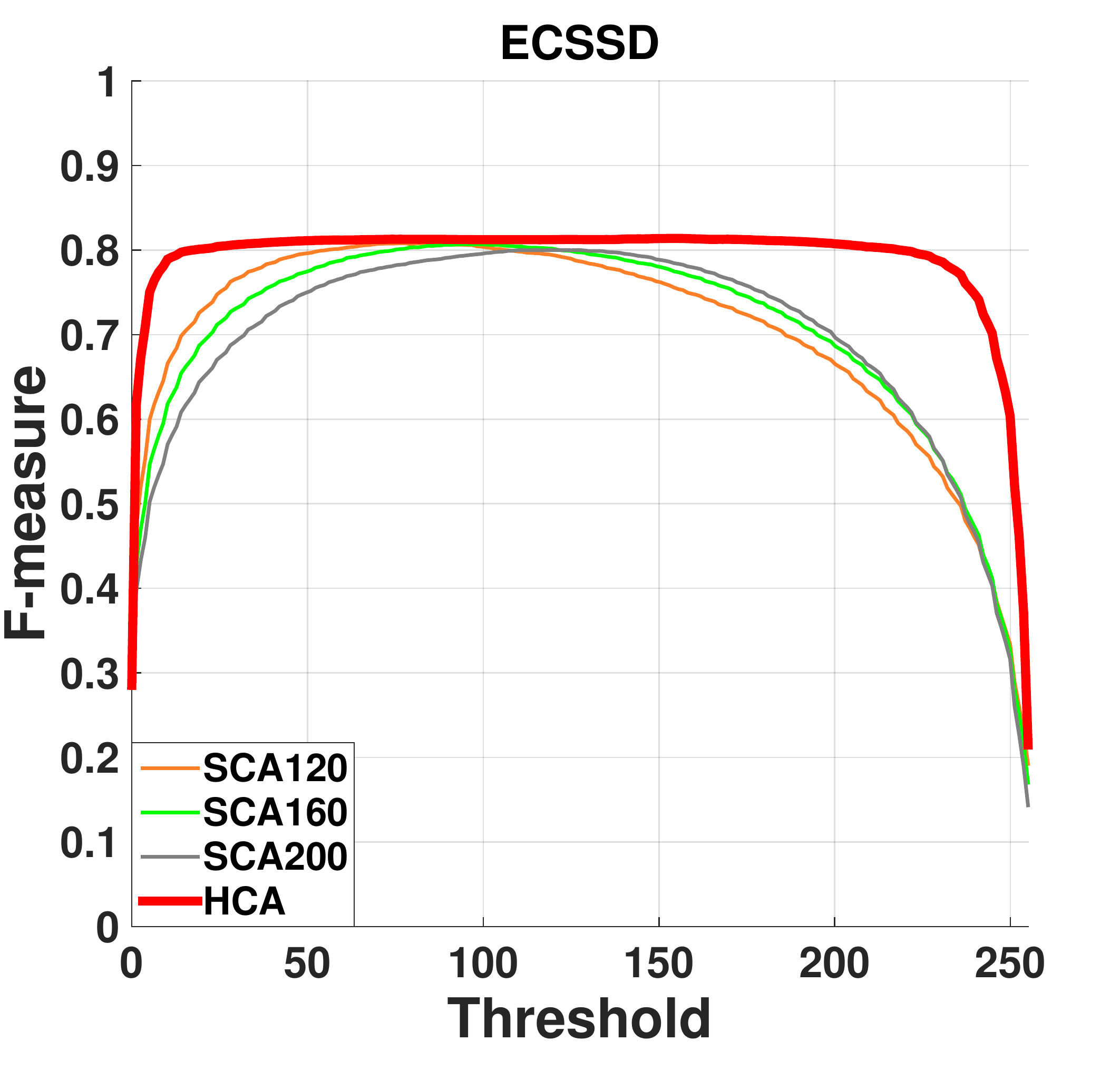}\label{mca-com-fmeasure}
\includegraphics[width=4.15cm]{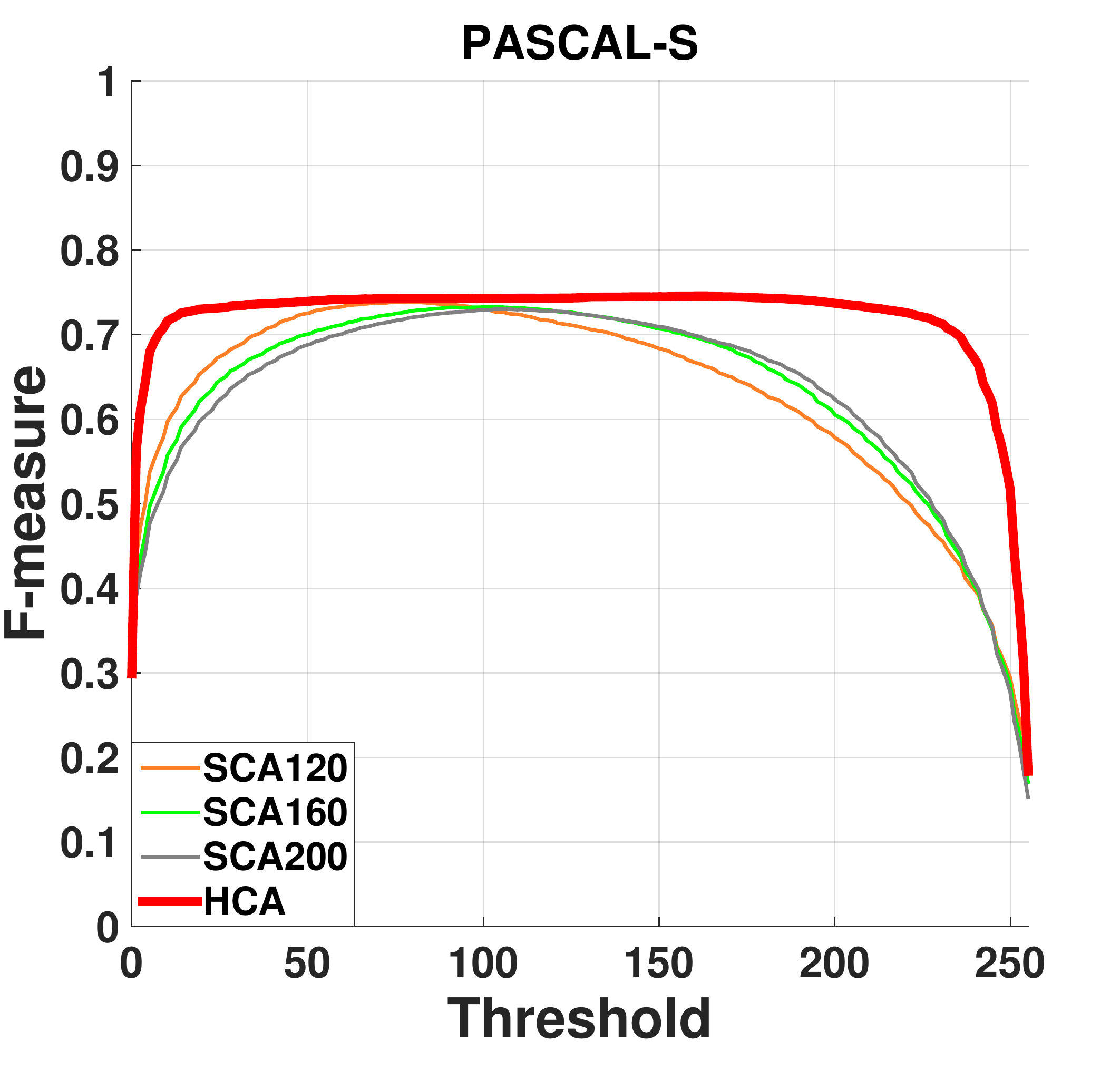}\label{mca-com-fmeasure2}\\

  \caption{F-measure - Threshold curves of saliency maps generated by SCA at different scales ($n_1=120, n_2=160, n_3 = 200$ respectively), and the integrated results by HCA on ECSSD and PASCAL-S. }\label{mcacom}
    \vspace{-4mm}
\end{figure}

\vspace{-2mm}
To demonstrate the effectiveness of our proposed algorithm, we test the results on the standard ECSSD and PASCAL-S datasets. We generate saliency maps at three scales: $n_1 = 120, n_2 = 160, n_3 = 200$ and use CCA to integrate them. FT curves in Figure.~\ref{mcacom} indicate that the results of the Single-layer Automata are already quite satisfying. In addition, CCA can improve the overall performance of SCA with a wider range of high F-measure scores than SCA alone.
 Similar results are also achieved on other datasets but are not presented here to be succinct.
\vspace{-4mm}
\subsubsection{Performance Comparison}
\begin{figure*}
  \includegraphics[width=17.5cm]{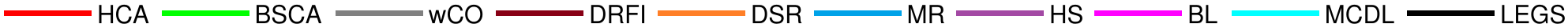}\\
  \includegraphics[width=0.55cm]{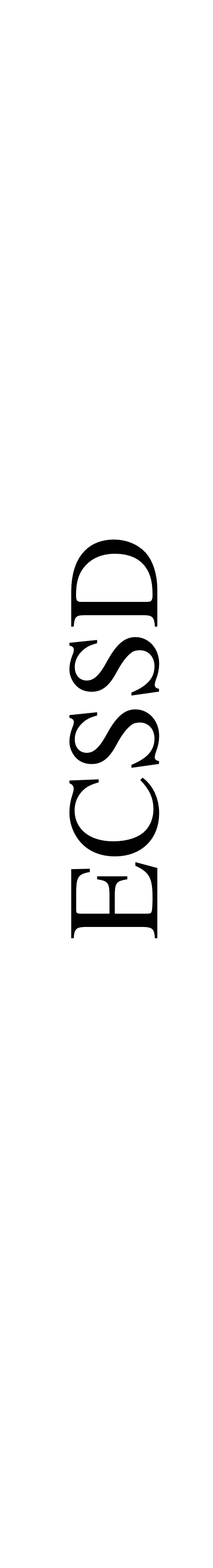}
  \includegraphics[width=5.7cm]{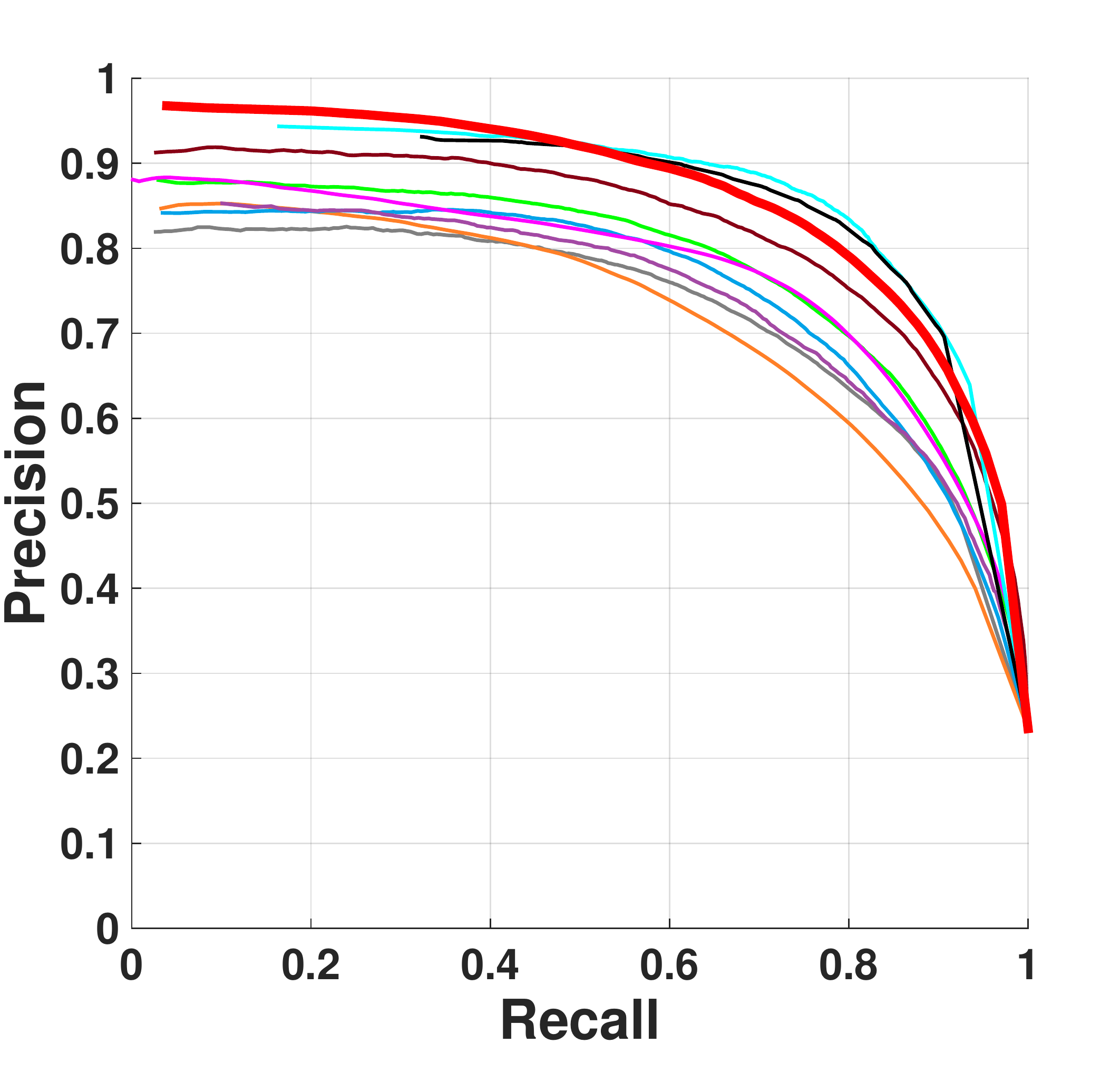}
  \includegraphics[width=5.7cm]{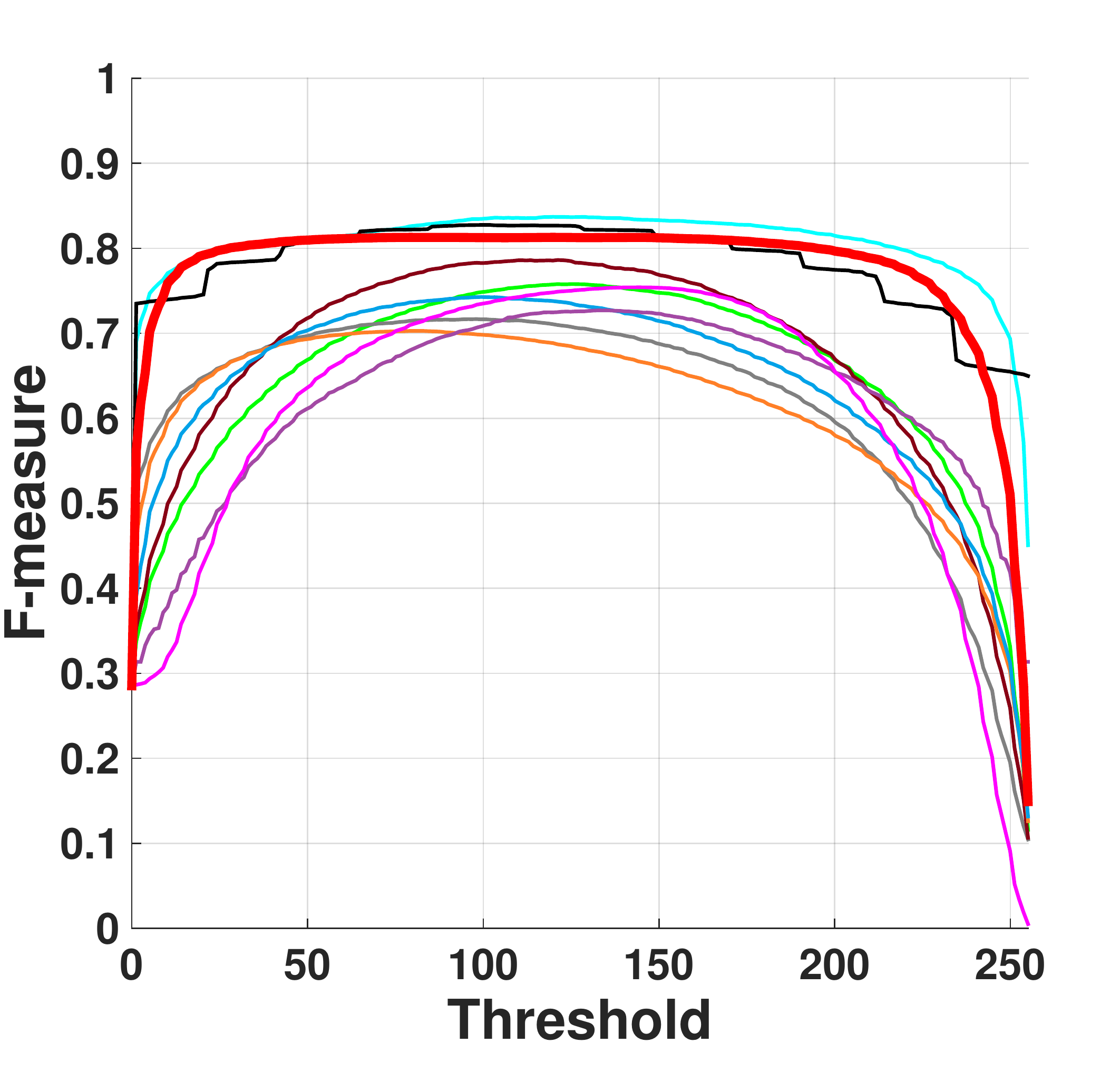}
  \includegraphics[width=5.7cm]{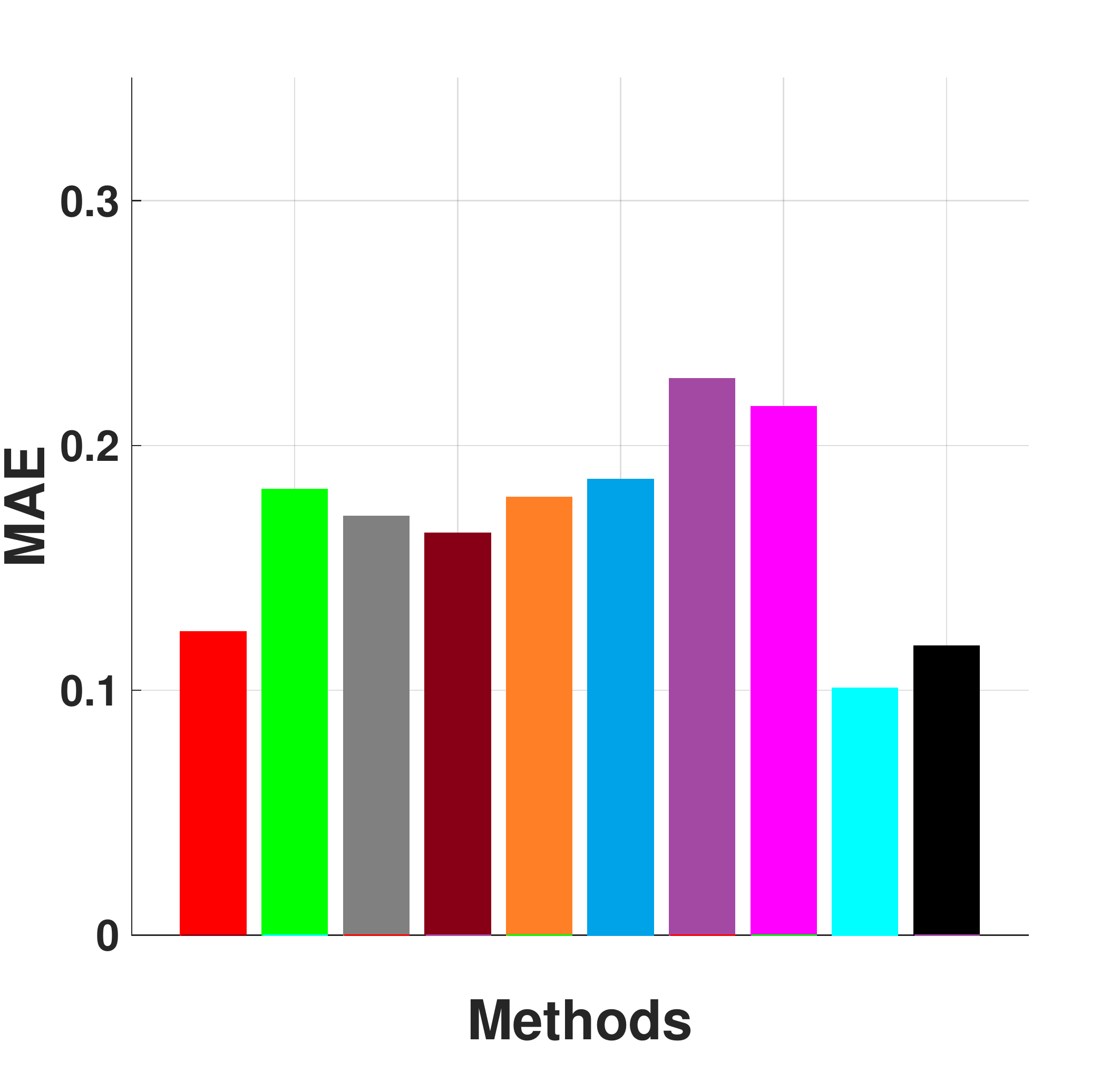}\\
  \includegraphics[width=0.55cm]{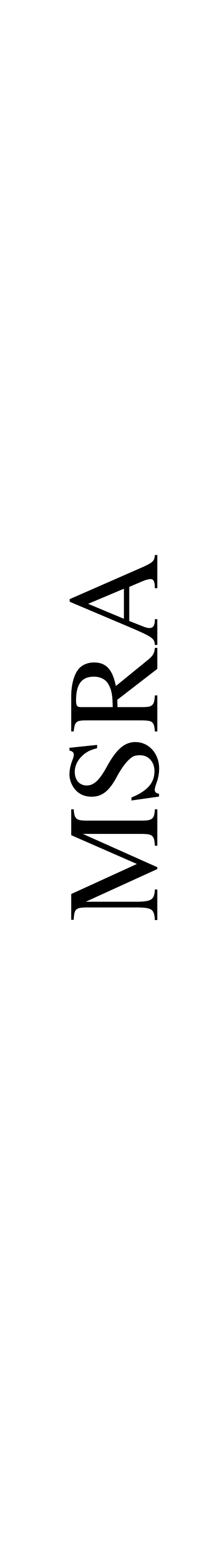}
  \includegraphics[width=5.7cm]{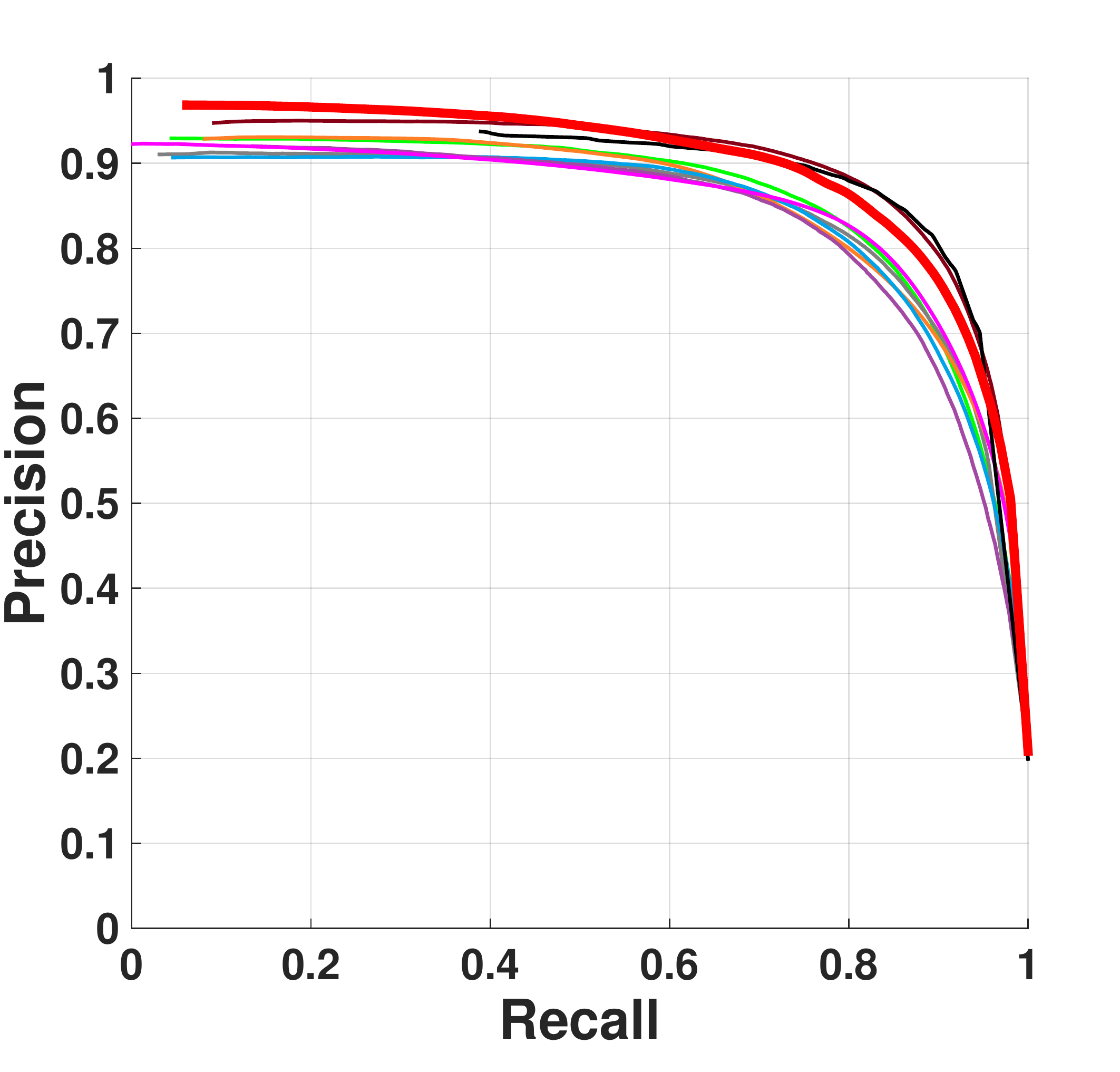}
  \includegraphics[width=5.7cm]{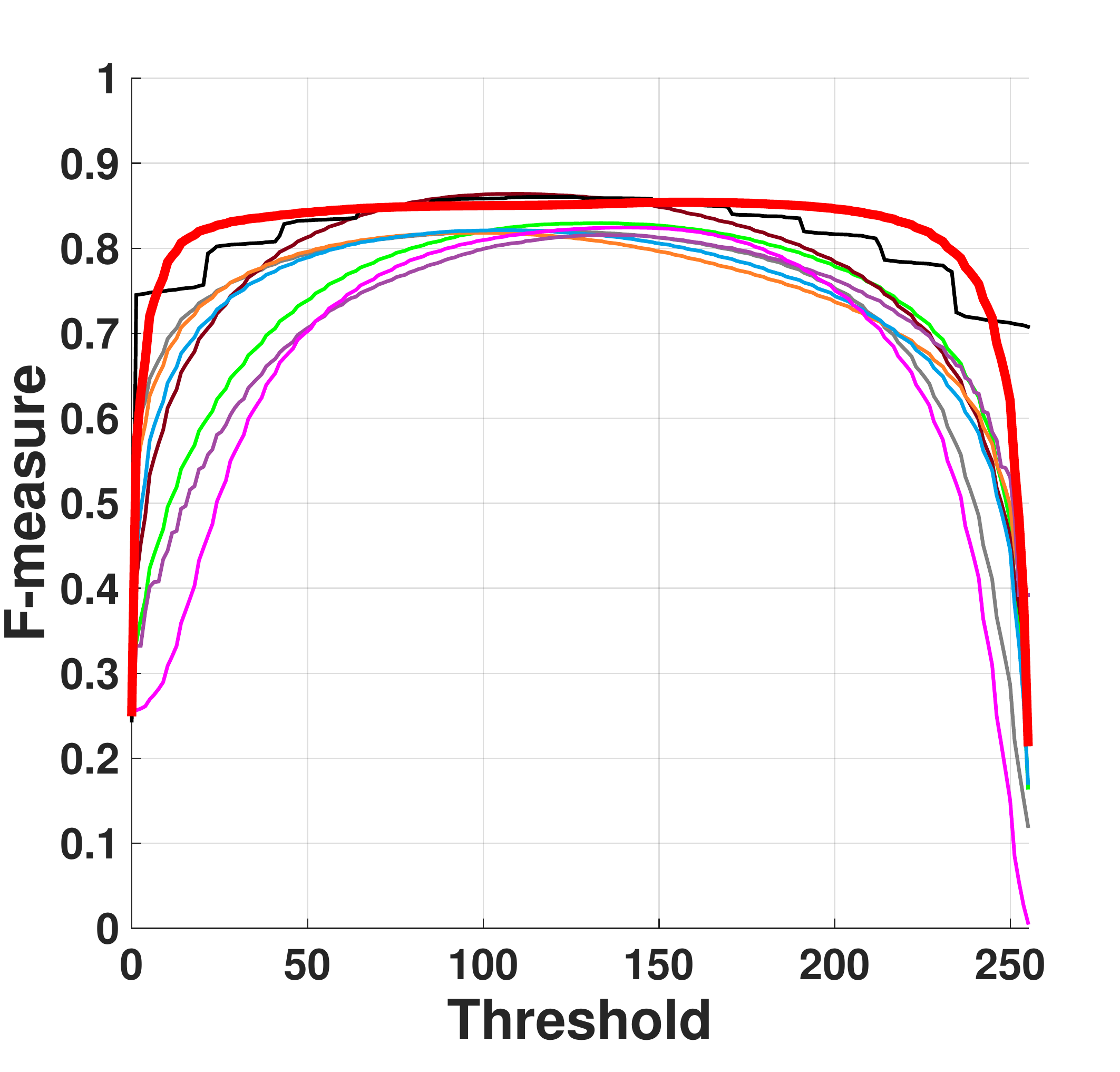}
  \includegraphics[width=5.7cm]{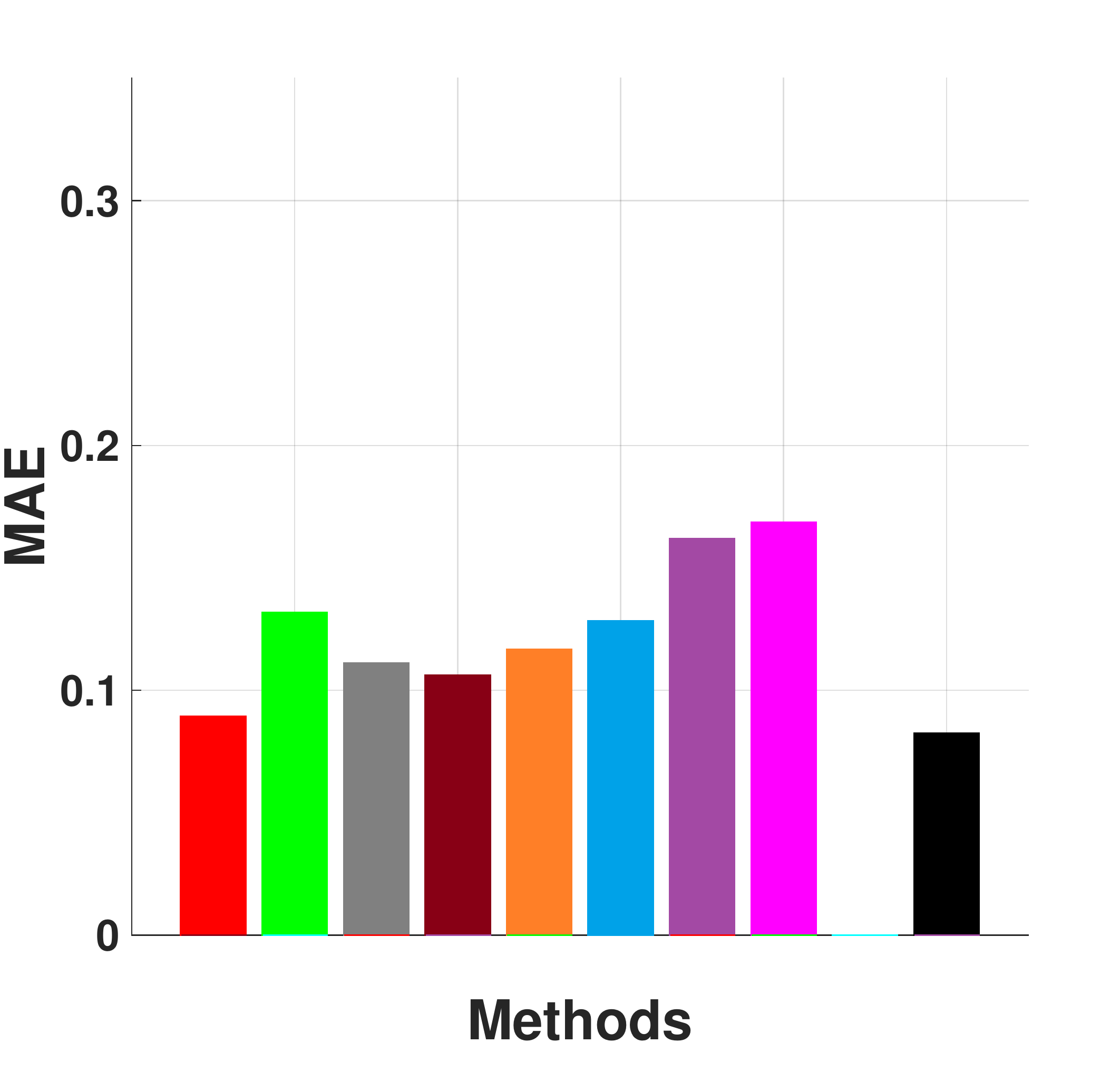}\\
  \includegraphics[width=0.55cm]{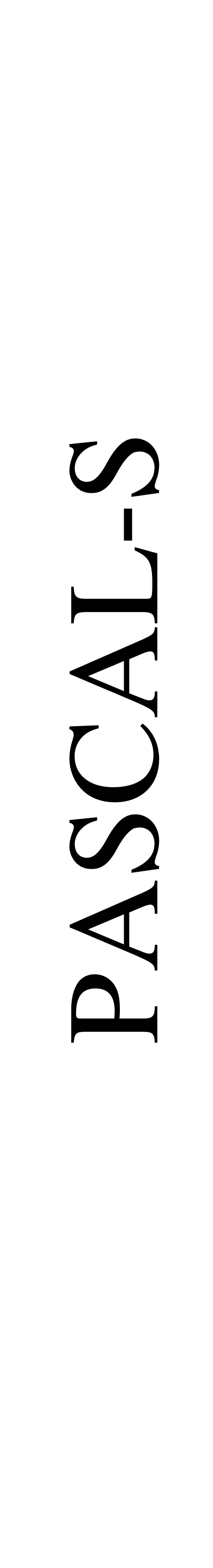}
  \includegraphics[width=5.7cm]{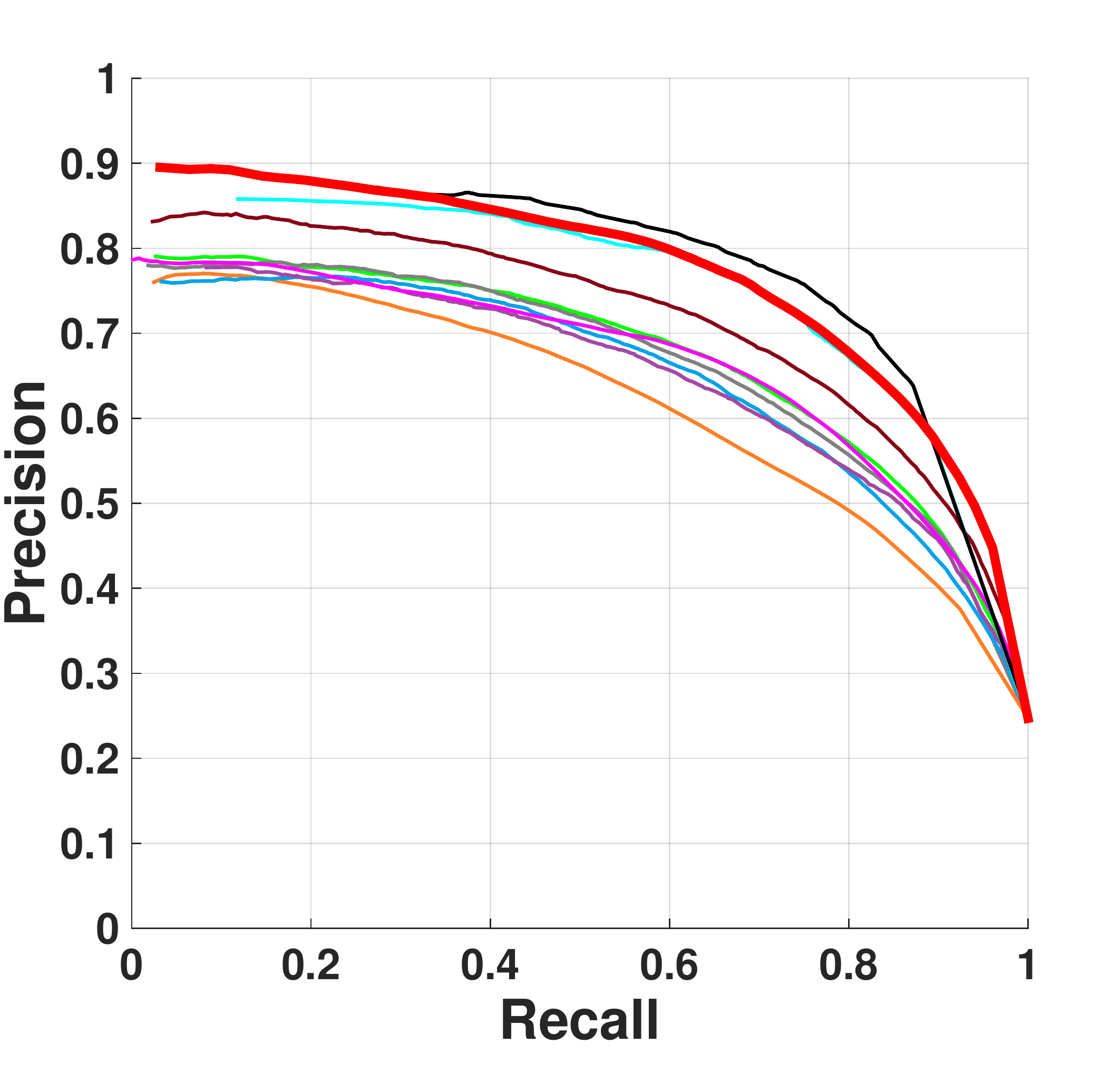}
  \includegraphics[width=5.7cm]{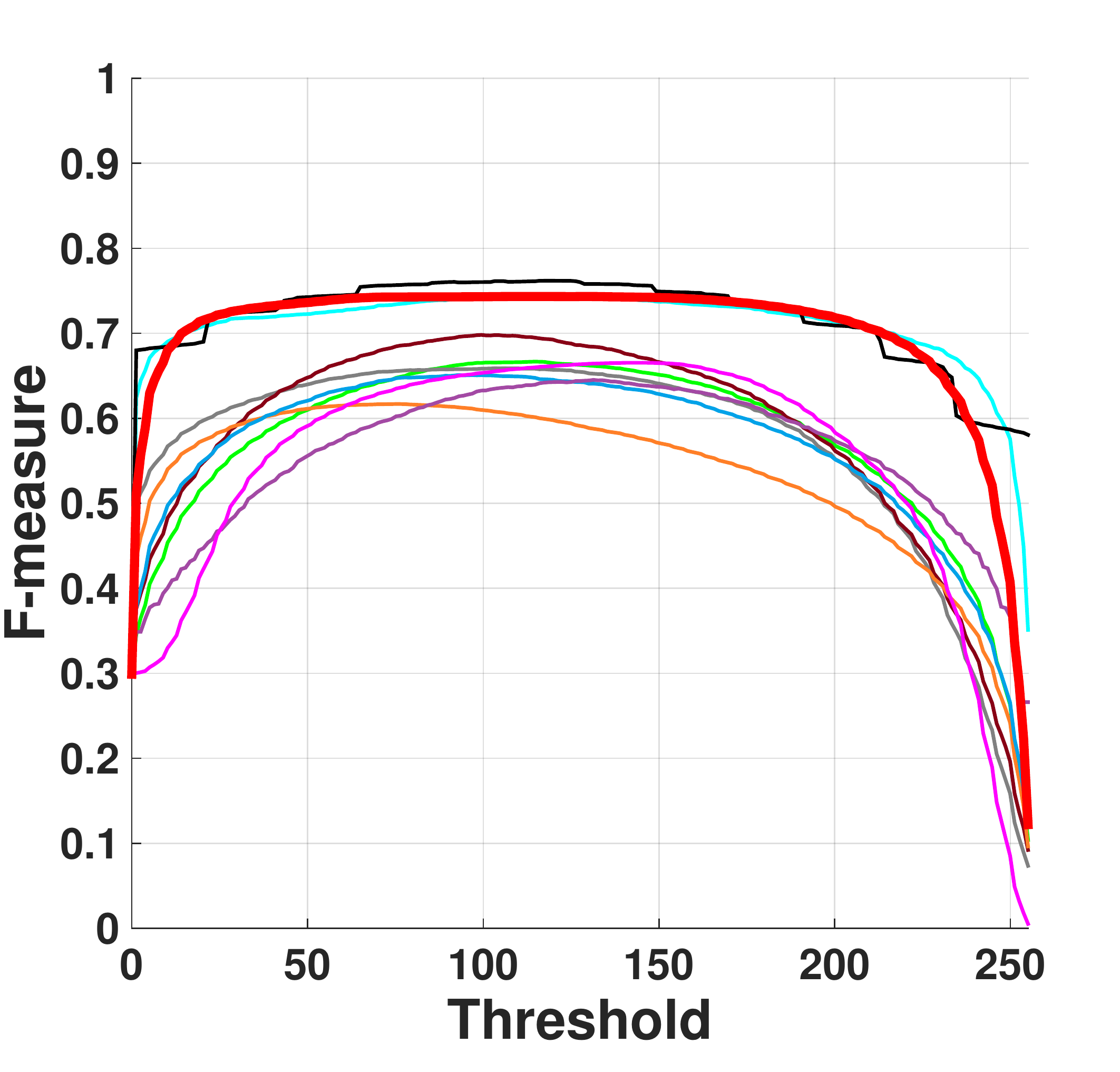}
  \includegraphics[width=5.7cm]{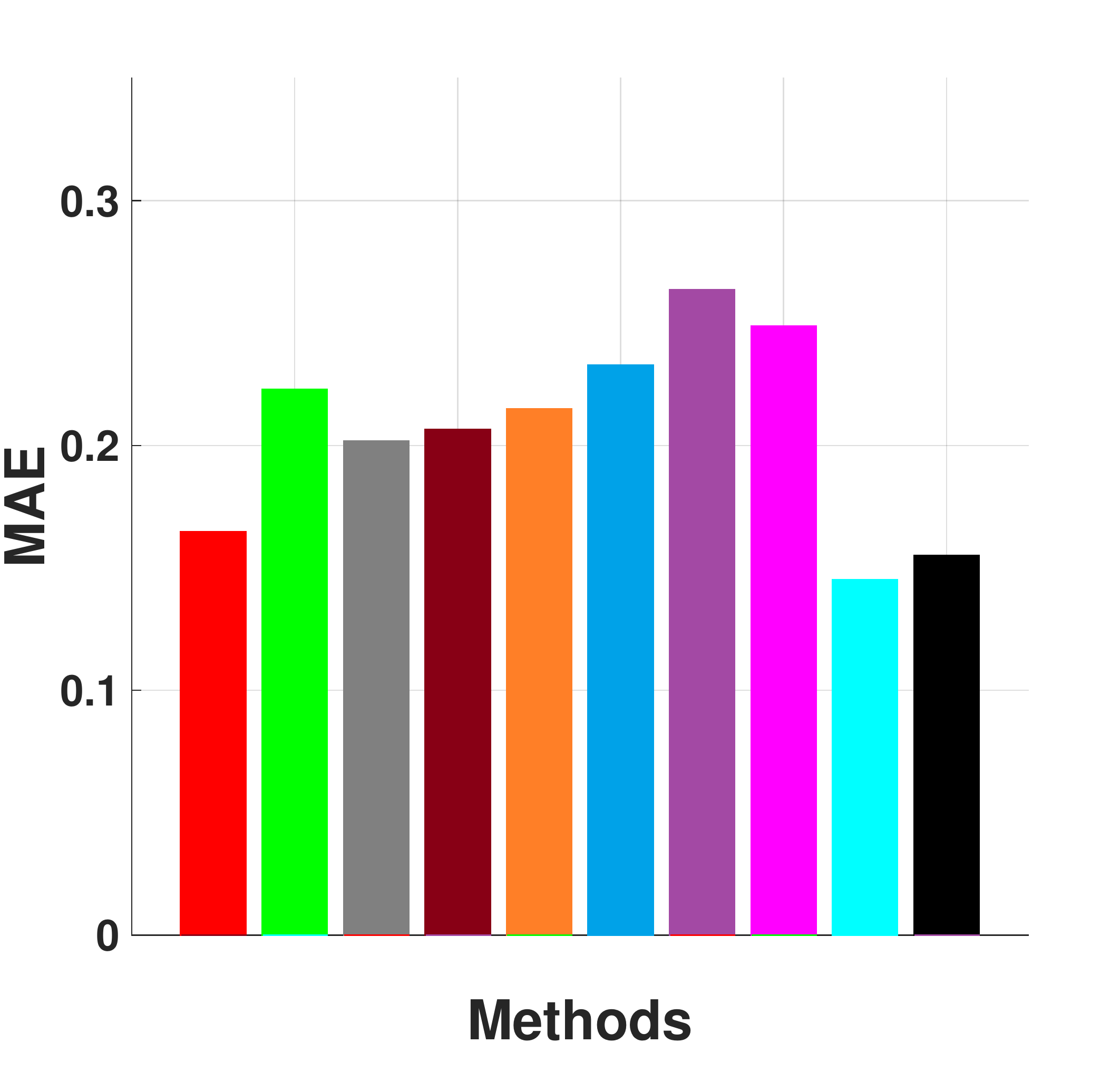}\\
  \includegraphics[width=0.55cm]{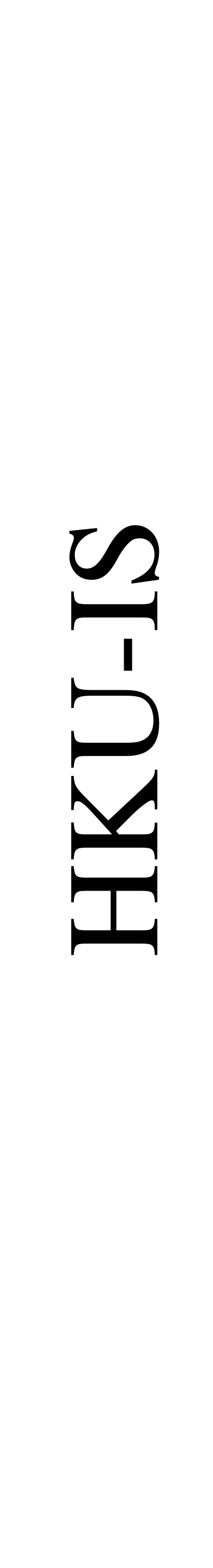}
  \subfigure[PR curves]{\includegraphics[width=5.7cm]{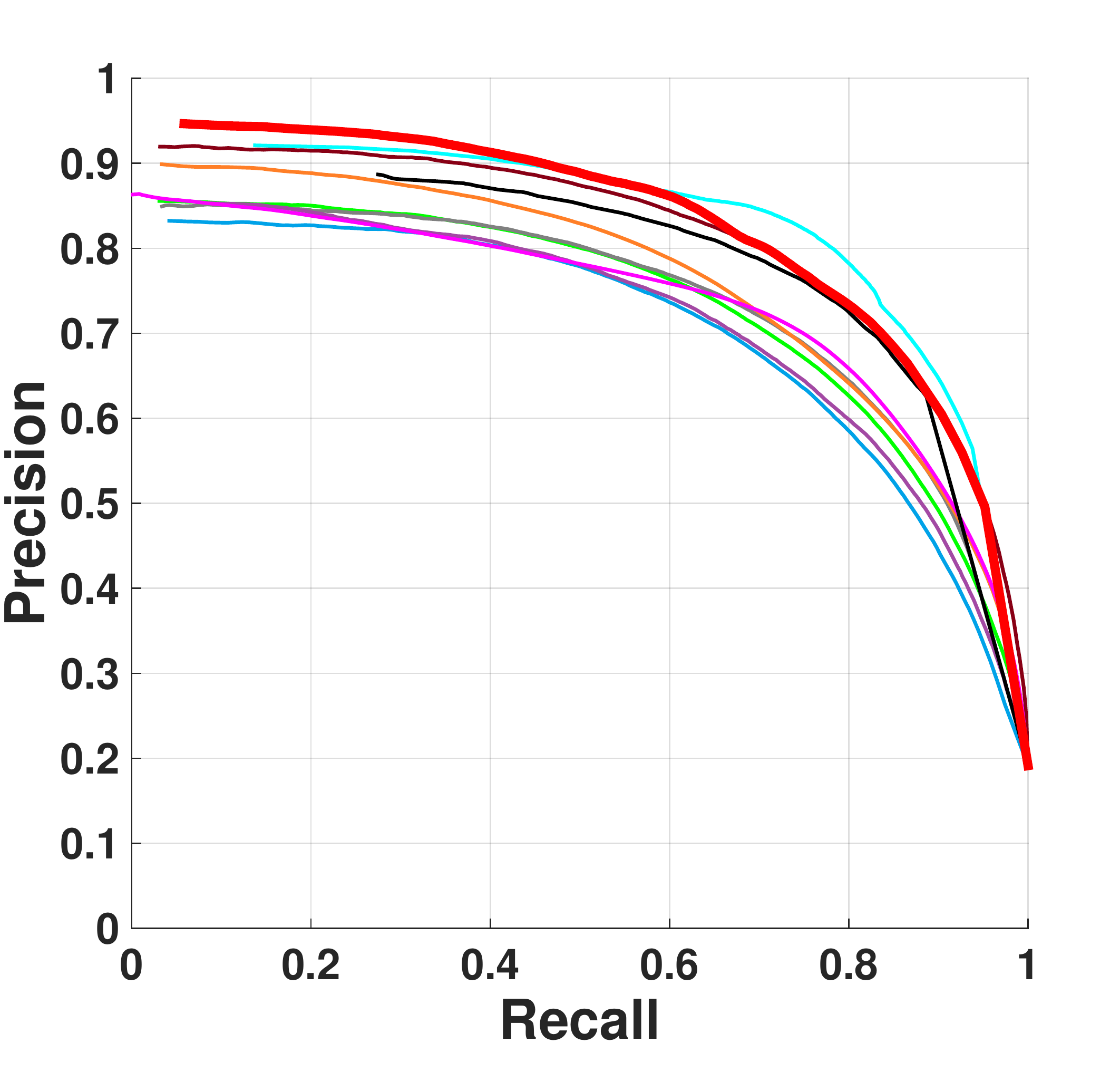}\label{com_with_STA-a}}
  \subfigure[FT curves]{\includegraphics[width=5.7cm]{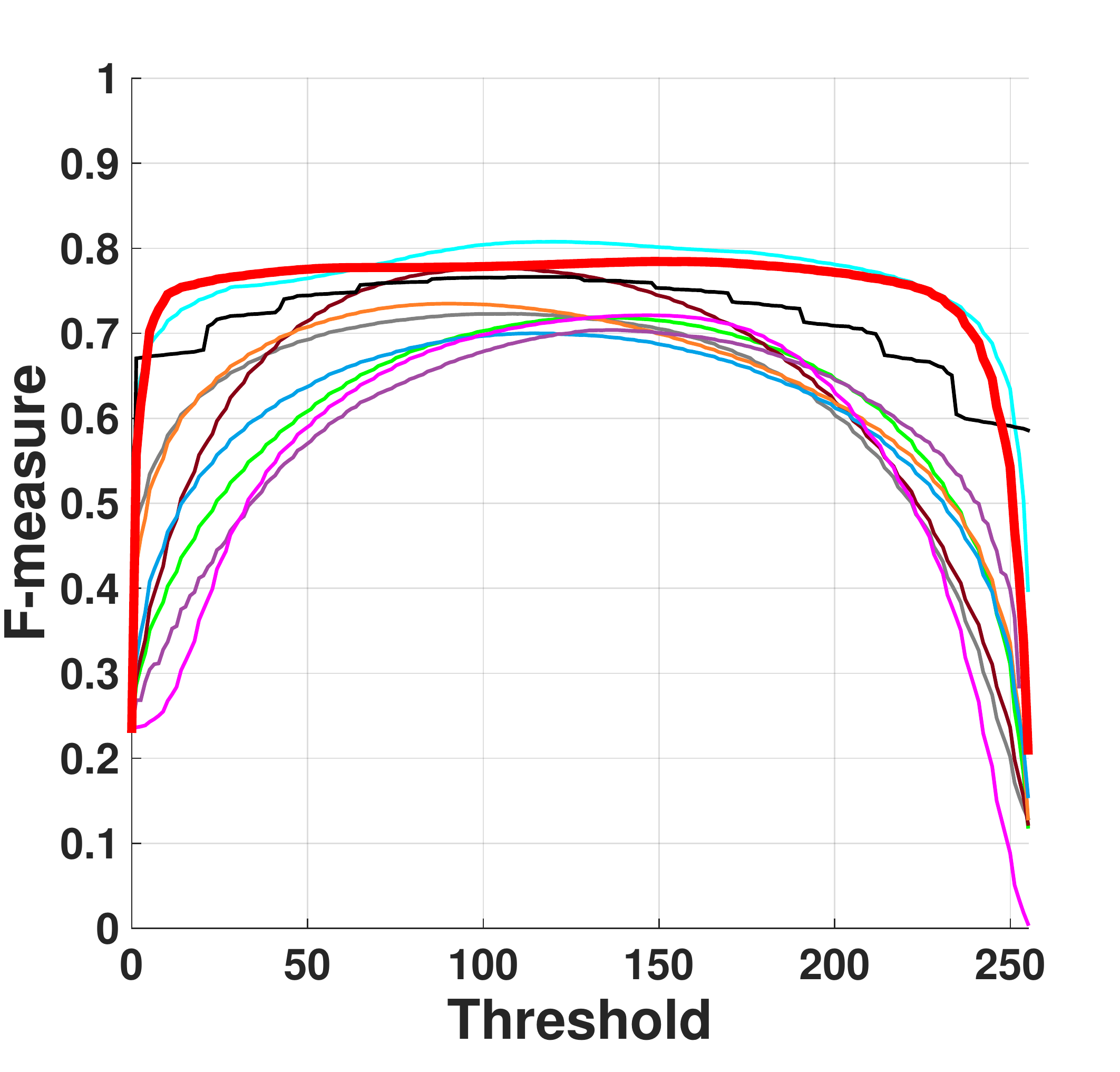}\label{com_with_STA-b}}
  \subfigure[MAE bars]{\includegraphics[width=5.7cm]{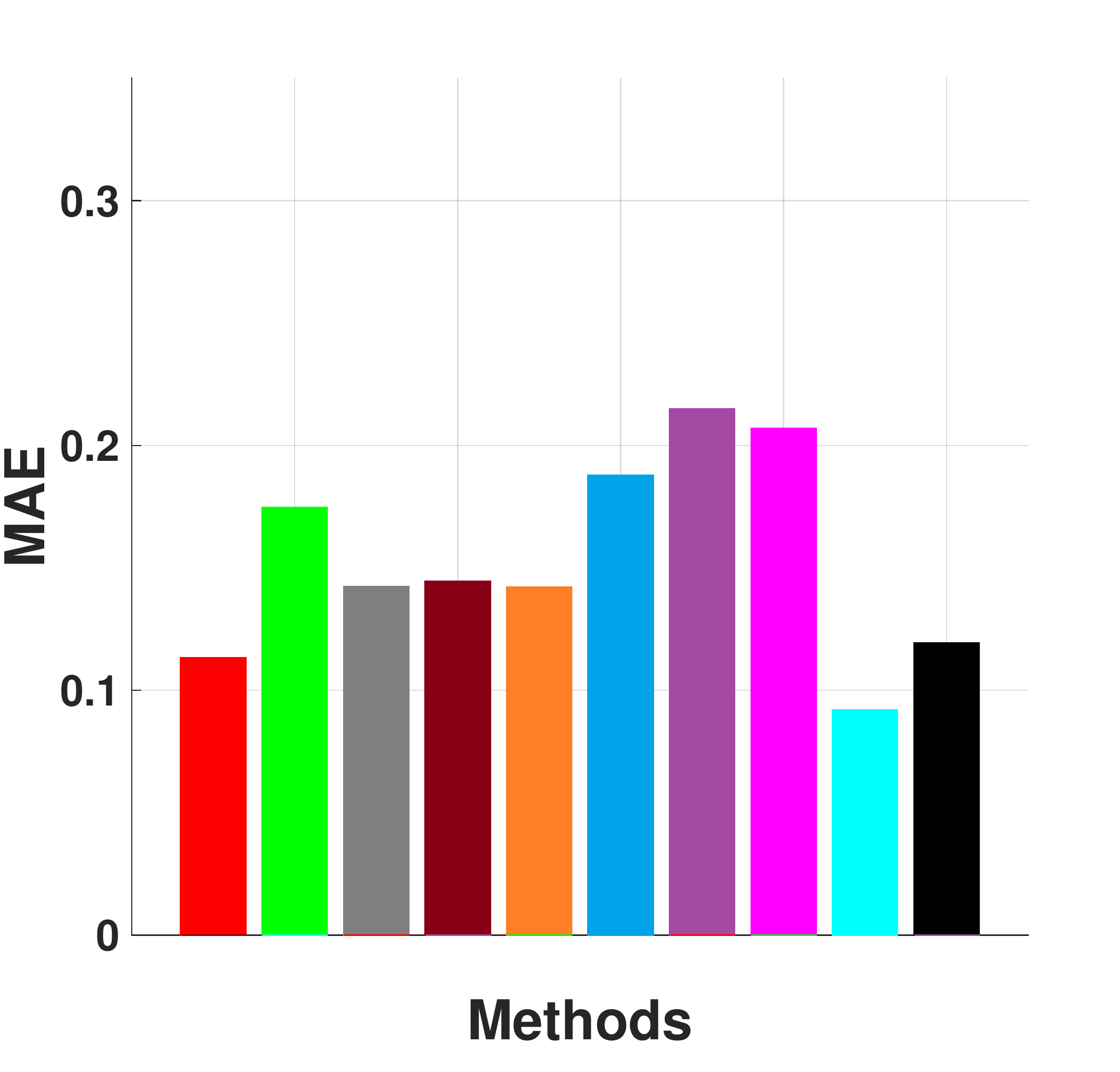}\label{com_with_STA-c}}\\
  \caption{PR curves, FT curves and MAE scores of different methods compared with our algorithm (HCA). From top to bottom: ECSSD, MSRA5000, PASCAL-S and HKU-IS are tested.}\label{com_with_STA}
\end{figure*}

\begin{figure*}
  \includegraphics[width=17.4cm]{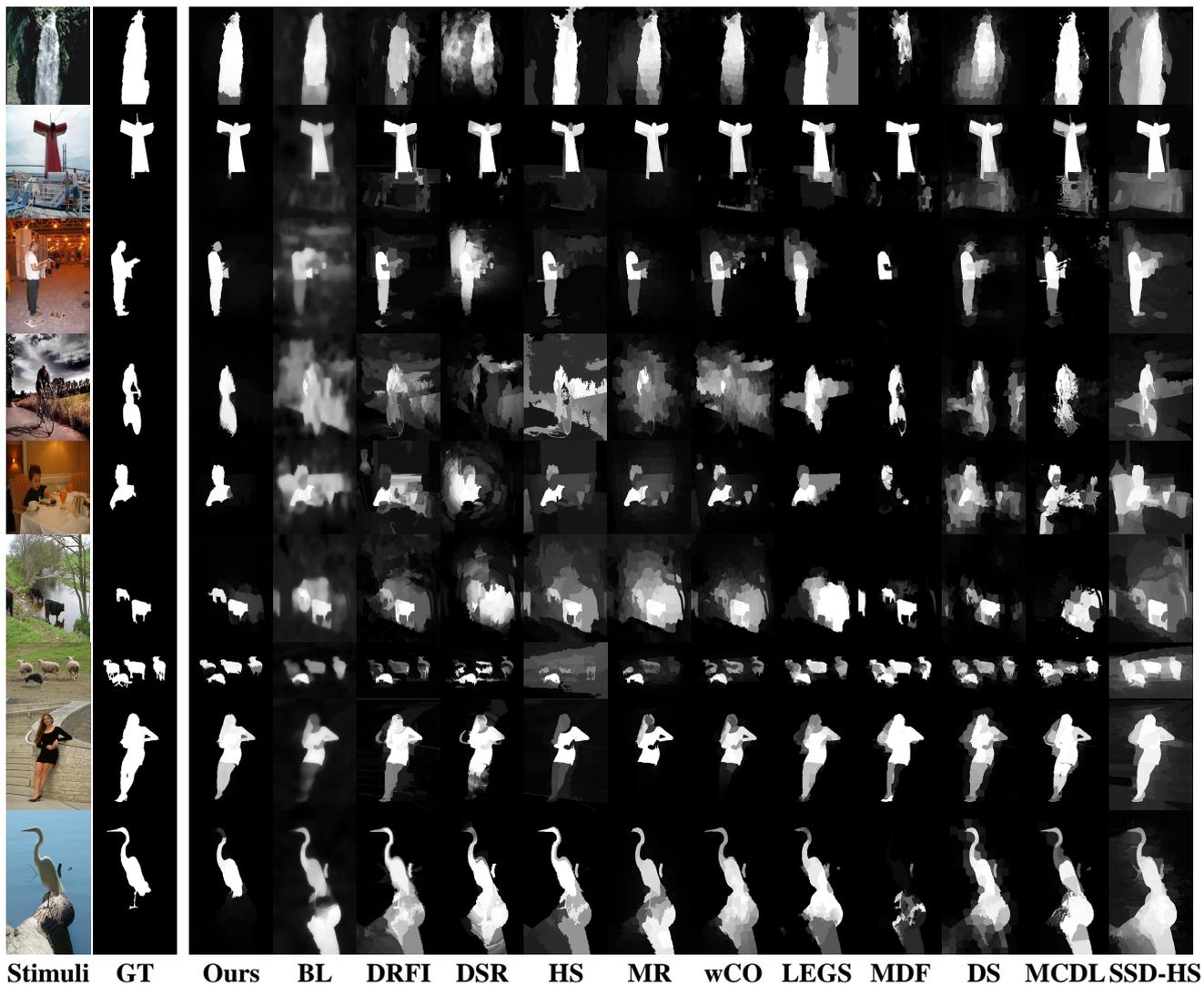}\\
  \caption{Visual comparison of saliency maps of different methods. GT: Ground Truth, Ours: Saliency maps generated by Hierarchical Cellular Automata (HCA).}\label{sm-com}
  \vspace{-2mm}
\end{figure*}

\vspace{-2mm}
As is shown in Figure~\ref{com_with_STA}, our proposed Hierarchical Cellular Automata performs favorably against state-of-the-art conventional algorithms with higher precision and recall values on four challenging datasets. HCA is competitive with deep learning based approaches and has higher precision at low levels of recall. Furthermore, the fairly low MAE scores, displayed in Figure.~\ref{com_with_STA-c}, indicate that our saliency maps are very close to the ground truth. As MCDL~\citep{zhao2015saliency} trained the network on the MSRA dataset, we do not report its result on this dataset in Figure.~\ref{com_with_STA}. In addition, LEGS~\citep{wang2015deep} used part of the images in the MSRA and PASCAL-S datasets as the training set. As a result, we only test LEGS with the test images on these two datasets. Saliency maps are shown in Figure.~\ref{sm-com} for visual comparison of our method with other models.
\vspace{-2mm}

  \vspace{-4mm}
\subsection{Optimization of state-of-the-art methods}
\vspace{-2mm}
In the previous sections, we showed qualitatively that our model creates better saliency maps by improving initial saliency maps with SCA, or by combining the results of multiple algorithms with CCA, or by applying SCA and CCA.  Here we compare our methods to other methods quantitatively. When the initial maps are imperfect, we apply SCA to improve them and then apply CCA. When the initial maps are already very good, we show that we can combine state-of-the-art methods to perform even better by simply using CCA.
\vspace{-4mm}
\subsubsection{Consistent Improvement}
\begin{figure*}
  \includegraphics[width=0.55cm]{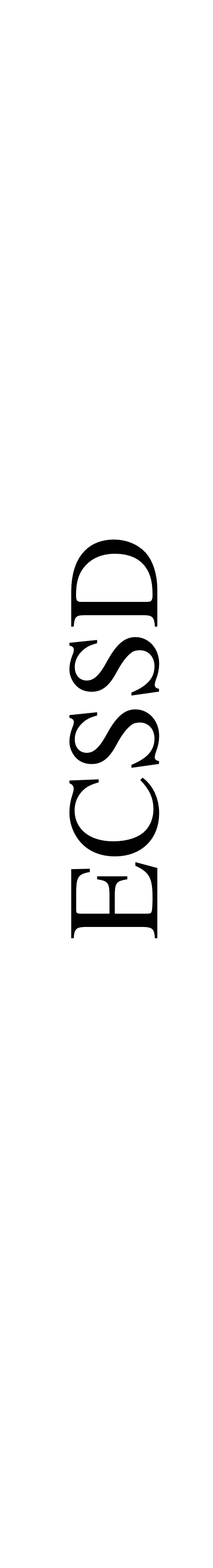}
  \includegraphics[width=5.7cm,height=4.5cm]{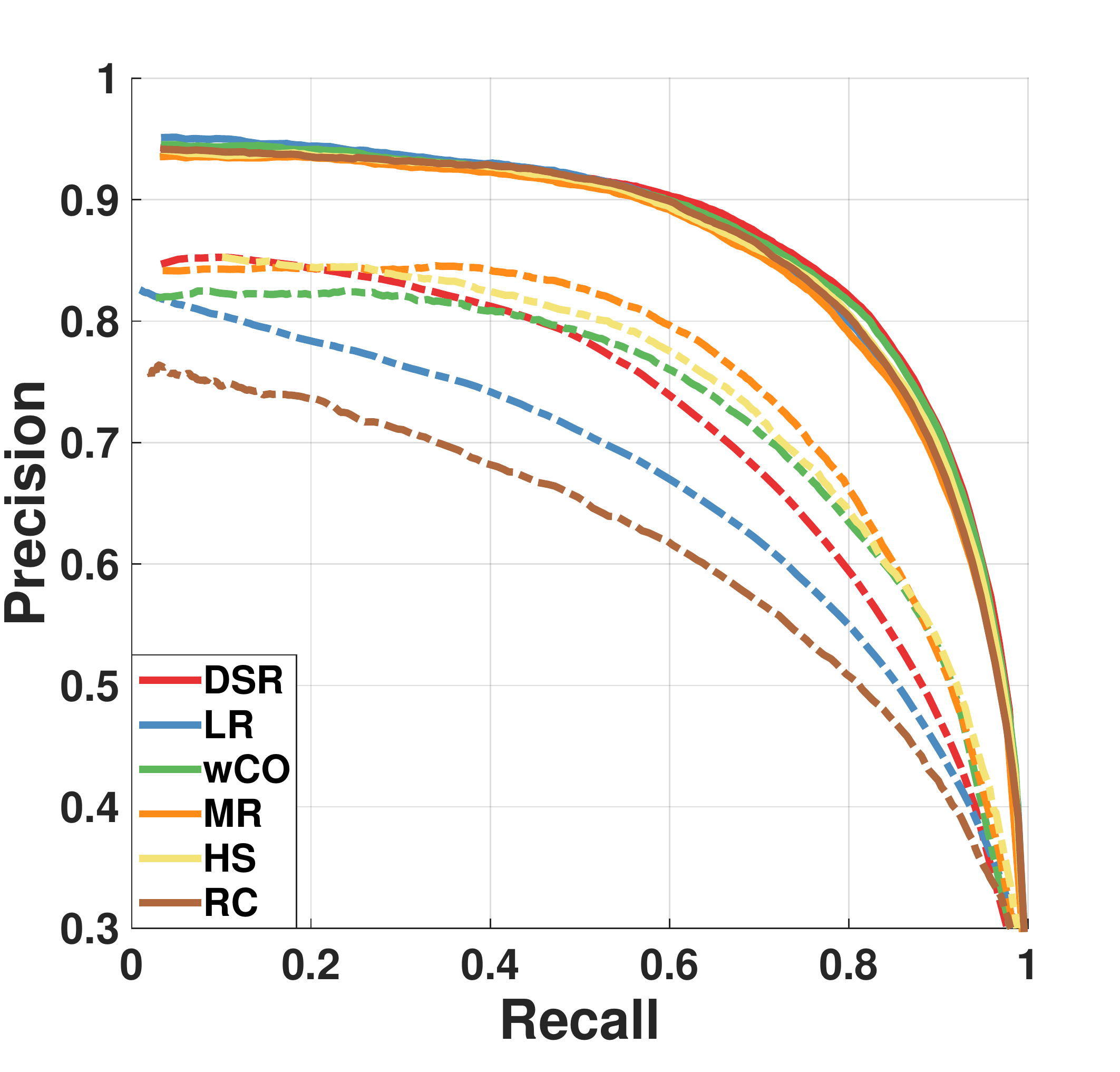}
  \includegraphics[width=5.7cm,height=4.5cm]{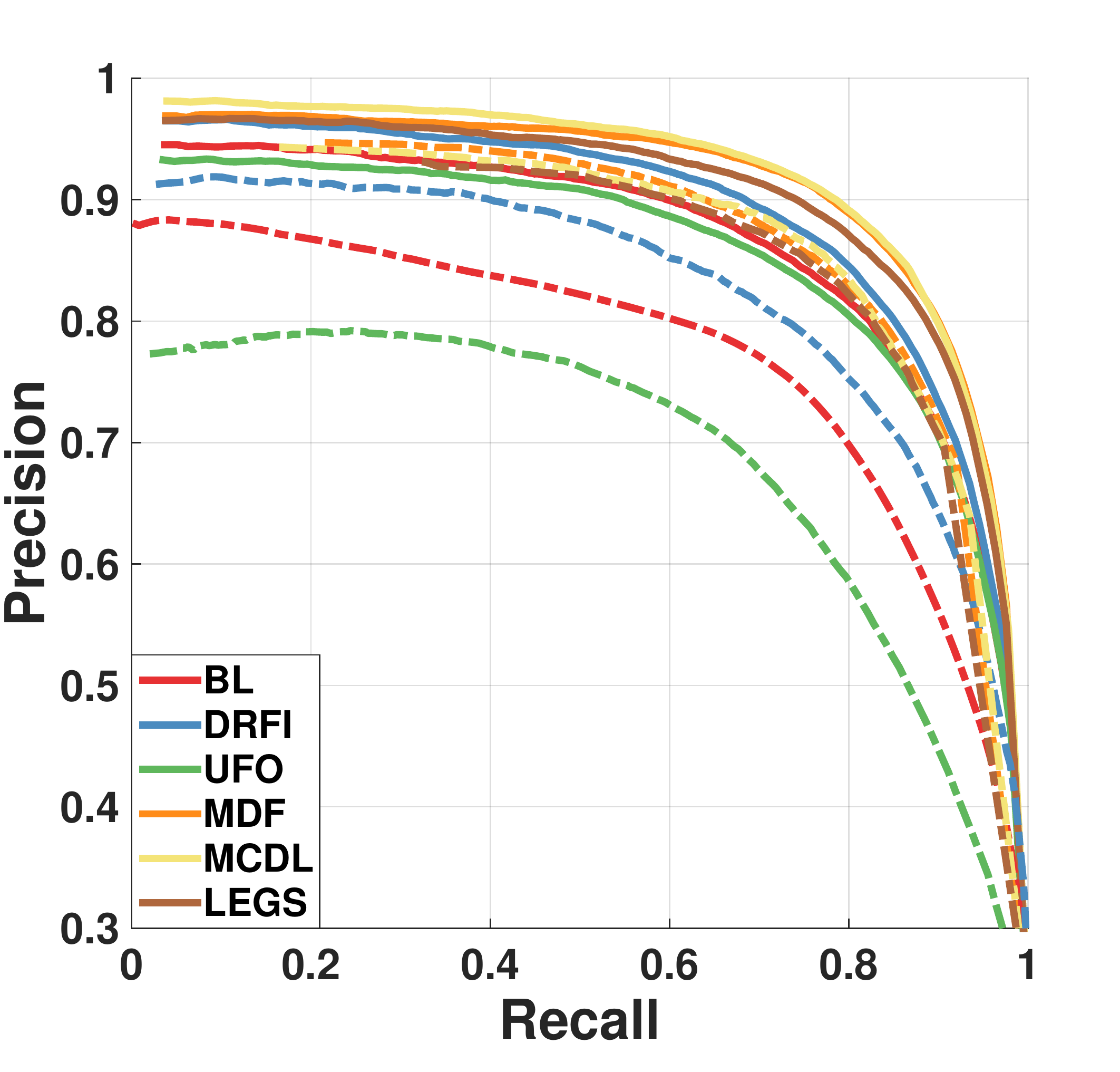}
  \includegraphics[width=5.7cm,height=4.5cm]{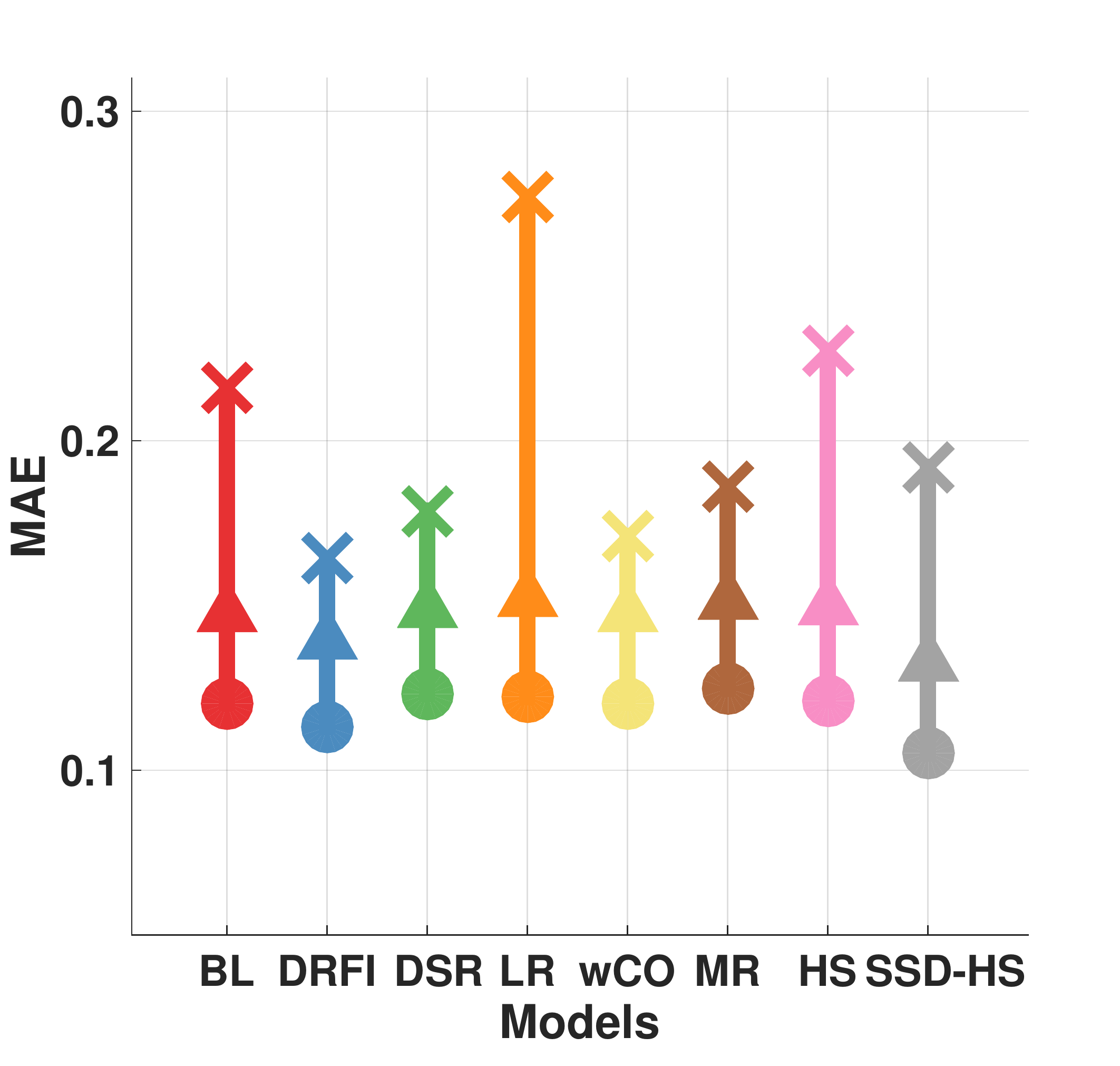}\\
  \includegraphics[width=0.55cm]{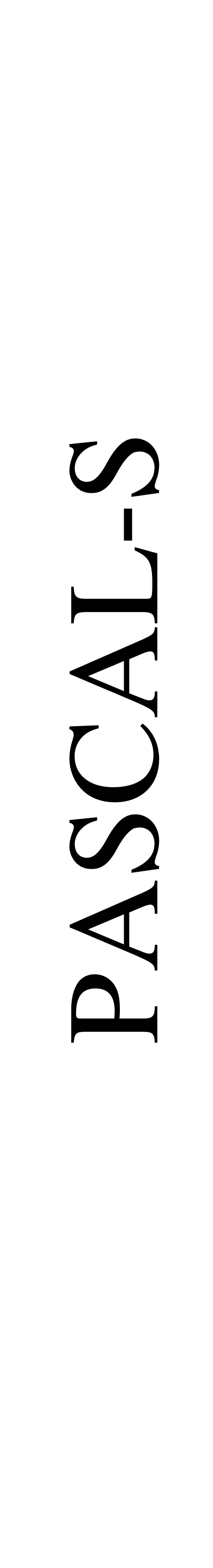}
  \subfigure[PR curves]{\includegraphics[width=5.7cm,height=4.5cm]{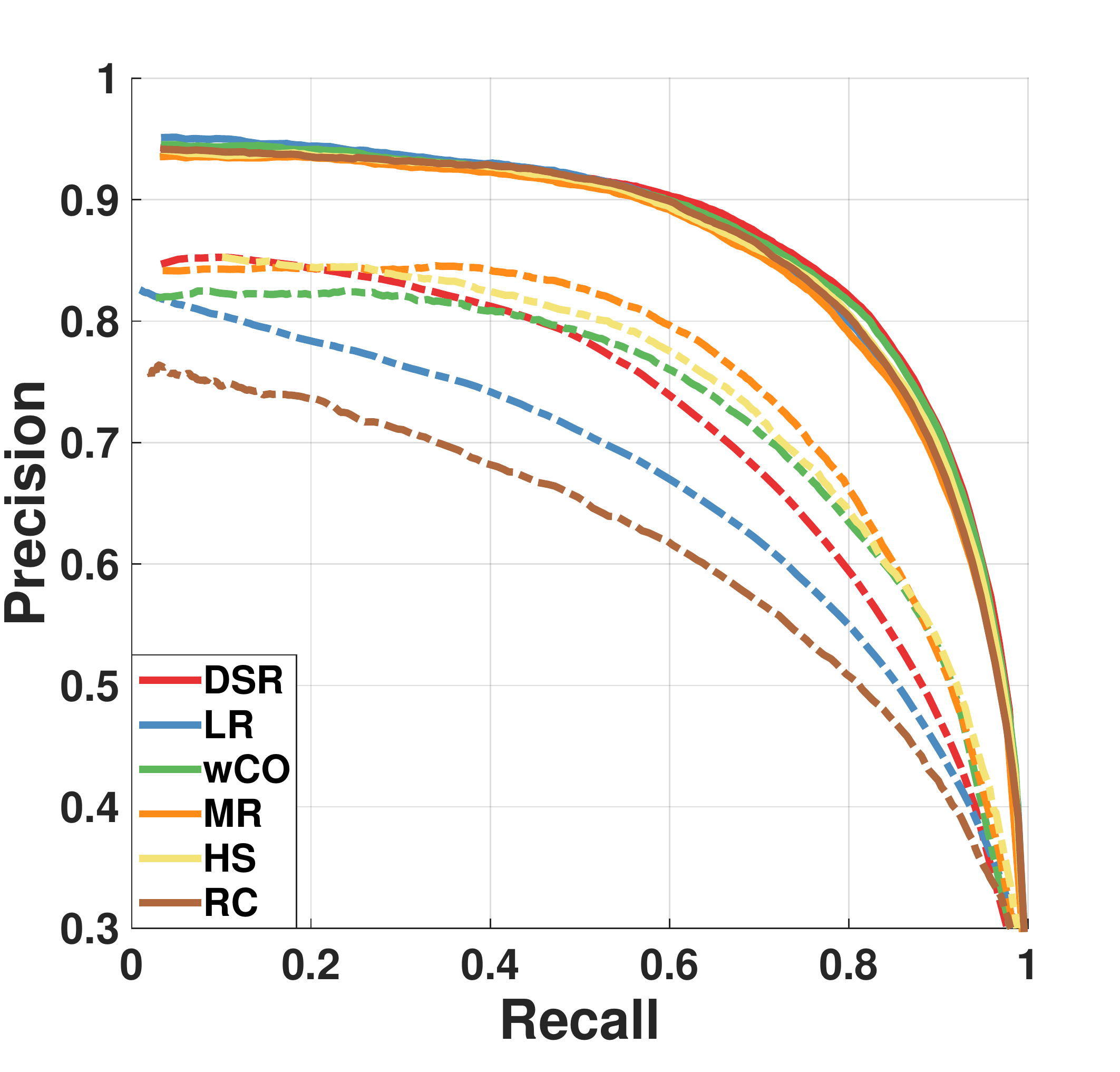}\label{M-vs-M-sca.a}}
  \subfigure[FT curves]{\includegraphics[width=5.7cm,height=4.5cm]{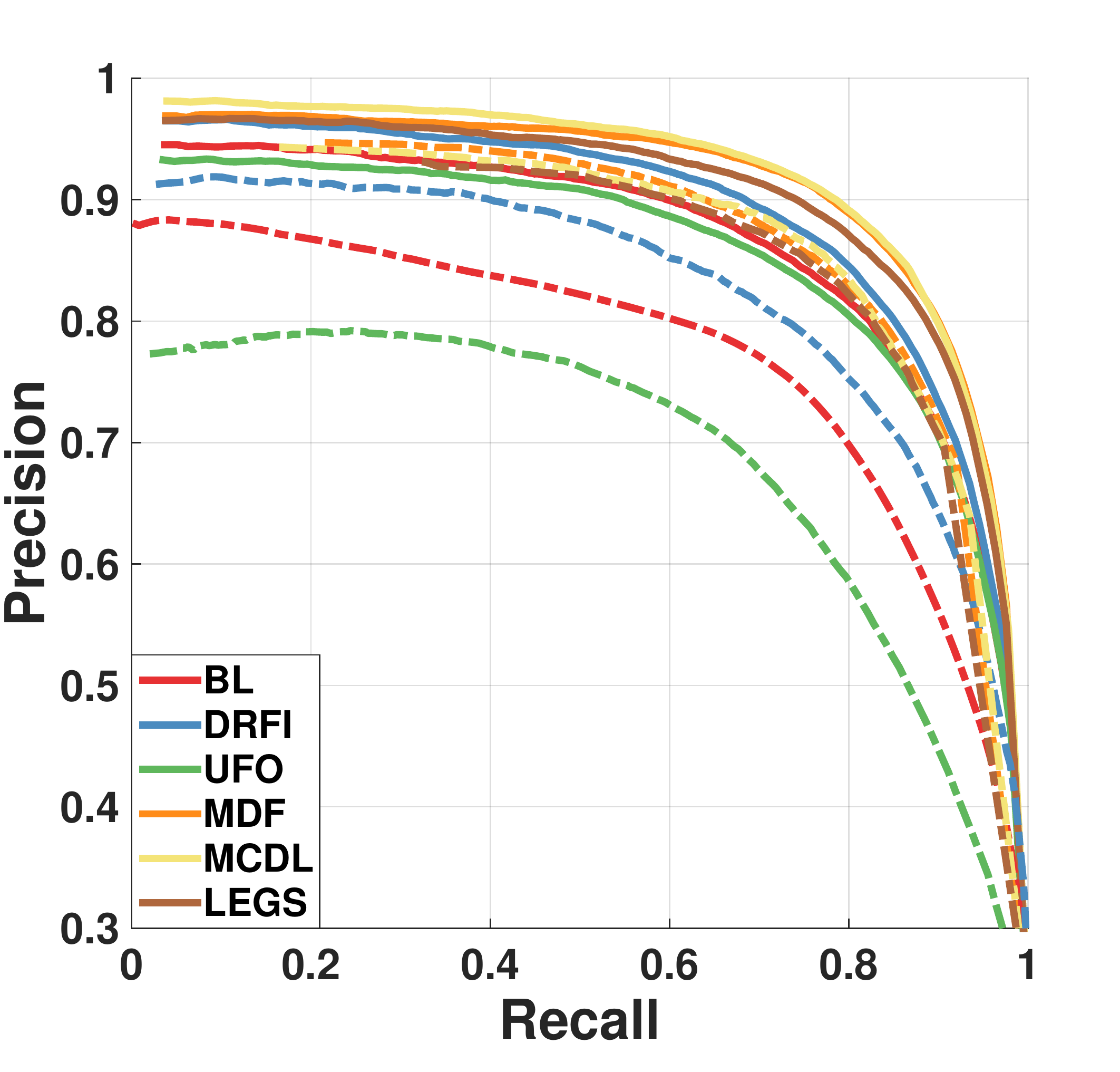}\label{M-vs-M-sca.b}}
  \subfigure[MAE scores]{\includegraphics[width=5.7cm,height=4.5cm]{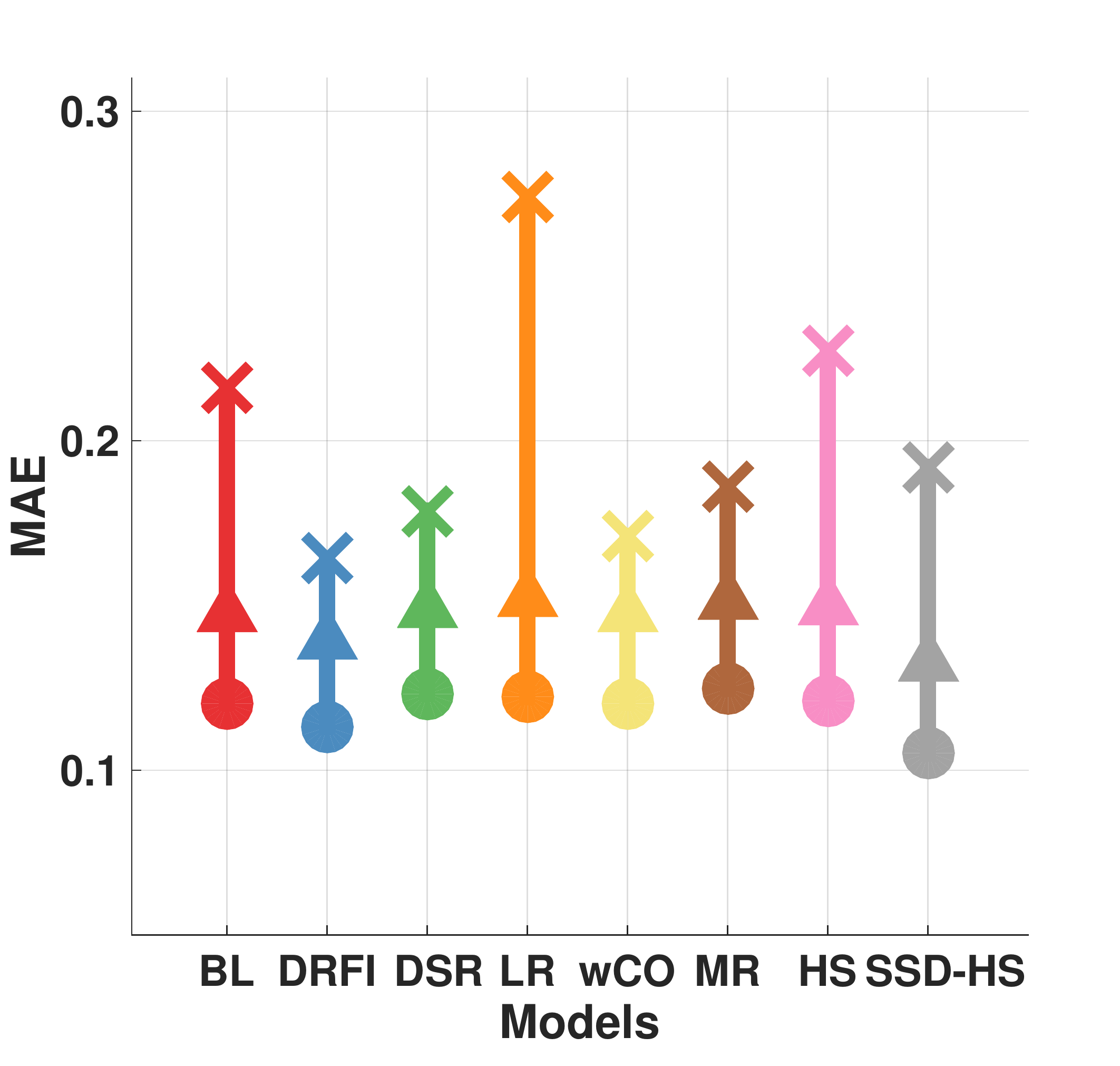}\label{M-vs-M-sca.c}}\\
  \vspace{-3mm}
  \caption{Consistent improvement on ECSSD and PASCALS datasets. (a) and (b): PR curves of different methods (dashed line) and their optimized version via SCA200 (solid line). The right column shows that SCA200 ($\bigtriangleup$), improves the MAEs of the original methods ($\times$)  and that HCA* ($\bigcirc$), here applied to SCA120, SCA160, and SCA200, further improves the results.}\label{M-vs-M-sca}
    \vspace{-1mm}
\end{figure*}

{\label{sca_im}}
\vspace{-2mm}
In Section~\ref{sop}, we concluded that results generated by different methods can be effectively optimized via Single-layer Cellular Automata. Figure~\ref{M-vs-M-sca} shows the precision-recall curves and mean absolute error bars of various saliency methods and their optimized results on two datasets. These results demonstrate that SCA can greatly improve existing results to a similar precision level. Even though the original saliency maps are not well constructed, the optimized results are comparable to the state-of-the-art methods. It should be noted that SCA can even optimize deep learning-based methods to a better precision level, e.g., MCDL~\citep{zhao2015saliency}, MDF~\citep{li2015visual}, LEGS~\citep{wang2015deep}, and SSD-HS~\citep{ssd2016eccv}. In addition, for one existing method, we can use SCA to optimize it at different scales and then use CCA to integrate the multi-scale saliency maps. The ultimate optimized result is denoted as HCA*. The lowest MAEs of saliency maps optimized by HCA in Figure~\ref{M-vs-M-sca} (c) show that HCA's use of CCA improves performance over SCA alone.
  \vspace{-3mm}
\begin{table}[t]
\center
\caption{Run Time of Each Component of HCA}\label{runtime-our}
\begin{adjustbox}{max width=0.48\textwidth}
\begin{tabu}{cccccc}
\tabucline[1pt]{-}
  \hline

  \footnotesize{\textbf{Method}} & \footnotesize{\textbf{SCA120}} & \footnotesize{\textbf{SCA160}} & \footnotesize{\textbf{SCA200}} & \footnotesize{\textbf{CCA}} & \footnotesize{\textbf{HCA}} \\
  \hline
  \footnotesize{\textbf{w/ SLIC(s)}} & .2889 & .3134 & .3380 & -  & 1.0240   \\

  \footnotesize{\textbf{wo/ SLIC(s)}} & .0704 & .0525 & .0355 & .0837 & .2421   \\
  \tabucline[1pt]{-}
\vspace{-7mm}
\end{tabu}
\end{adjustbox}

\end{table}

\begin{figure*}
\raggedleft
  \includegraphics[width=5.7cm,height=4.5cm]{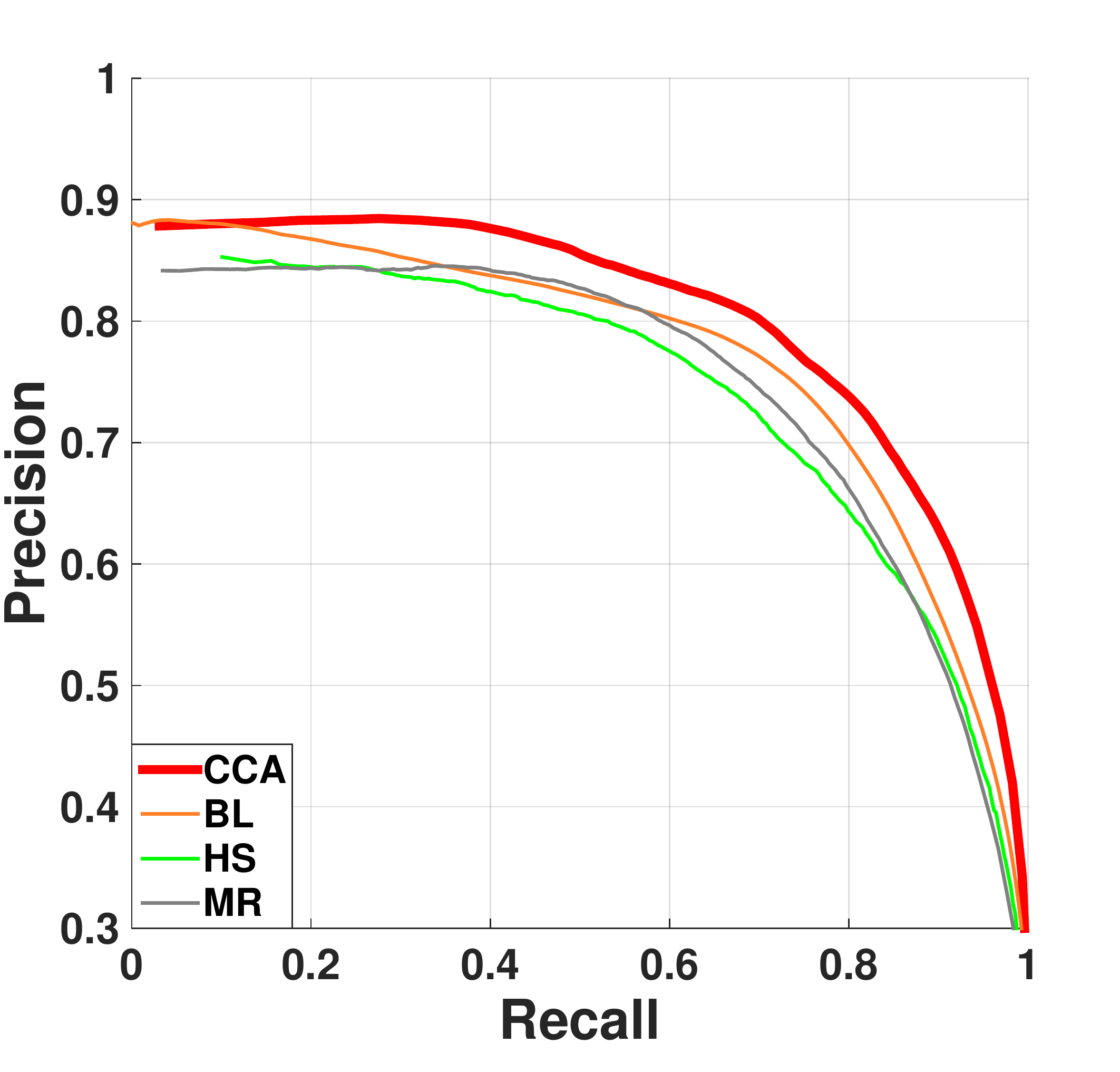}
  \includegraphics[width=5.7cm,height=4.5cm]{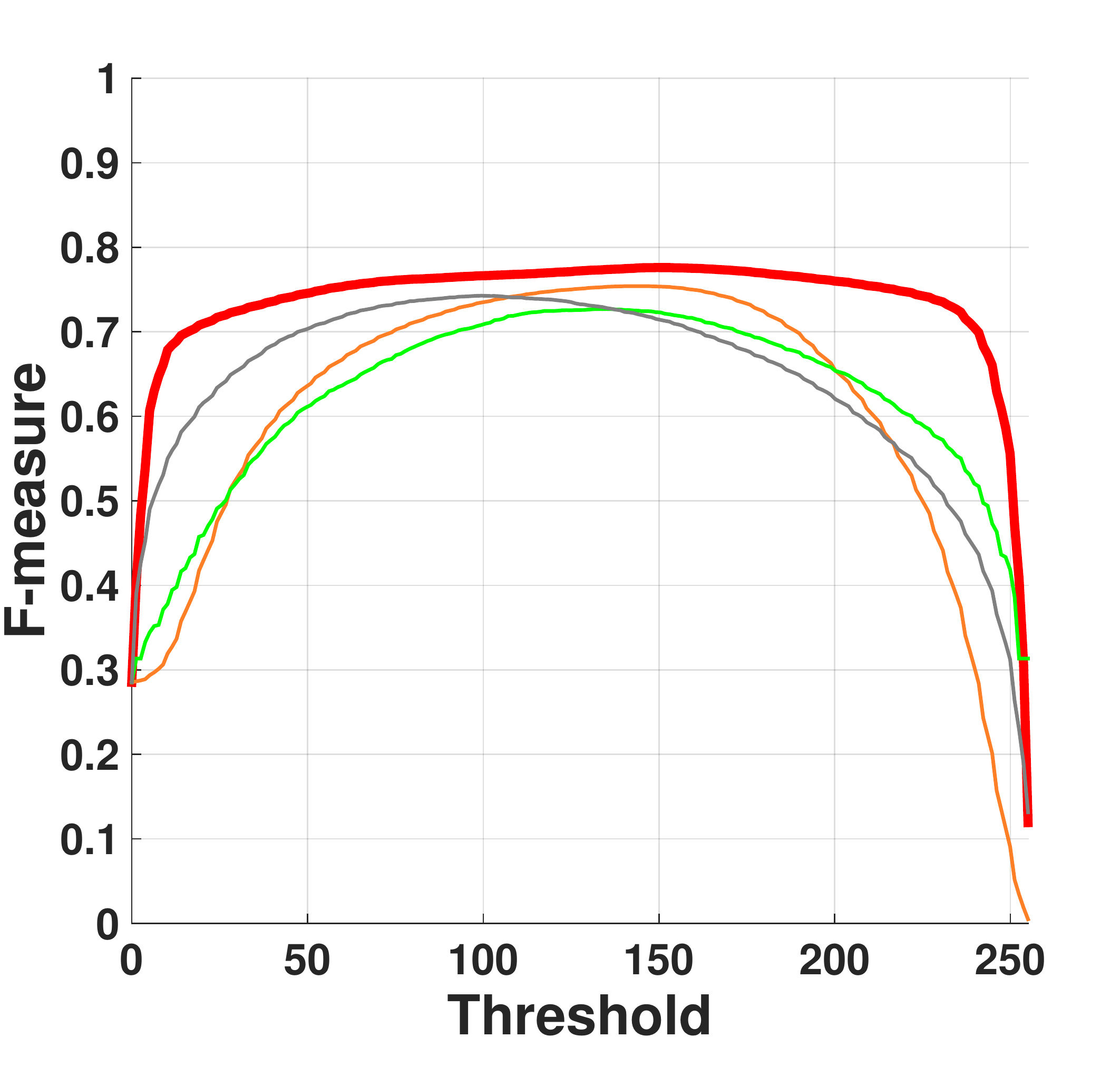}
  \includegraphics[width=5.2cm,height=4.5cm]{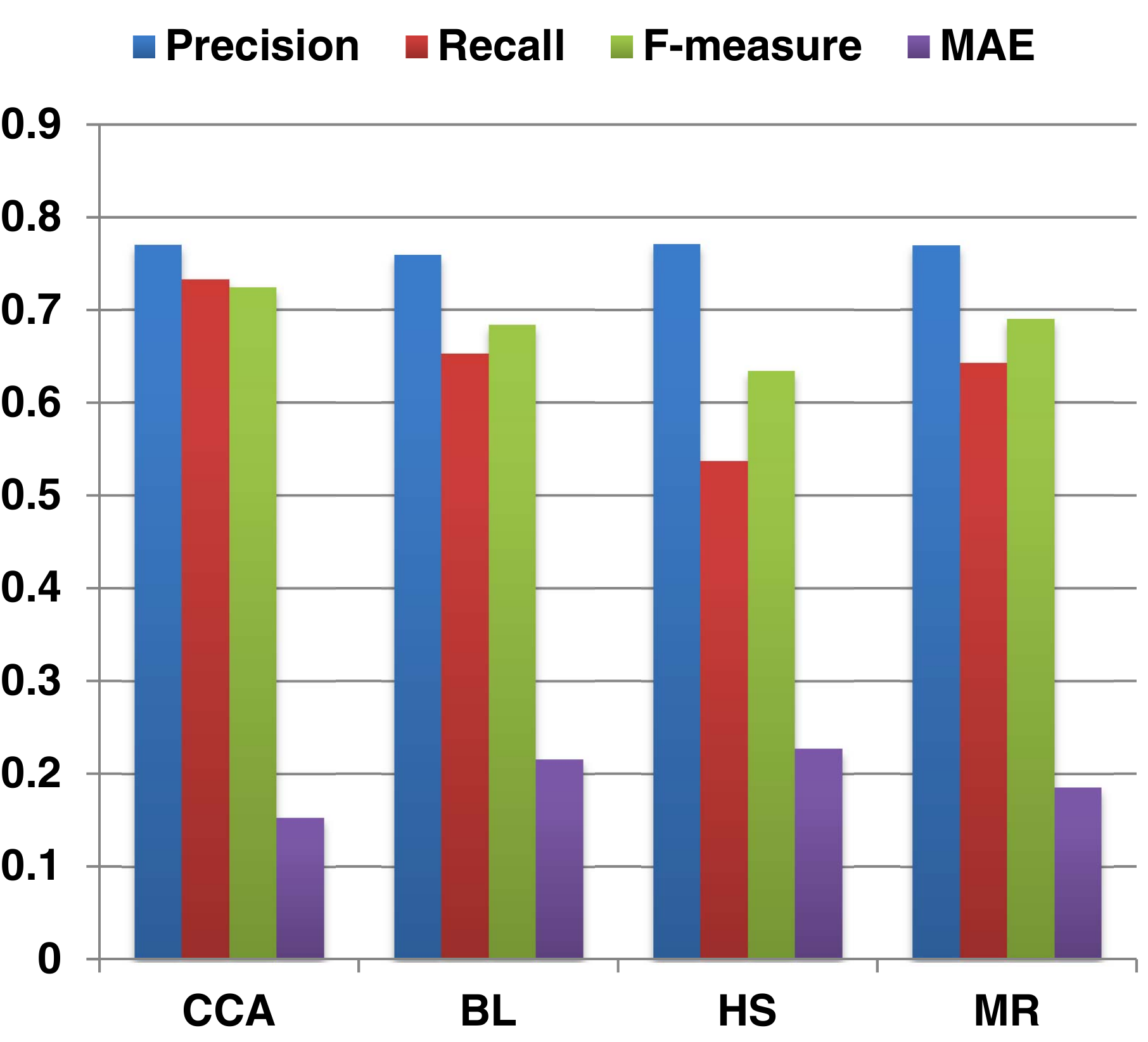}\\

  \subfigure[PR curves]{\includegraphics[width=5.7cm,height=4.5cm]{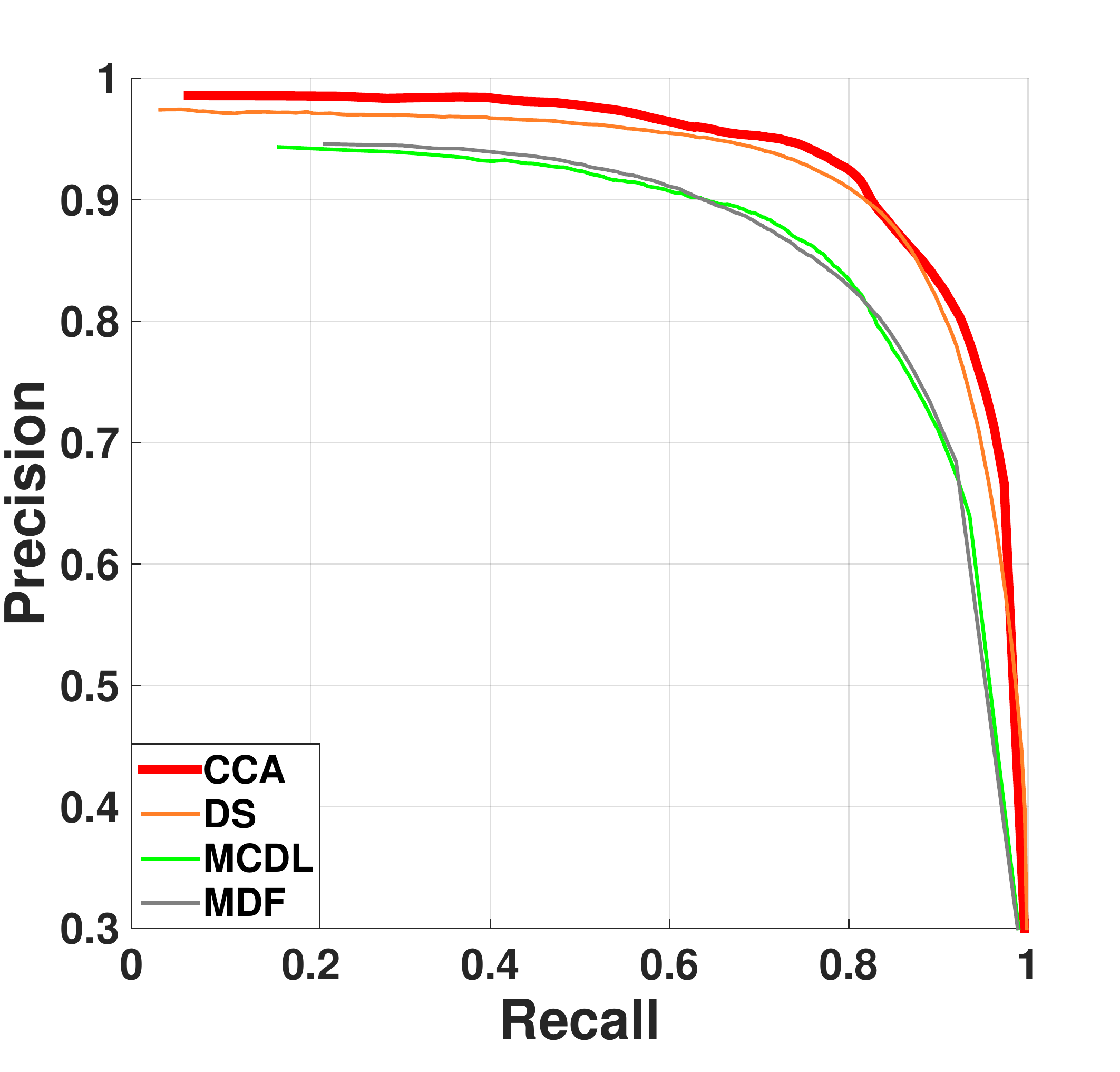}\label{integration-sta.a}}
  \subfigure[FT curves]{\includegraphics[width=5.7cm,height=4.5cm]{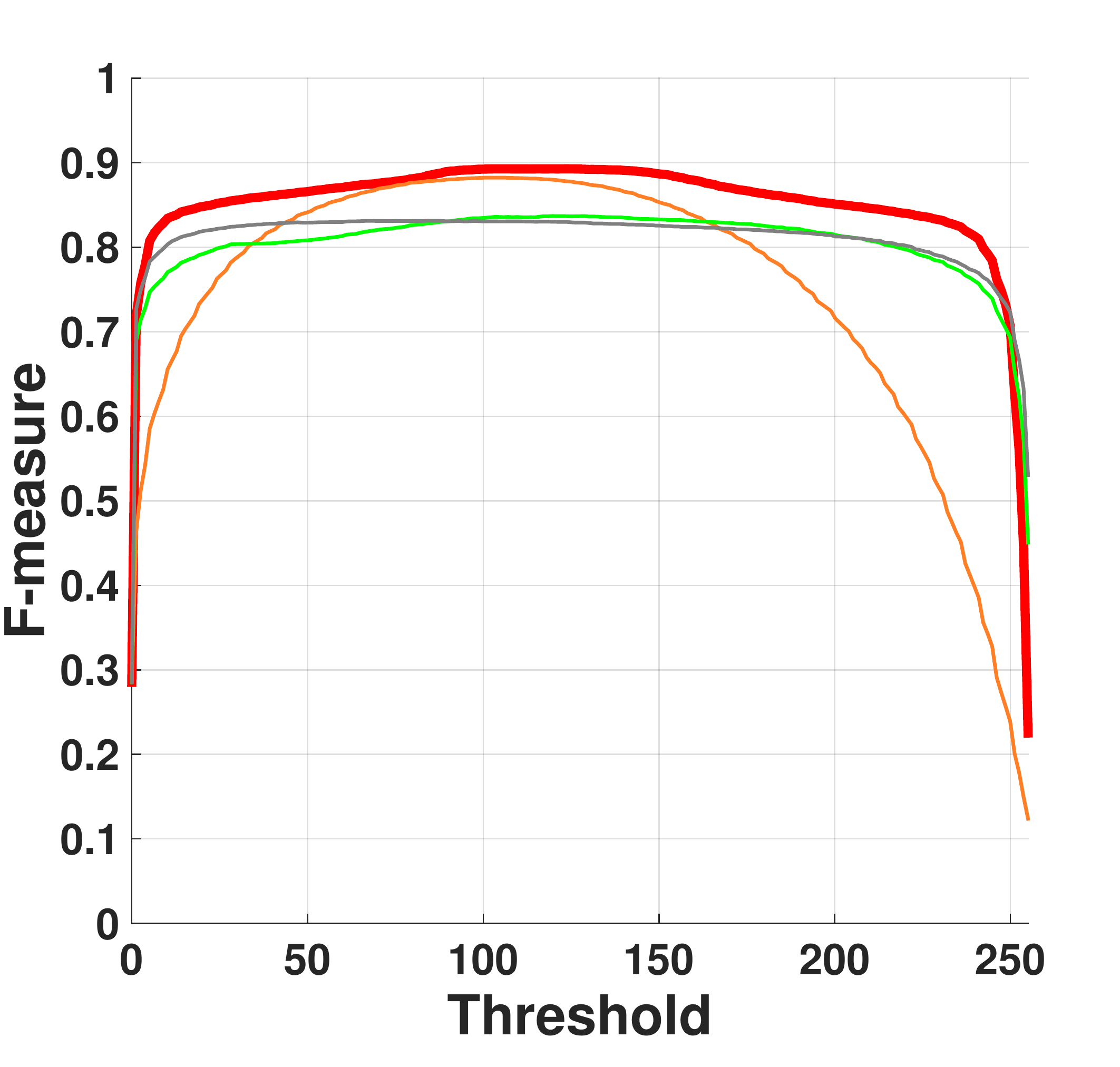}\label{integration-sta.b}}
  \subfigure[Score bars]{\includegraphics[width=5.2cm,height=4.5cm]{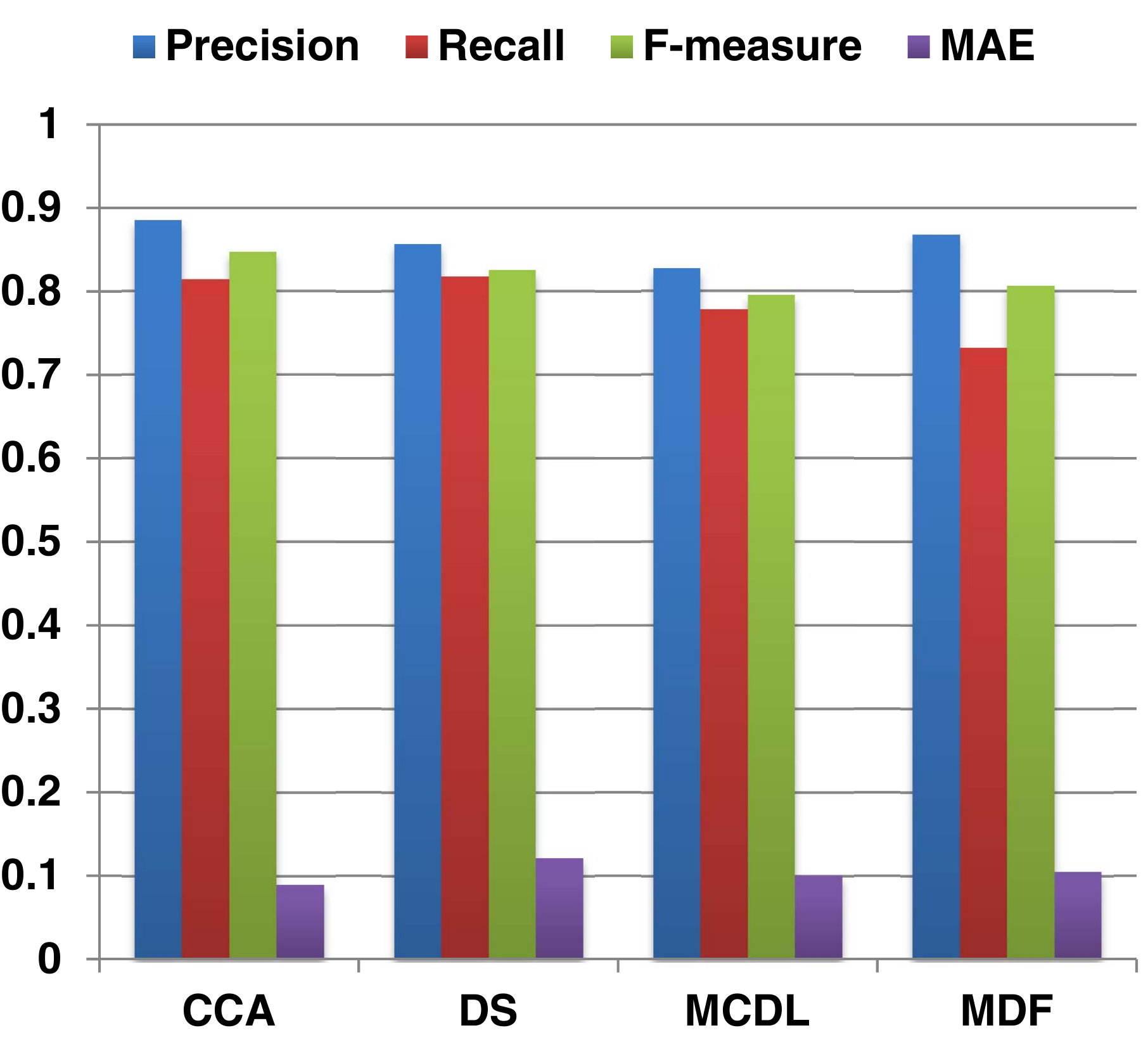}\label{integration-sta.c}}\\
  \caption{Effects of pixel-wise aggregation via Cuboid Cellular Automata on ECSSD dataset. The first row compares three conventional methods BL~\citep{tong2015bootstrap}, HS~\citep{yan2013hierarchical}, MR~\citep{yang2013saliency} and their integrated results via Cuboid Cellular Automata, denoted as CCA. The second row compares three deep learning models, e.g.~DS~\citep{Li2016DeepSaliency}, MCDL~\citep{zhao2015saliency}, MDF~\citep{li2015visual} and their integrated results. The precision, recall and F-measure scores in the right column are obtained by thresholding the saliency maps at twice the mean saliency value.}\label{Integration-sta}
\end{figure*}
\begin{figure*}
\center
  \includegraphics[width=5.7cm,height=4.5cm]{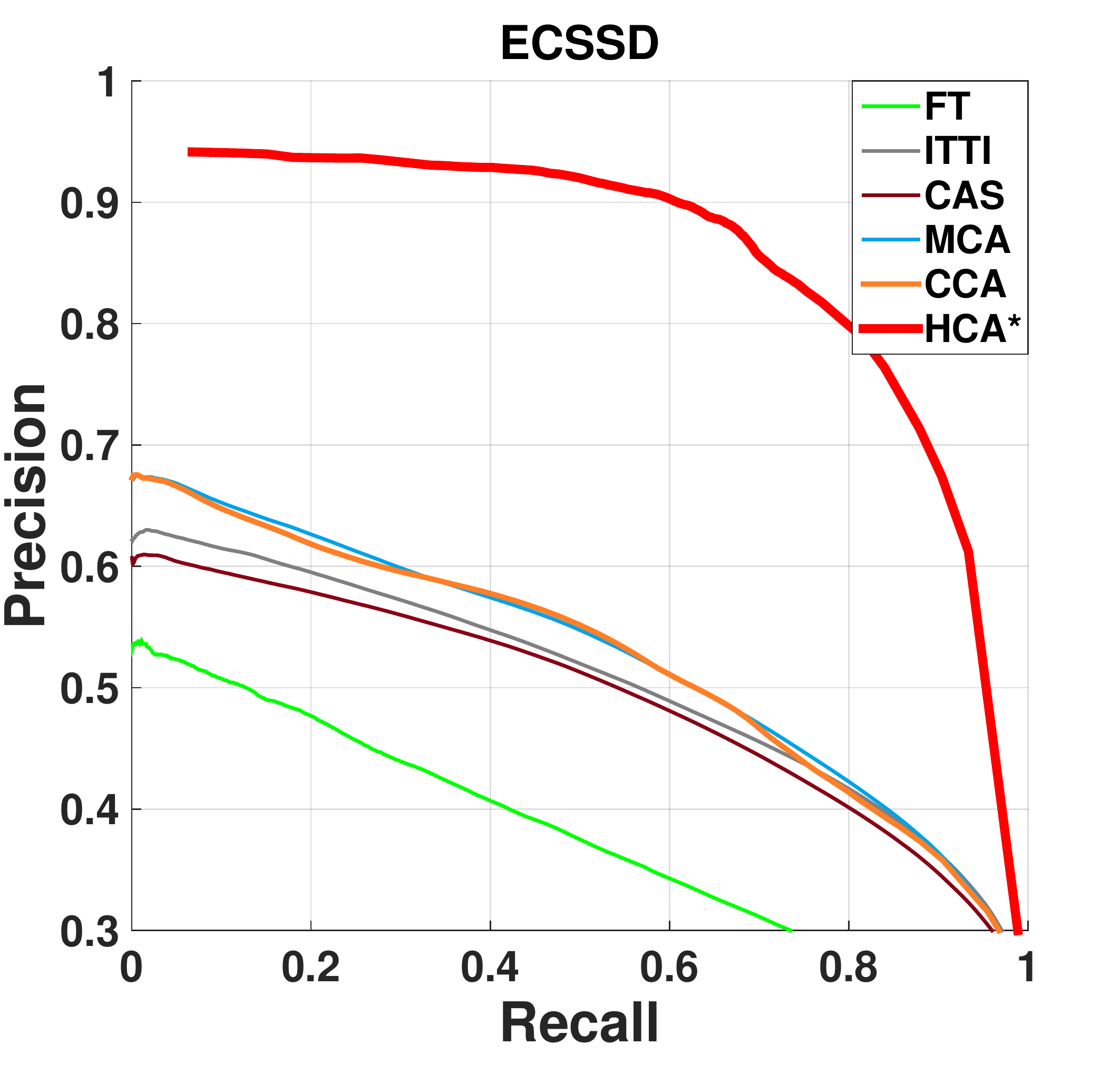}
  \includegraphics[width=5.7cm,height=4.5cm]{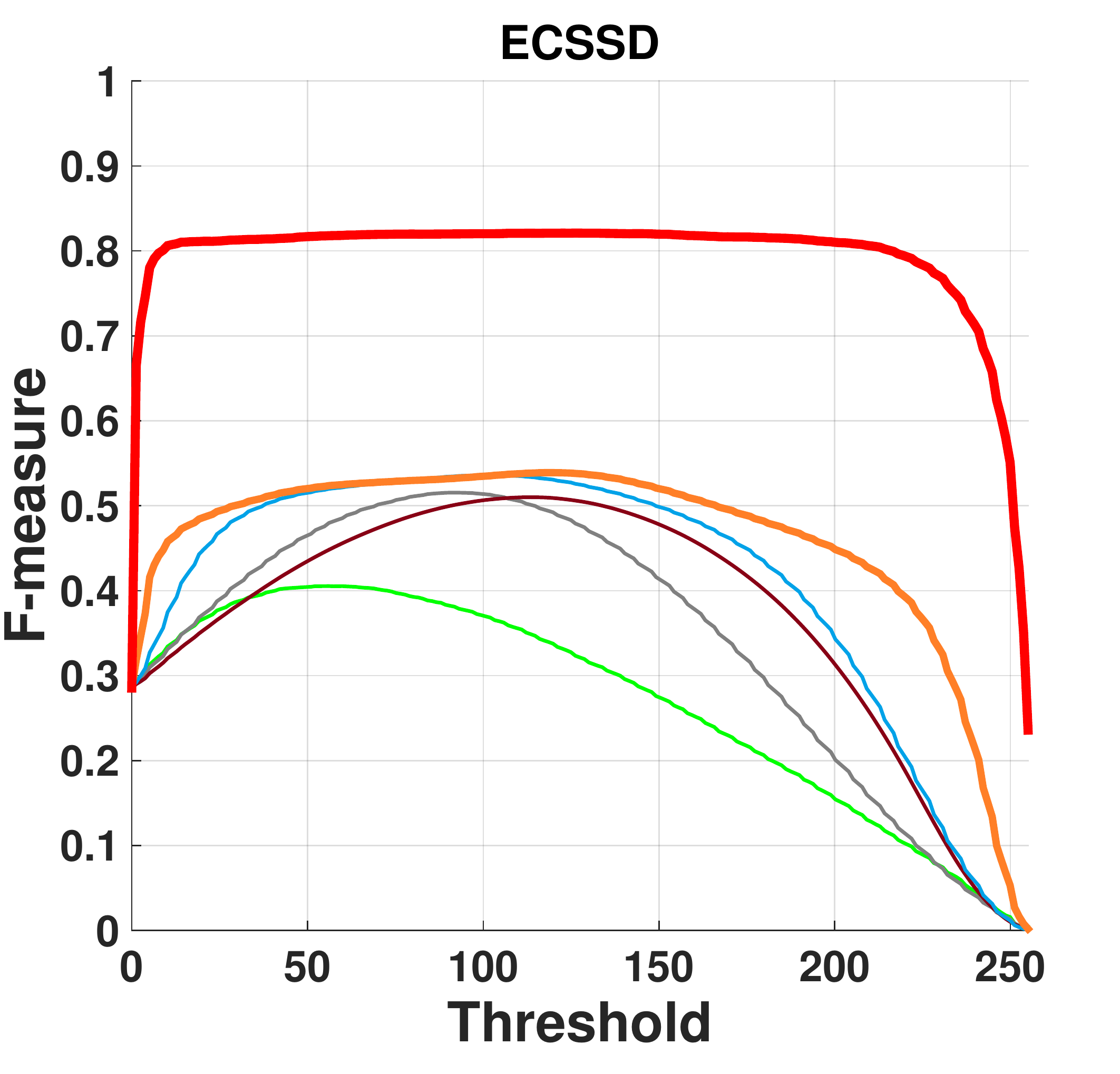}
  \includegraphics[width=5.7cm,height=4.5cm]{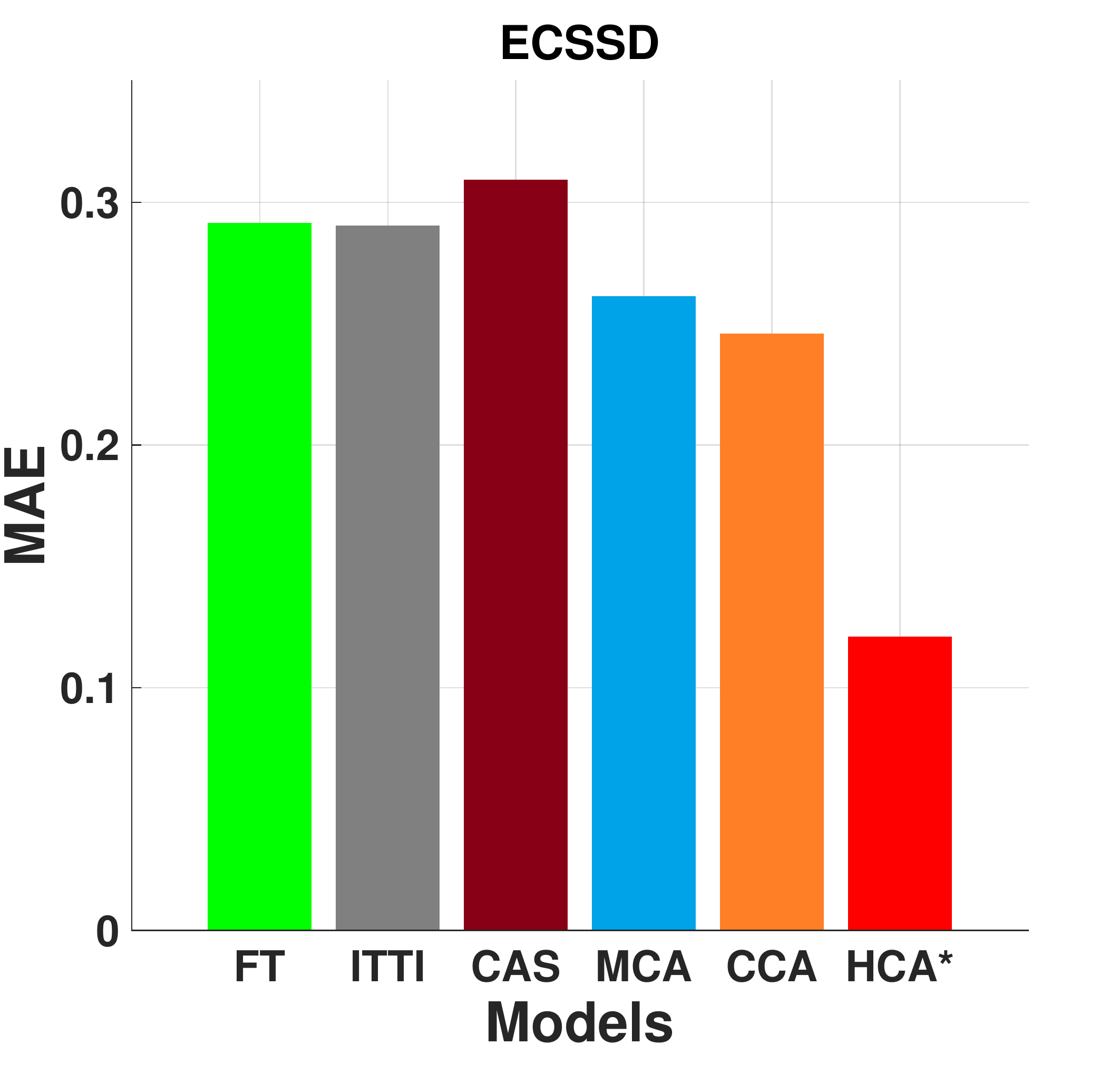}\\

  \subfigure[PR curves]{\includegraphics[width=5.7cm,height=4.5cm]{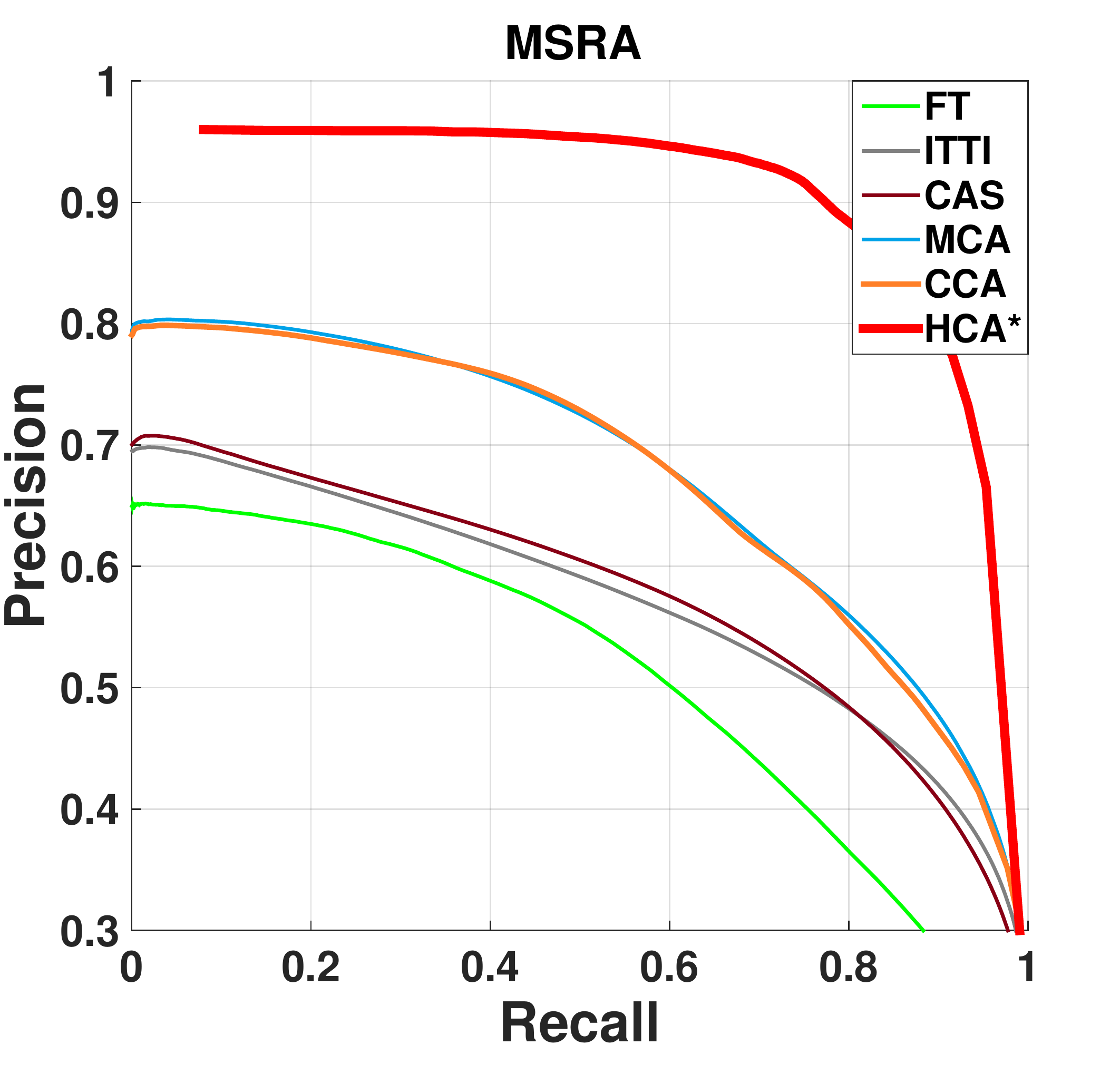}\label{cca-hca.a}}
  \subfigure[FT curves]{\includegraphics[width=5.7cm,height=4.5cm]{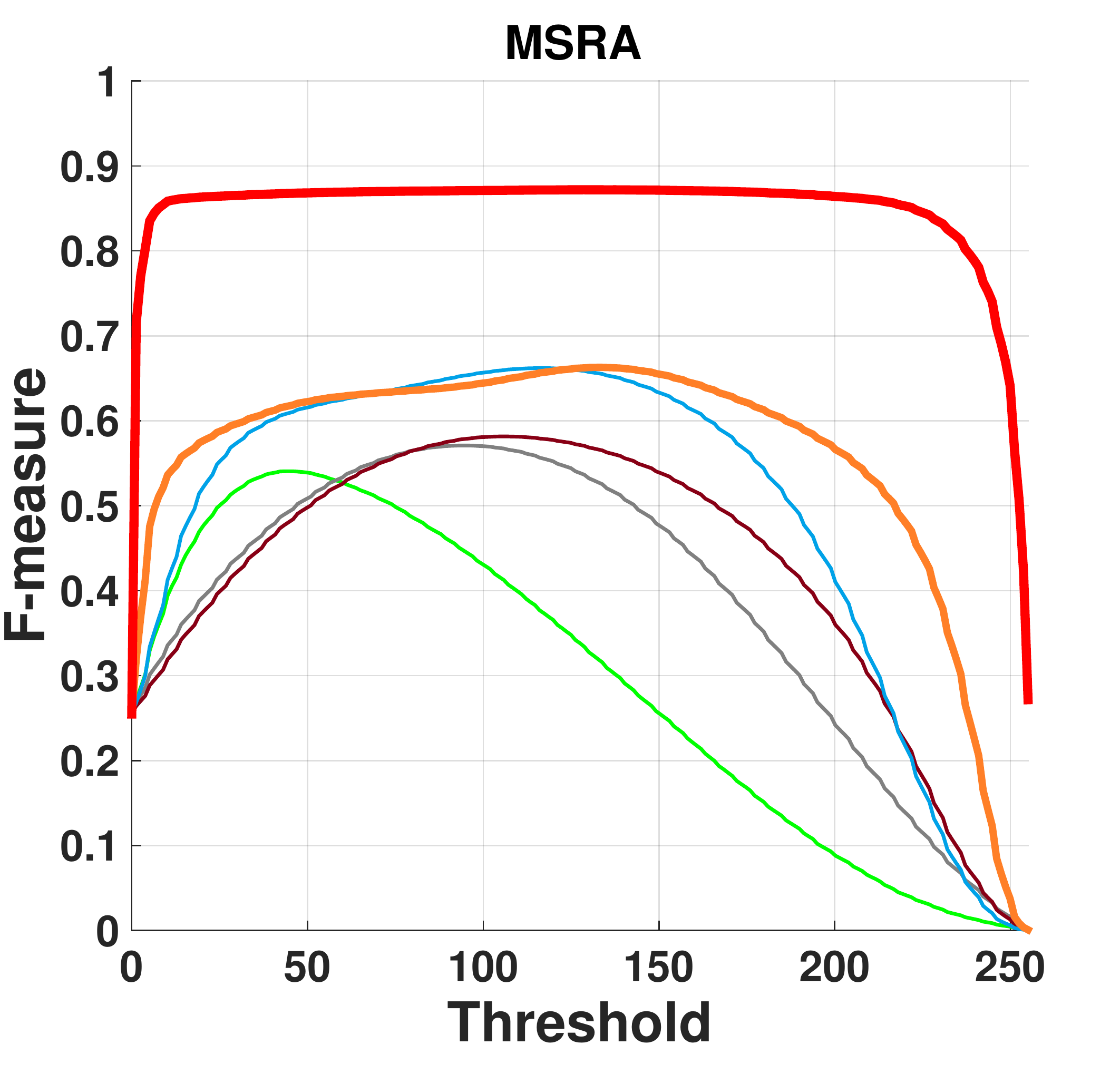}\label{cca-hca.b}}
  \subfigure[MAE scores]{\includegraphics[width=5.7cm,height=4.5cm]{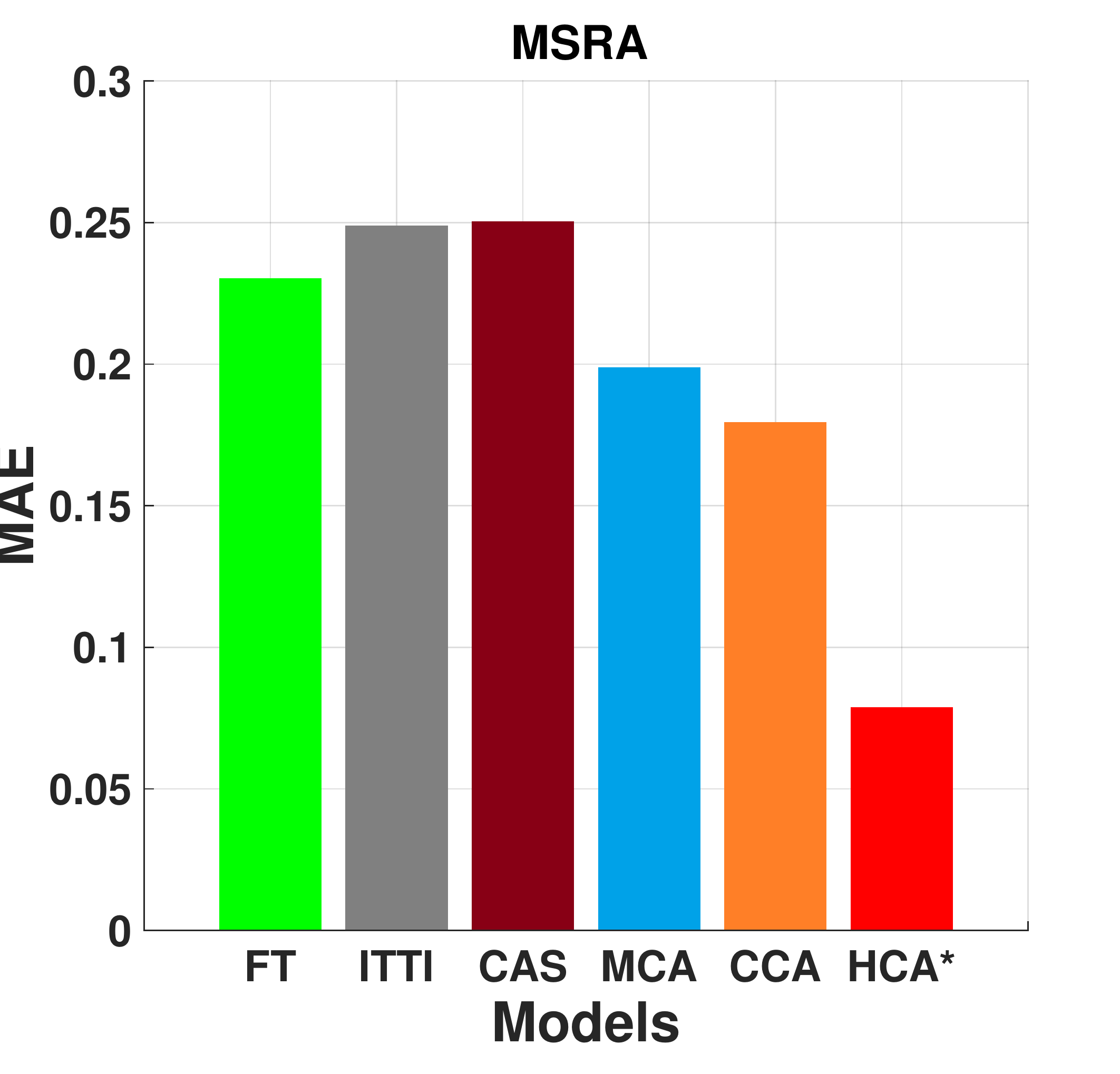}\label{cca-hca.c}}\\
  \vspace{-3mm}
  \caption{Comparison between three different integration methods MCA~\citep{Qin_2015_CVPR}, CCA and HCA when integrating FT~\citep{achanta2009frequency}, ITTI~\citep{itti1998model} and CAS~\citep{goferman2010context} on ECSSD and MSRA datasets.}\label{cca-hca-cmp}
\end{figure*}
\begin{table*}
\center
\vspace{-3mm}
\caption{Comparison of Run Time}\label{runtime-other}
\begin{adjustbox}{max width=0.996\textwidth}
\begin{tabu}{cccc | cccc | cccc}
\tabucline[1pt]{-}
  \hline
  \textbf{Model}  & \textbf{Year} & \textbf{Code} & \textbf{Time(s)}  &\textbf{Model}  & \textbf{Year} & \textbf{Code} & \textbf{Time(s)}& \textbf{Model}&\textbf{Year} & \textbf{Code} & \textbf{Time(s)} \\
  \hline
  \textbf{HCA} &     & Matlab & 1.4917 &\textbf{HDCT} & 2014 & Matlab & 5.1248 & \textbf{MR}  & 2013 & Matlab & 0.4542 \\

  \textbf{MCDL}  & 2015 & Python & 2.2521 &\textbf{wCO}  & 2014 & Matlab & 0.1484 &\textbf{XL13} & 2013 & Matlab & 65.5491\\

  \textbf{LEGS}  & 2015 & Matlab + C & 1.9050  &\textbf{DRFI} & 2013 & Matlab & 8.0104 &\textbf{LR}  & 2012 & Matlab & 10.0259      \\
\textbf{MDF} & 2015 & Matlab & 25.7328  & \textbf{DSR}  & 2013 & Matlab & 3.4796  &\textbf{RC}  & 2011 & C& 0.1360   \\

   \textbf{BL}  & 2015 & Matlab & 21.5161&  \textbf{HS}  & 2013 & EXE & 0.3821  &\textbf{CAS}  & 2010 & Matlab + C& 44.3270  \\
  \tabucline[1pt]{-}
\end{tabu}
\end{adjustbox}
\vspace{-3mm}
\end{table*}

\subsubsection{Effective Integration}
\vspace{-2mm}
In Section.~\ref{pwa}, we used Cuboid Cellular Automata as a pixel-wise aggregation method to integrate two groups of state-of-the-art methods. One group includes three of the latest conventional methods while the other contains three deep learning-based methods. We test the various methods on the ECSSD dataset, and the integrated result is denoted as CCA. PR curves in Figure.~\ref{integration-sta.a} demonstrate the effectiveness of CCA over all the individual methods. FT curves of CCA in Figure.~\ref{integration-sta.b} are fixed at high values that are insensitive to the thresholds. In addition, we binarize the saliency map with two times mean saliency value. From Figure.~\ref{integration-sta.c} we can see that the integrated result has higher precision, recall and F-measure scores compared to each method that is integrated. Also, the mean absolute errors of CCA are always the lowest. The fairly low mean absolute errors indicate that the integrated results are quite similar to the ground truth.

Although Cuboid Cellular Automata have exhibited great strength in integrating multiple saliency maps, they have a major drawback in that the integrated result highly relies on the precision of the saliency detection methods used as input. If saliency maps fed into Cuboid Cellular Automata are not well constructed, it cannot naturally detect the salient objects via interactions between these candidate saliency maps. HCA, however, can easily address this problem by incorporating single-layer propagation and multi-layer integration into a unified framework. Unlike MCA~\citep{Qin_2015_CVPR} and CCA, HCA can achieve better integrated saliency map regardless of their original detection performance through the application of SCA to clean up the initial maps. PR curves, FT curves and MAE scores in Figure.~\ref{cca-hca-cmp} show that 1) CCA has a better performance than MCA, as it considers the influence of adjacent cells on different layers. 2) HCA can greatly improve the aggregation results compared to MCA and CCA because it is independent of the initial saliency maps.
\vspace{-8mm}
\subsection{Run Time}
\vspace{-3mm}
The run time to process one image in{} the MSRA5000 dataset via Matlab R2014b-64bit with a PC equipped with an i7-4790k 3.60 GHz CPU and 32GB RAM is shown in Table~\ref{runtime-our}. The Table displays the average run time of each component in our algorithm, not including the time for extracting deep features. We can see that the Single-layer Cellular Automata and Cuboid Cellular Automata are very fast at processing one image, on average 0.06s. Their combination HCA takes only 0.2421s to process one image without superpixel segmentation and 1.0240s with SLIC.

We also compare the run time of our method with other state-of-the-art methods in Table~\ref{runtime-other}. Here we compute the run time including superpixel segmentation and feature extraction for all models. Our algorithm has the least run time compared to other deep learning based methods and is the fifth fastest overall.

\vspace{-6mm}
\section{Conclusion}
\vspace{-2mm}
In this paper, we propose an unsupervised Hierarchical Cellular Automata, a temporally evolving system for saliency detection. It incorporates two components,  Single-layer Cellular Automata (SCA), which can clean up noisy saliency maps, and Cuboid Cellular Automata (CCA), that can integrate multiple saliency maps. SCA is designed to exploit the intrinsic connectivity of saliency objects through interactions with neighbors. Low-level image features and high-level semantic information are both extracted from deep neural networks and incorporated into SCA to measure the similarity between neighbors. With superpixels on the image boundary chosen as the background seeds, SCA iteratively updates the saliency maps according to well-defined update rules, and salient objects naturally emerge under the influence of  their neighbors. This context-based propagation mechanism can improve the  saliency maps generated by existing methods to a high performance level. We used this in two ways: First, given a single saliency map, SCA can be applied to superpixels generated from the saliency map at multiple scales, and CCA can then integrate these into an improved saliency map. Second, we can take saliency maps generated by multiple methods, apply SCA (if necessary) to improve them, and then apply CCA to integrate them into better saliency maps. Our experimental results demonstrate the superior performance of our algorithms compared to existing methods.

\bibliographystyle{spbasic}      

%
%

\end{document}